\documentclass[10pt, conference, letterpaper]{IEEEtran}

\IEEEoverridecommandlockouts
\ifCLASSINFOpdf
   \usepackage[pdftex]{graphicx}
\else
   \usepackage[dvips]{graphicx}
\fi

\usepackage[nocompress]{cite}
\usepackage{amsmath,amssymb,amsfonts,amsthm}
\usepackage{algorithmic}
\usepackage[ruled,vlined]{algorithm2e}
\usepackage{wrapfig}
\usepackage{graphicx}
\usepackage{textcomp}
\usepackage{xcolor}
\usepackage{comment}
\usepackage{bm,bbm}
\usepackage{wrapfig}
\usepackage{subfigure}
\usepackage{caption}
\usepackage{multicol}
\usepackage{stfloats}
\usepackage{balance}

\usepackage{booktabs}
\usepackage{threeparttable}
\usepackage{multicol}
\usepackage{multirow}
\usepackage{url}

\newtheorem{definition}{\bf Definition}
\newtheorem{assumption}{Assumption}
\newtheorem{theorem}{Theorem}

\newtheorem{lemma}{Lemma}
\newtheorem{proposition}{Proposition}

\begin{document}
\title{Optimizing the Numbers of Queries and Replies in Federated Learning with Differential Privacy
}

\author{Yipeng~Zhou,~\IEEEmembership{}
        Xuezheng~Liu,~\IEEEmembership{}
        Yao~Fu,~\IEEEmembership{}
        Di~Wu,~\IEEEmembership{Senior Member,~IEEE,}
        Chao~Li~\IEEEmembership{}
        and~Shui~Yu~\IEEEmembership{}
\IEEEcompsocitemizethanks{\IEEEcompsocthanksitem Y. Zhou is with the Department of Computing, Faculty of Science and Engineering, Macquarie University, Sydney, NSW 2109, Australia.\protect\\
E-mail: yipeng.zhou@mq.edu.au.
\IEEEcompsocthanksitem X. Liu, F. Yao and D. Wu are with the School of Data and Computer Science, Sun Yat-sen University, Guangzhou 510275, China, and also
with the Guangdong Key Laboratory of Big Data Analysis and Processing,
Guangzhou, 510006, China.\protect\\
E-mail: {lxuezh, fuyao7}@mail2.sysu.edu.cn, wudi27@mail.sysu.edu.cn
\IEEEcompsocthanksitem Chao Li  is with Tencent Technology (Shenzhen) Co. Ltd, China.\protect\\
E-mail:  ethancli@tencent.com.
\IEEEcompsocthanksitem Shui Yu is with the School of Computer Science, University of Technology Sydney, Australia.\protect\\ E-mail: Shui.Yu@uts.edu.au.
}
}

\maketitle
\begin{abstract}
Federated learning (FL) empowers distributed clients to collaboratively train a shared machine learning model through exchanging parameter information. Despite the fact that FL can protect clients' raw data, malicious users can still crack original data with  disclosed parameters. To amend this flaw,  differential privacy (DP) is incorporated into FL clients to disturb original parameters, which however can significantly impair the accuracy of the trained model.  In this work, we study a  crucial question which has been vastly overlooked by existing works: what are the optimal numbers of queries and replies in FL with DP so that the final model accuracy is maximized. In FL,  the parameter server (PS) needs to query participating clients for multiple global iterations to complete training. Each client responds a query from the PS by conducting a local iteration. Our work investigates how many times the PS should query clients and how many times each client should reply the PS. We investigate two most extensively used DP mechanisms (i.e., the Laplace mechanism and Gaussian mechanisms).  Through conducting convergence rate analysis, we can determine the optimal numbers of queries and replies  in FL with DP so that the final model accuracy can be maximized.  Finally, extensive experiments are conducted with publicly available datasets: MNIST and FEMNIST, to verify our analysis and the results demonstrate that properly setting the numbers of queries and replies can significantly improve the final model accuracy in FL with DP. 
\end{abstract}


{\section{Introduction}\label{sec:introduction}}

\IEEEPARstart{F}{ederated} learning (FL) is contrived to preserve data privacy by maintaining local data private. A central parameter server (PS) is deployed to coordinate federated learning among decentralized clients, which is responsible for  aggregating parameters submitted by different clients and distributing aggregated parameters back to clients \cite{pmlr-v54-mcmahan17a, bonawitz2019towards}. Clients need to exchange parameters  with the PS for multiple global iterations to complete FL. In each global iteration,  participating clients conduct local iterations first, and then submit their parameters to the PS \cite{kairouz2019advances, yang2019federated}. In this process, participating clients never disclose their raw data samples to any entity. However, it can still result in severe privacy leakage. In works \cite{zhu2019deep, zhao2020idlg, wei2020framework, hitaj2017deep}, various attack algorithms have been designed to crack raw data samples based on  exposed parameter information.

In order to enhance the privacy protection level in FL,  tremendous efforts have been dedicated to exploring how to incorporate differential privacy (DP) mechanisms into FL clients \cite{10.1145/3378679.3394533,  geyer2017differentially, choudhury2019differential}. Rather than submitting plain parameters to the PS, DP mechanisms add additional zero-mean noises to distort parameters, which also inevitably impairs the model accuracy.  From a particular client's perspective, the variance of DP noises should be amplified with  the number of times the client needs to submit parameters, \emph{i.e.}, the number of queries the client needs to reply, because replying more queries leaks more privacy unless the noise variance is amplified accordingly  \cite{wu2019value, wei2020federated, Abadi_2016}. 

Existing works mainly focused on designing various DP mechanisms applicable in FL. Wei et al. in \cite{wei2020federated} incorporated the Laplace mechanism into FL clients, and the convergence rate with Laplace noises was analyzed. In \cite{Abadi_2016}, Abadi et al.  refined the Gaussian mechanism by proposing a variant with a much lower variance in deep learning. In \cite{10.1145/3378679.3394533}, Truex et al.  discussed how to incorporate LDP (local differential privacy) into FL. In \cite{geyer2017differentially}, Geyer et al. proposed to add DP noises on aggregated parameters instead of parameters of individual clients.  Aleksei et al. incorporated relaxed Bayesian DP into FL \cite{triastcyn2019federated} to improve the bound of privacy loss. 

Orthogonal to existing works regarding the design of  DP mechanisms in FL, our work investigates how the numbers of queries and replies affect the final model accuracy.  In fact, no matter which DP mechanism is adopted in FL, the numbers of queries and replies should always be optimized because the variance of DP noises is dependent on the number of queries a client needs to respond. The utility  of FL lies in achieving a shared useful model with certain accuracy.  Poorly setting the numbers of queries and replies in FL with DP can  considerably deteriorate the model accuracy, and hence prohibits the implementation of DP in real FL systems \cite{jayaraman2019evaluating}.  
   
Prior to discussing how to optimize the numbers of queries and replies in FL with DP, we need to understand the role of each query. Considering FL without DP, each iteration can improve the model accuracy a little bit on expectation.  Thus, the model is more accurate if  clients return query results, \emph{i.e.}, model parameters, to the PS for more times  \cite{Li2020On}. \emph{However, with DP in FL,  the variance of a client's DP noises is  amplified with the number of queries the client needs to reply \cite{dwork2014algorithmic}.} The negative influence of DP noises can be very significant if clients reply an excessive number of queries from the PS, which has been validated in the work \cite{jayaraman2019evaluating}.  
Thus, to reach the highest model accuracy with DP in FL,  it is necessary to take prons and cons of each query into account.

To address this problem, we investigate how each query influences the final model accuracy with two typical DP mechanisms, {i.e.}, the Laplace mechanism and Gaussian mechanisms). To ease our discussion, we propose the FedSGD with DP on clients (FedSGD-DPC) algorithmic framework that  can  incorporate different DP mechanisms into FL. It is designed  based on the FedSGD algorithm \cite{pmlr-v54-mcmahan17a}, one of the most fundamental model average algorithms in FL. Based on FedSGD-DPC, the convergence rate with each DP mechanism is derived. Optimizing the final  model accuracy is converted to minimizing the convergence rate function (equivalent to minimizing the loss function) with respect to the numbers of queries by the PS and replies by clients.  We prove that this is a biconvex problem, which can be solved efficiently with methods in \cite{gorski2007biconvex}. In addition,  each closed-form solution of this problem is derived and discussed to reveal the implications of each feasible solution. At last, public datasets (MNIST and FEMNIST) are exploited to conduct experiments to verify the correctness of  our analysis.  

In summary, our work yielded the following contributions.
\begin{itemize}
    \item We investigate the optimization of the numbers of queries and replies with two most classical DP mechanisms (\emph{i.e.}, the Laplace and the Gaussian mechanisms) in FL. Our analysis leverages a generic algorithm framework, FedSGD-DPC, which can incorporate different DP mechanisms into FL clients. The convergence rate with each DP mechanism is derived. 
    \item For each DP mechanism, through minimizing the convergence rate function, we derive the optimal numbers of queries and replies in order to maximize the model accuracy. 
    \item Extensive experiments are carried out with the MNIST and FEMNIST datasets to verify the correctness of our analysis, and also demonstrate the importance to set the numbers of queries and replies  properly in FL with DP mechanisms. 
\end{itemize}

The rest is organized as below. Related works are discussed in Sec.~\ref{Sec:Related}. Preliminaries regarding FedSGD and differential privacy mechanisms are introduced in Sec.~\ref{sec:preliminary}. FedSGD-DPC and two DP mechanisms are explored in Sec.~\ref{Sec:Alg}. The convergence rates and the closed-form solutions of the optimal number of queries are derived for the Laplace mechanism and the Gaussian mechanism  in Sec.~\ref{Sec:AnalysisLaplace} and Sec.~\ref{Sec:AnalysisGaussian}, respectively.  Experiment results are shown and discussed in Sec.~\ref{Sec:Exp}. Finally, we conclude our work in Sec.~\ref{Sec:Conclusion}.

\section{Related Work}
\label{Sec:Related}

\subsection{Federated Learning}

The FL paradigm is still at the infant stage. It was firstly  proposed and applied by Google to predict the next word for mobile users in \cite{pmlr-v54-mcmahan17a,googleFL, bonawitz2019towards}.  In~\cite{pmlr-v54-mcmahan17a},  FedAvg and FedSGD algorithms were designed  and empirically studied with extensive experiments. FedAvg and FedSGD are two most fundamental model average algorithms in FL, which have been extensively used by various FL systems \cite{luo2020cost, chen2020convergence, luping2019cmfl, liu2020accelerating}. Later on, Li et al. \cite{Li2020On}  derived the convergence rate of FedAvg/FedSGD with non-iid (not independent and identically distributed) data sample distributions.

Due to the capability to preserve data privacy, FL has received tremendous research efforts. {Kairouz et al.}~\cite{kairouz2019advances} and {Li et al.}~\cite{Li2020FederatedLC} have conducted holistic overviews of FL with  in-depth  discussions on the potential applications of FL and its weaknesses calling for further research efforts. A particularly emphasized problem is that transmitting unprotected parameter information can make FL vulnerable to malicious attacks. Then, it was extensively studied in works \cite{wei2020framework, zhao2020idlg, zhu2019deep,236216,8737416}, which verified that the exposure of parameters or gradients can cause the leakage of data privacy because malicious attackers can utilize exposed parameter information to largely crack original samples. Thereby, in a word, how to protect parameters in FL is a challenging but  vital problem.

\subsection{Differential Privacy}

Differential privacy was originally proposed to protect the data privacy in databases.  DP mechanisms add zero-mean noises to query results before they are exposed. Malicious attackers cannot exactly crack users' private information if they can only access query results disturbed by noises. Two most frequently used mechanisms to generate noises in FL are the Laplace  and the Gaussian mechanisms \cite{dwork2014algorithmic}. 
Due to the capability to protect data privacy, DP has been applied in various applications such as  personalized recommendation \cite{8290673}, location based services \cite{ding2017collecting}, meta learning \cite{li2020differentially} and databases \cite{inan2020sensitivity}.

However, the drawback of DP is that the accuracy of the query results  is lowered  because true results are disturbed by DP noises \cite{bhowmick2018protection, wu2019value, liao2019di-prida}.  How to alleviate the influence caused by DP noises without compromising the privacy protection level is always a challenging research problem.

\subsection{Applying DP in FL}

In FL, each client maintains a private dataset. In each round of global iteration, each participating client needs to respond the query of the PS by uploading its computation results obtained with local samples to the PS. It is equivalent to exposing its query results to the public. Naturally, DP can be applied to disturb the query results against differential attacks \cite{lyu2020threats}.

Recently, various works have been dedicated to exploiting the design of DP mechanisms in  order to protect exposed parameters in FL such as \cite{bhowmick2018protection,pihur2018differentiallyprivate,seif2020wireless, wei2020federated, wu2019value}. 
In particular, in \cite{wu2019value}, Wu et al. derived the convergence rate of a distributed learning system by incorporating the Laplace mechanism with the composition rule into clients. In \cite{wei2020federated}, Wei et al. further studied this problem specialized for FL, in which only a fraction of clients can be selected to participate each global iteration. 
LDP (local differential privacy) is particularly suitable for training personalized machine learning models in FL. In \cite{10.1145/3378679.3394533}, Truex et al. explored the feasibility to incorporate LDP into FL. In \cite{8290673} Shin et al. proposed the privacy enhanced matrix factorization with LDP in FL systems, in which  individual users add noises to their rating data before they collaboratively train personalized recommendation models.  Liu et al. proposed the FedSel algorithm that only selects the most significant dimension for adding noises  so as to reduce the variance of LDP noises  \cite{liu2020fedsel}. 
To alleviate the influence of DP noises, {Seif et al.} proposed to apply  relaxed local DP with wireless FL settings \cite{seif2020wireless}, which yielded fixed noises in each iteration. An advanced composition rule is adopted to track the total privacy leakage over the entire process of FL training. However,  the privacy leakage will be  infinity if the number of iterations goes to infinity.

In summary,  existing works have exhibited that it is crucial to incorporate DP mechanisms into FL  which can effectively ban attacks towards cracking data privacy in  FL. Nevertheless, achieving high model accuracy is also the main goal of FL. If DP noises severely deteriorate the final model accuracy, DP is an impractical technique in real systems. Thus, Optimizing the final model accuracy in FL without compromising the privacy protection level significantly is a vital unsolved problem. 


\section{Preliminaries}\label{sec:preliminary}

We introduce the FedSGD algorithm~\cite{pmlr-v54-mcmahan17a} and the concept of DP in this section. 
\subsection{FedSGD Algorithm}
In a typical FL system, we can assume that there are a certain number of clients denoted by $\mathcal{N}$ with cardinality $N$ who will  train a shared model together. The model is denoted by the function  $F(\theta)$.  Here $\theta\in\mathbb{R}^p$ represents the parameters of the model and $p$ is the dimension of paramters.  Each client $i$ maintains a local dataset $\mathcal{D}_i$ with cardinality $d_i$. Let $d = \sum_{i=1}^Nd_i$.

In FL, a parameter server (PS) must be deployed that is responsible for  aggregating the computation results submitted by clients and distributing the aggregated model parameters back to clients in each round of global iteration. 
FL systems commonly   train the model $F(\theta)$ via exchanging model parameters between the PS and clients for multiple iterations (such as the  FedSGD algorithm  proposed in \cite{pmlr-v54-mcmahan17a}) which can iteratively reduce the loss function.   

Without loss of generality, we assume that the entire training process lasts $T$ rounds of global iterations. In the $t^{th}$ global iteration, the PS randomly involves a number of clients, denoted by the set $\mathcal{P}_t$, for participation. Each participating client  updates the model parameters received from the PS by conducting a local iteration with its own dataset. Then, the client returns its computation results to the PS. In summary, the FedSGD algorithm works as follows.

\begin{itemize}
    \item{\bf Step 1 (Start of a global iteration):} The PS sends out the latest model parameters $\theta$ (which may be randomly initialized in the first round of global iteration) to participating clients in the set $\mathcal{P}_t$. 
    \item{\bf Step 2 (Local iteration):} Each participating client conducts a local iteration with the local dateset and the  model parameters are updated as below: 
    \begin{equation}
    \label{EQ:LocalIteration}
        \theta\leftarrow\theta-\eta\nabla F(\theta, \mathcal{D}_i),   
    \end{equation}
    where $\mathcal{D}_i$ represents the local dataset of client $i$ and $\eta$ is the learning rate.  Then, each participating client returns the updated model parameters to the PS. 
    \item {\bf Step 3 (Parameter aggregation):} The PS aggregates results returned from all participating clients. Suppose it is the $(t+1)^{th}$ global iteration and the size of $\mathcal{P}_t $ is $b$, the aggregation rule according to \cite{Li2020On} is 
    \begin{equation*}
        \theta_{t+1}\leftarrow\frac{N}{b}\sum_{i\in\mathcal{P}_t}\beta_i\theta_{t+1}^{i}.
    \end{equation*}
    Here $\beta_i$ is the weight of client $i$. Conventionally, $\beta_i$ is set as $\frac{d_i}{d}$.
    \item {\bf Step 4 (End of a global iteration):} The algorithm terminates if the termination conditions are met, Otherwise, go back to step 1 to kick off a new round of  global iteration. 
\end{itemize}

In FedSGD,  gradients are computed as $\nabla F(\theta, \mathcal{D}) = \frac{1}{|\mathcal{D}|}\sum_{\forall\zeta \in \mathcal{D}}\nabla F(\theta, \zeta) $ where $\zeta$ represents a single sample in any given sample set $\mathcal{D}$. For simplicity, we let $ F_i(\theta ) = \sum_{\forall \zeta\in \mathcal{D}_i }\frac{1}{d_i}F(\theta, \zeta) $ and $ \nabla F_i(\theta ) = \sum_{\forall \zeta\in \mathcal{D}_i }\frac{1}{d_i}\nabla F(\theta, \zeta) $.
To facilitate our discussion, let $T$ denote the total number of global iterations and $t$  denote the index of global iterations.

\subsection{Differential Privacy}

Differential privacy (DP) is a technique to disturb the output of a query function with randomly generated noises so as to protect the true information. 
We briefly introduce the  concepts or  DP according to previous works \cite{10.1007/11787006_1, dwork2014algorithmic}.

\begin{definition}[($\epsilon,\delta$)-Differential Privacy]
Let $\mathfrak{M}$ denote a randomized function. $\mathfrak{M}$ gives ($\epsilon,\delta$)-differential privacy if for any data sets $\mathcal{D}$ and $\mathcal{D}'$ differing on at most one entry and
\begin{equation*}
    \Pr\left\{\mathfrak{M}(\mathcal{D})\in \mathcal{S}\right\}\le\exp(\epsilon)\times\Pr\left\{\mathfrak{M}(\mathcal{D}')\in \mathcal{S}\right\}+\delta,
\end{equation*}
$\forall \mathcal{S}\subset$ where $Range(\mathfrak{M}) $ is the range of the output of $\mathfrak{M}$ and $\mathcal{S}$ is any subset of $Range(\mathfrak{M}) $.
\end{definition}

Here, $(\epsilon,\delta)$ is regarded as the privacy budget. Smaller $\epsilon$ and $\delta$ imply a stronger level of privacy protection.  In our work, $(\epsilon,\delta)$ are considered as constants determined by how much the loss of  privacy a client can tolerate.


According to prior works  \cite{zhu2019deep,zhao2020idlg,236216,8737416}, clients' sample information can be cracked through analyzing exposed parameter information. 
Thus, the purpose to incorporate DP into FL is to protect the sensitive parameter information of each client. According to Eq.~\eqref{EQ:LocalIteration}, local iterations are conducted based on the global parameters distributed by the PS. We can deduce that disturbing gradients is equivalent to disturbing parameters. 
Let $\mathbf{g}_t^i$ denote the gradients of client $i$ in global iteration $t$.  
Following the FedSGD algorithm, in  global iteration $t$, the  gradients $\mathbf{g}_{t}^i$ returned by client $i$ can be distorted by DP noises denoted by $\mathbf{w}^i_{t}$.  It turns out that the output gradients of  client $i$ are
\begin{equation}
\label{EQ:ThetaAddNoise}
\mathfrak{M}_i(\mathbf{g}_t^i) =  \mathbf{g}_t^i+ \mathbf{w}^i_{t}.
\end{equation}

The resulted parameters returned by client $i$ in global iteration $t$ are
\begin{equation}
\label{EQ:Returnbyit}
\theta^i_{t+1}  = \theta_t-\eta_t \mathfrak{M}_i(\mathbf{g}_t^i).
\end{equation}
Here $\eta_t$ is the learning rate in global iteration $t$.
The distribution of $\mathbf{w}^i_{t}$ is jointly determined by the DP mechanism, the privacy budget $(\epsilon_i, \delta_i)$, and the number of replies, \emph{i.e.} how many times each client needs to return its parameters to the PS. $\mathbf{w}^i_{t}$ will be specified later after we introduce the detailed DP mechanisms.  

\section{FL Framework to Incorporate DP}
\label{Sec:Alg}

In this section, we sketch the FedSGD algorithm with DP mechanisms in FL, and particularly introduce the Laplace and Gaussian mechanisms.

\subsection{FedSGD-DPC Algorithm}

\begin{algorithm}
\SetAlgoLined
\DontPrintSemicolon

\SetKwProg{Fn}{ServerUpdate}{:}{}
\Fn{}{
 initialize $\theta_0$\;
\For{$t\leftarrow 0$ \KwTo\ $T-1$}{
 The PS choose $b$ clients  with a round robin manner, which  form the set $\mathcal{P}_t\subset\mathcal{N}$\;
 \ForEach{client $i\in\mathcal{P}_t$ \textbf{in parallel}}{
  ${\theta}_{t+1}^i\leftarrow$ClientUpdate($\theta_{t}$)\;
 }
 $\theta_{t+1}\leftarrow\frac{N}{b}\sum_{i\in\mathcal{P}_t}\frac{d_i}{d}{\theta}_{t+1}^i$\;
 }
}
\;
\SetKwProg{Fn}{ClientUpdate}{:}{}
\Fn{$(\theta_t)$}{
$\theta_t^i\leftarrow\theta_t$\;
Randomly select $q$ fraction of samples from $\mathcal{D}_i$ to form batch $\mathcal{B}_t^i$. Here $q\in (0, 1]$ is determined by $\mathfrak{M}_i$  \;
$\mathbf{g}_{t+1}^i\leftarrow\nabla F_i(\theta_t^i, \mathcal{B}_t^i)$\;
Generate $\mathbf{w}_{t+1}^i$ according to the DP mechanism $\mathfrak{M}_i$\;
${\theta}_{t+1}^i\leftarrow \theta_{t}^i-\eta_t (\mathbf{g}_{t+1}^i+\mathbf{w}_{t+1}^i)$\;
\KwRet{${\theta}_{t+1}^i$}
}

\caption{The FedSGD with DP on Clients (FedSGD-DPC) Algorithm.}
\label{alg:fedminibatchsgd}
\end{algorithm}

The FedSGD with DP on Clients (FedSGD-DPC) Algorithm is presented in Alg.~\ref{alg:fedminibatchsgd}. FedSGD-DPC totally  conducts $T$ global iterations and each time $b$ out of $N$ clients are randomly selected to participate each global iteration. 
It is worth to mention that FedSGD-DPC is a generic algorithm framework for FL with DP since  the distribution of the noises $\mathbf{w}_{t+1}^i$ has not been specified. 

In Alg.~\ref{alg:fedminibatchsgd}, \emph{the total number of queries issued by the PS  is 
$T$ and the total number of replies
from each client is  $\frac{Tb}{N}$.}
Since $N$ is regarded as a constant, our problem is to optimize the values of $T$ and $b$ so as to maximize the final model accuracy. 

Note that the aggregation rule in Alg.~\ref{alg:fedminibatchsgd}, \emph{i.e.}, $\theta_{t+1}\leftarrow\frac{N}{b}\sum_{i\in\mathcal{P}_t}\frac{d_i}{d}{\theta}_{t+1}^i$, is from \cite{Li2020On}. $q$ is the sampling rate determining the fraction of samples selected to participate local iterations. For the Laplace mechanism $q=1$, while for the Gaussian mechanism $q<1$,\footnote{Note that the sample batch $\mathcal{B}_t^i$ will be identical with $\mathcal{D}_i$ if $q=1$.} which will be further discussed in the next subsection.

For convenience, we show the overall iteration rules of Algorithm \ref{alg:fedminibatchsgd} as below for our analysis:
\begin{equation}\label{equ:update}
\begin{split}
 &\nu_{t+1}^i=\theta_t-\eta_t\nabla F_i(\theta_t),\\
        &\theta_{t+1}=\frac{N}{b}\sum_{i\in\mathcal{P}_t}\frac{d_i}{d}\nu_{t+1}^i-\mathbf{w}_{t}^b,
\end{split}
\end{equation}
where
\begin{equation*}
 \mathbf{w}_{t}^b=\frac{N}{b}\sum_{i\in\mathcal{P}_t}\frac{d_i}{d}\eta_t\mathbf{w}_{t}^i.  
\end{equation*}
Here $ \mathbf{w}_{t}^b$ is the average of noises from $b$ selected clients. To ease our discussion, we define $\nu_{t+1}^i$ as the intermediate variable to represent parameters without noises.

\subsection{Applicable DP Mechanism}

In our work, we analyze two most frequently used DP mechanisms: the Laplace mechanism and the Gaussian mechanism.\footnote{In fact, the FedSGD-DPC algorithm framework is also applicable for other DP mechanisms in FL. } 

\subsubsection{Laplace Mechanism}
\begin{definition}
($l_1$-sensitivity). Given an arbitrary query function $f(\mathcal{D})$ where $\mathcal{D}$ is the set of input, the $l_1$-sensitivity is defined as 
$\max_{\forall \mathcal{D}, \mathcal{D}': \|\mathcal{D} - \mathcal{D}'\|_1  = 1}\| f(\mathcal{D}) - f(\mathcal{D}') \|_1$
where $\mathcal{D}$ and $\mathcal{D}'$ are two adjacent input sets, \emph{i.e.}, $\|\mathcal{D}-\mathcal{D}'\|_1 =1 $.   
\end{definition}

\begin{assumption}
\label{ASP:L1Norm}
The $l_1$-norm of the gradients of the loss function $F_i$ is uniformly bounded: $\xi_1 = \max_{t= 1, \dots, T,\forall i, \theta_t^i, \forall\zeta\in \mathcal{D}_i}\| \nabla F_i(\theta_t^i, \zeta)\|_1< \infty$.
\end{assumption}

Let $(\epsilon_i, \delta_i)$ denote the privacy budget of client $i$. According to Alg.~\ref{alg:fedminibatchsgd},  each client $i$ needs to respond $\frac{bT}{N}$ queries to the PS given that the probability to participate each global iteration is $\frac{b}{N}$ for each client.
Then, $\mathfrak{M}_i $ is  $(\epsilon_i, 0)$-differential privacy ($\epsilon_i$-DP) if the following conditions are met. 

\begin{theorem}[Laplace Mechanism \cite{wu2019value}]\label{theorem:Lap}
$\mathfrak{M}_i$ is $(\epsilon_i, 0)$-differential privacy over $T$ global iterations if $\mathbf{w}_t^i$ is draw from $Lap\left(0, \frac{2bT\xi_1}{Nd_i\epsilon_i}\mathbb{I}_p\right)$ 
where $\mathbb{I}_p$ represents the identity matrix with rank $p$ and $Lap$ represents the Laplace distribution. 
\end{theorem}
Here $\frac{bT}{N}$ is the number of global iterations the client $i$ participates and  $\frac{2\xi_1}{d_i}$ is the $l_1$-sensitivity of $\nabla F_i$.  The detailed proof is provided by Theorem 1 in \cite{wu2019value}.

The Laplace mechanism can provide a very strong privacy protection.
Because the noise variance of each individual client is  $\mathbb{E}\{\|\mathbf{w}_t^i\|_2^2\}= \frac{8pb^2T^2\xi_1^2}{N^2d_i^2\epsilon_i^2}$, we derive:

\begin{proposition}[Variance of aggregated Laplace noises]\label{pro:noise}
The  variance of the aggregated DP noises of $b$ randomly selected clients using the Laplace mechanism in FedSGD-DPC is,
\begin{equation*}
    \mathbb{E}\left\{\left\|\mathbf{w}_{t}^b\right\|_2^2\right\}=\frac{8\eta_t^2pbT^2\xi_1^2}{Nd^2}\sum_{i\in \mathcal{N}} \frac{1}{\epsilon_i^2}.
\end{equation*}
\end{proposition}
The proof is presented in Appendix \ref{APP:Prop1}.
From the Proposition~\ref{pro:noise}, we can observe that the variance of $\mathbf{w}_t^b$ is directly determined by $b$ and $T$. 
Lowering $b$ and $T$ can reduce the variance of DP noises, which however also lowers the number of iterations. We investigate how to maximize the model accuracy by choosing proper $b$ and $T$ through convergence rate analysis in the next section. 

\subsubsection{Gaussian Mechanism}

Other than the Laplace mechanism, the Gaussian mechanism has been widely applied in FL as well.  The protection of the Gaussian mechanism replies on  adding noises generated from the Gaussian distribution. According to the work \cite{Abadi_2016}, the Gaussian mechanism can have a much tighter bound of noise variance. Hence,  the noise influence  of the Gaussian mechanism is much smaller  than that of the Laplace mechanism. 
\begin{definition}
($l_2$-sensitivity). Given an arbitrary query function $f(\mathcal{D})$ where $\mathcal{D}$ is the set of input, the $l_2$-sensitivity is defined as 
$\max_{\forall \mathcal{D}, \mathcal{D}': \|\mathcal{D} - \mathcal{D}'\|_1  = 1}\| f(\mathcal{D}) - f(\mathcal{D}') \|_2$
where $\mathcal{D}$ and $\mathcal{D}'$ are two adjacent datasets.   
\end{definition}

\begin{assumption}
\label{ASP:GradientBound}
The $l_2$-norm of $\nabla F_i$ is uniformly bounded: $\xi_2 = \max_{t=1,\dots, T,\forall i, \theta_t^i, \forall \zeta\in \mathcal{D}_i}\| \nabla F_i(\theta_t^i, \zeta)\|_2< \infty$.
\end{assumption}

According to \cite{Abadi_2016}, the variance of Gaussian noises can grow  much slower than that of Laplace noises  with the numbers of queries and replies. It turns out that:

\begin{theorem}[Gaussian Mechanism~\cite{Abadi_2016}]\label{THE:Gaussian}
There exist constants $c_1$ and $c_2$ so that given the sampling probability $q\in(0,1)$ (\emph{i.e.}, randomly select $q$ fraction of all data samples to respond queries) and $T$,  $\mathfrak{M}_i $ is $(\epsilon_i, \delta_i)$-differential privacy for any $\epsilon_i < c_1q^2T$, $\delta_i>0$ and $\mathbf{w}_t^i$ draw from the Gaussian distribution $\mathcal{N}(0,\sigma_i^2\mathbb{I}_p)$ where 
\begin{equation}
\label{EQ:signmaGau}
\sigma_i^2 \geq 
\frac{c_2^2\xi_2^2}{d_i^2\epsilon_i^2} \frac{bT}{N}\log{\frac{1}{\delta_i}}.
\end{equation}
\end{theorem}
The proof is straightforward by extending the Theorem 1 provided in \cite{Abadi_2016}. 
Through comparing the noise variances  of two DP mechanisms, we can observe that $\sigma_i^2$ of Gaussian increases with $\frac{bT}{N}$, which is much slower than that with Laplace. The cost lies in that the client $i$ can tolerate the failure of DP with a small probability $\delta_i$, and only $q<1$ fraction of samples can be used to participate each iteration. 

\begin{proposition}[Variance of aggregated Gaussian noises]\label{pro:noiseGau}
The  variance of the aggregated DP noises of $b$ randomly selected clients using the Gaussian mechanism in FedSGD-DPC is,
\begin{equation*}
    \mathbb{E}\left\{\left\|\mathbf{w}_{t}^b\right\|_2^2\right\}=\frac{c_2^2\eta_t^2pT\xi_2^2}{d^2}\sum_{i\in \mathcal{N}} \frac{1}{\epsilon_i^2}\log{\frac{1}{\delta_i}}.
\end{equation*}
\end{proposition}
The proof is very similar to that of Proposition~\ref{pro:noise} by substituting the variance of Gaussian noises into $\mathbf{w}_t^b$. The detailed proof is omitted. 

{\bf Remark: } Without DP, we should set $\frac{b}{N}$ and $T$ as large as possible because the model is more accurate if we query clients more exhaustively. However, if DP is involved, the noise variances increase with $\frac{b}{N}$ and $T$, which can deteriorate model accuracy. Our target is to determine  $b$ and $T$ which can maximize the final model accuracy through convergence rate analysis with DP noises.


\section{Analysis of FedSGD-DPC with  Laplace Mechanism}\label{sec:laplace}
\label{Sec:AnalysisLaplace}
We study how to tune the numbers of queries and replies, \emph{i.e.}, tuning $b$ and $T$,\footnote{Recall that the number of queries each client needs to reply is $\frac{bT}{N}$ where $N$ is a fixed number.  } so as to maximize the model accuracy (equivalent to minimizing the loss function) based on the convergence rate of FedSGD-DPC.

\subsection{Convergence Rate with Laplace Mechanism}

In FL, the distribution of data samples is non-iid (not independent and identically distributed) because the training samples in different clients can be generated from distinct distributions. According to the work~\cite{Li2020On}, the degree of non-iid can be quantified as below. 
\begin{definition}\label{def:non-iid} The degree of non-iid in FL is quantified by
\begin{equation*}
    \Gamma=F^*-\sum_{i\in\mathcal{N}}\frac{d_i}{d}F_i^*,
\end{equation*}
where $F^*$ is the minimum loss function achieved with global optimal parameters and $F_i^*$ is the minimum loss function achieved with local optimal parameters.
\end{definition}

We leverage the assumption in \cite{Li2020On} to simplify the participation scheme by assuming that the weight of each client $i$ is identical, \emph{i.e.}, $d_1=d_2= \cdots =d_{N}$. To achieve the highest convergence rate,  each client utilizes all its local samples to conduct local iterations when they adopt the Laplace mechanism, \emph{i.e.},  $\mathcal{B}_t^i= \mathcal{D}_i, \forall i\in \mathcal{N}$.\footnote{There is no constraint of $q$ with the Laplace mechanism, and thus we set $q=1$. We will restrict $q$ when analyzing the Gaussian mechanism. }

To ease our analysis, we define $\bar{\theta}_t=\sum_{i\in\mathcal{N}}\frac{d_i}{d}\theta_t^i$ and $\bar{\nu}_t=\sum_{i\in\mathcal{N}}\frac{d_i}{d}\nu_t^i$ to denote the global parameters by averaging over all clients. Similarly, we also define
$\bar{\nu}_t^b=\frac{N}{b}\sum_{i\in\mathcal{P}_t}\frac{d_i}{d}\nu_t^i$ to denote the sampled global parameters of $b$ participating clients. We use $\mathbf{g}_t=\sum_{i\in\mathcal{N}}\frac{d_i}{d}\nabla F_i(\theta_t^i, \mathcal{D}_i)$ to denote the global gradient and $\mathbf{g}_t^b=\frac{N}{b}\sum_{i\in\mathcal{P}_t}\frac{d_i}{d}\nabla F_i(\theta_t^i, \mathcal{D}_i)$ to denote the sampled global gradients. 

Let $Y_t=\mathbb{E}\left\{\left\|\bar{\theta}_t-\theta^*\right\|_2^2\right\}$ denote the expected distance between the global parameters after $t$ global iterations and the optimal global parameters.  We introduce the assumptions made in  previous works~\cite{JMLR:v19:17-650, pmlr-v80-nguyen18c,Nguyen2019TightDI,Li2020On} to simplify our analysis. These assumptions are introduced as below. 
\begin{assumption}\label{assumption:smooth}
The loss functions  are $\lambda$-smooth. Formally, there exists a constant $\lambda>0$ so that
\begin{equation*}
    F_i(\theta)\le F_i(\theta')+\left<\nabla F_i(\theta'),\theta-\theta'\right>+\frac{\lambda}{2}\left\|\theta-\theta' \right\|_2^2,
\end{equation*}
for $i=1, \dots, N$.
\end{assumption}
\begin{assumption}\label{assumption:strong_convex}
The loss functions are  $\mu$-strongly convex. Formally, there exists a constant $\mu>0$ so that 
\begin{equation*}
    F_i(\theta)\ge F_i(\theta')+\left<\nabla F_i(\theta'),\theta-\theta'\right>+\frac{\mu}{2}\left\|\theta-\theta'\right\|_2^2,
\end{equation*}
for $i=1, \dots, N$.
\end{assumption}

\begin{assumption}\label{assumption:variance of gradient}
	The expected squared norm of stochastic gradients is uniformly bounded:
	\begin{equation*}
	\mathbb{E}\left\{\left\|\nabla F_i(\theta_t, \zeta )\right\|_2^2\right\}\le G^2,
	\end{equation*}
	for $t=0, \dots, T-1$, $i=1,\dots, N$ and $\forall \zeta\in \mathcal{D}_i$ .
\end{assumption}

Based on the above assumptions, we prove the following lemmas. 
\begin{lemma}[Unbiased client sampling]\label{lemma:unbaised sample}
Sampling clients with a round robin manner is unbiased, and thus we have
	\begin{equation*}
	\mathbb{E}\left\{\bar{\nu}_t^b\right\}=\bar{\nu}_t.
	\end{equation*}
\end{lemma}

\begin{lemma}[Bounding the variance of SGD]\label{lemma:SGD variance}
Let Assumption~\ref{assumption:variance of gradient} hold, the variance of gradients from sampled clients is bounded:
\begin{equation*}
\mathbb{E}
\left\{\left\|\mathbf{g}_t^b - \mathbf{g}_t\right\|_2^2\right\}\le2\frac{N-b}{N-1}\frac{G^2}{b}.
\end{equation*}
\end{lemma}

\begin{lemma}[Upper bound in one step of FedSGD with Laplace mechanism ]\label{lemma:one step SGD}
Let Assumption~\ref{assumption:smooth} and~\ref{assumption:strong_convex} hold and assume $\eta_t\le\frac{1}{\lambda}$, we prove that
\begin{equation*}
\left\|\bar{\nu}_{t+1}-\theta^*\right\|_2^2\le(1-\mu\eta_{t})\left\|\theta_{t}-\theta^*\right\|_2^2+2\lambda\eta_{t}^2\Gamma,
\end{equation*}
 in one step of FedSGD-DPC using the Laplace mechanism
\end{lemma}

The detailed proof of Lemmas \ref{lemma:unbaised sample}- \ref{lemma:one step SGD} are presented in the Appendix~\ref{APP:Lemma1} to \ref{APP:Lemma3}. 
Based on these  lemmas, we further prove the main theorem  as follows.
\begin{theorem}\label{theorem:SGD upper buond}
Let Assumptions~\ref{assumption:smooth},~\ref{assumption:strong_convex} and~\ref{assumption:variance of gradient} hold. By setting $\eta_t=\frac{2}{\mu}\frac{1}{t+\gamma}$ where $\gamma=2\frac{\lambda}{\mu}$, the upper bound of the convergence rate of FedSGD-DPC using the Laplace mechanism is
\begin{equation*}
\begin{split}
&Y_t\le\frac{1}{t+\gamma}\left(\frac{4}{\mu^2}\omega_0+\gamma Y_0\right)+
\frac{1}{\mu^2}\frac{1}{t+\gamma}\omega_1 ,\\
&\omega_0=2\frac{N-b}{N-1}\frac{G^2}{b}+2\lambda\Gamma,\\
&\omega_1 = \frac{32pbT^2\xi_1^2}{Nd^2}\sum_{i\in \mathcal{N}}\frac{1}{\epsilon_i^2}.
\end{split}
\end{equation*}
\end{theorem}
Theorem~\ref{theorem:SGD upper buond} is proved in Appendix \ref{APP:Theorem3} by leveraging Lemmas~\ref{lemma:unbaised sample}-\ref{lemma:one step SGD}. 

{\bf Remark I:}
It is worth to discussing the difference between $Y_t$ and $Y_T$. $Y_t$ represents the distance to $\theta^*$ after $t$ iterations, while $Y_T$ is the final gap after $T$ total iterations. From Theorem~\ref{theorem:SGD upper buond}, the bound of $Y_t$ is a monotonic decreasing series with $t$.  However, $Y_T$ is not a monotonic decreasing function with $T$. 
According to Theorem~\ref{theorem:SGD upper buond},  by taking all other variables as constants, $Y_T$ diverges with the rate $O(\frac{1}{T}+T)$ asymptotically  as $T$ approaches infinity due to the influence of DP noises. 

{\bf Remark II:}
We cannot set $T$ and $b$ arbitrarily large  to increase the number of query times of  each client, \emph{i.e.}, $\frac{bT}{N}$. As $T$ approaches infinity, the term $\omega_1$ approaches infinity as well so that $Y_T$ finally diverge with the rate $O(\frac{1}{T}+T) $ asymptotically. 

{\bf Remark III:}
According to Theorem~\ref{theorem:SGD upper buond}, there exist optimal $b$ and $T$ such that $Y_T$ can be minimized over  $T\geq 0$ and $1\leq b\leq N$. 

In the rest analysis, let $b^*$ and $T^*$ denote the optimal values of $b$ and $T$ respectively.  

\subsection{Optimizing $T$ and $b$ with Laplace Mechanism}

By leveraging the bound of convenience rate, we define $U(T, b) =Y_T$ as the function with variables $T$ and $b$. It is complicated to jointly optimize $T$ and $b$. Thus, we simplify the analysis by fixing either $b$ or $T$.  

If $b$ is fixed, $U(T, b)$ is simplified as $U(T)$, which can be expressed as
\begin{equation*}
U(T)=\frac{A_1+A_2T^2+\gamma Y_0}{T+\gamma},
\end{equation*}
where $A_1=\frac{4}{\mu^2}\left(2\frac{N-b}{N-1}\frac{1}{b}G^2+2\lambda\Gamma\right)$ and $A_2=\frac{32}{\mu^2}\frac{pb\xi_1^2}{Nd^2}\sum_{i\in\mathcal{N}}\frac{1}{\epsilon_i^2}$ are regarded as constant numbers.
It is easy to verify that $U(T)$ is a convex function with $T$. By letting $\frac{\partial U}{\partial T}=0$, we obtain 
\begin{equation}\label{equ:optimal T}
T^*=\sqrt{\gamma^2+\frac{A_1+\gamma Y_0}{A_2}}-\gamma. 
\end{equation}
This result can be explained intuitively. 
\begin{itemize}
    \item $A_1$ is related with the sum of gradient variance and the degree of non-iid. $A_2$ is determined by the variance of the DP noises.
    \item If the privacy budget $\epsilon_i$ is smaller implying higher privacy protection level and larger variance of the DP noises, $T^*$ should be smaller. In the extreme case when $\epsilon_i, \forall i$ approaches $0$, $T^*$ approaches $0$. It implies  that DP noises  make the collaborative learning between clients useless. 
    \item If $A_1$ is larger, $T^*$ should be set as a larger value accordingly because increasing iteration times $T$ is the way to reduce the variance of gradients and the non-iid degree in convergence rate.  In contrast, if $A_2$ is larger, $T^*$ should be reduced because the variance of DP noises is inflated by $T$.
\end{itemize}


Similarly, we fix $T$  to analyze the optimal the value of $b$. The upper bound function can be rewritten  as 
\begin{equation*}
U(b)=\frac{1}{b}B_1+bB_2+B_3,
\end{equation*}
where $B_1=\frac{4}{\mu^2}\frac{G^2}{T+\gamma}\frac{2N}{N-1}$, $B_2=\frac{1}{\mu^2}\frac{1}{T+\gamma} \frac{32pT^2\xi_1^2}{Nd^2}\sum_{i\in \mathcal{N}}\frac{1}{\epsilon_i^2}$ and $B_3=\frac{\gamma Y_0}{T+\gamma}+\frac{4}{\mu^2}\frac{1}{T+\gamma}\left(2\lambda\Gamma-\frac{2G^2}{N-1}\right)$ are regarded as constant numbers.  Interestingly, this is a convex function again. 
By letting $\frac{\partial U}{\partial b}=0$, we obtain
\begin{equation*}
b^*=\sqrt{\frac{B_1}{B_2}}=\frac{GNd}{2T\xi_1}\frac{1}{\sqrt{p(N-1)}}\frac{1}{\sqrt{\sum_{i\in \mathcal{N}}\frac{1}{\epsilon_i^2}}}.
\end{equation*}
The expression of $b^*$ can be interpreted as below. 
\begin{itemize}
    \item $b^*$ is not affected by the level of non-iid degree, \emph{i.e.}, $\Gamma$, because $\Gamma$ is only included in $B_3$ and involving more clients into $\mathcal{P}_t$ cannot reduce the level of non-iid.
    \item If the privacy budget is smaller, $b^*$ should be smaller to alleviate the influence of DP noises. 
    According to  Proposition~\ref{pro:noise}, the  variance of aggregated DP noises is smaller if $b$ is smaller, \emph{i.e.}, fewer  clients are involved into $\mathcal{P}_t$.
    This result is counter-intuitive because the aggregated variance averaged over more independent clients should be smaller. However, increasing $b$ will enlarge the query times because each client needs to  reply $\frac{bT}{N}$ times, which will inflate the noise variance. The overall effect is that the aggregated noise variance increases with $b$. 
    \item $b^*$ should be larger if the gradient variance, \emph{i.e.}, $G$, is larger because increasing the size of $\mathcal{P}_t$ can effectively reduce the influence of gradient variance. 
\end{itemize}

We now minimize $U(T, b)$ by jointly analyzing $T$ and $b$. 
\begin{proposition}
$U(T,b)$ is strictly  biconvex with respect to $T$ and $b$. 
\end{proposition}
The proof is straightforward since $U(T)$ and $U(b)$ are strictly convex with respect to $T$ and $b$ respectively. 

Although a strictly biconvex problem can be solved efficiently  with existing algorithms such as the Alternate Convex Search (ACS) approach \cite{gorski2007biconvex}, it is worth to exploring the closed-form solutions of $T^*$ and $b^*$ to not only reveal the implications of the optimal solution but also gain a faster algorithm to achieve $T^*$ and $b^*$.

We formally define the optimization problem of $U(T,b)$ as
\begin{equation*}
    \begin{split}
        &\text{minimize }\quad U(T,b)=\frac{1}{T+\gamma}\left(\frac{C_1}{b}+C_2bT^2+C_3\right),\\
        &\text{s.t. }\quad 0\le T,1\le b\le N,\\
    \end{split}
\end{equation*}
where $C_1=\frac{4}{\mu^2}\frac{2NG^2}{N-1}$, $C_2=\frac{1}{\mu^2}\frac{32p\xi_1^2}{Nd^2}\sum_{i\in \mathcal{N}}\frac{1}{\epsilon_i^2}$ and  $C_3=\gamma Y_0+\frac{4}{\mu^2}\left(2\lambda\Gamma-\frac{2G^2}{N-1}\right)$ are constant numbers.

All possible optimal solutions of $T$ and $b$ can be analyzed through  K.K.T. conditions.  The  Lagrangian function of $U(T, b)$ can be defined as
\begin{equation*}
\mathcal{L}(T,b,\rho_1,\rho_2,\rho_3)=U(T,b)-\rho_1T+\rho_2(1-b)+\rho_3(b-N),
\end{equation*}
where $\rho_1$, $\rho_2$ and $\rho_3$ are Lagrangian multipliers. 
The K.K.T. conditions are listed in Appendix~\ref{sec:K.K.T.}, which give rise to the following three feasible  solutions.
\begin{itemize}
    
    \item \textbf{Solution 1 \& 2}: $T^*=\sqrt{\gamma^2+\left(\frac{C_1}{b}+C_3\right)\frac{1}{C_2b}}-\gamma$, $b^*=1\text{ or } b^*= N$.\footnote{According to our experiments results, solution 1 with $b^*=1$ is the optimal solution in most cases.}
     It is interesting to observe that $b^*$ should be assigned with boundary values, \emph{i.e.}, either $1$ or $N$. This result can be intuitively explained as follows. If the gradient variance term, \emph{i.e.}, $C_1$, dominates, $b^*$ should be assigned with $N$ to minimize the influence of gradient variance. Otherwise, if the variance of DP noises  dominates, we should have $b^*=1$. The implication is that the FedSGD-DPC algorithm with $b=1$ minimizes the number of times each client replies, \emph{i.e.}, $\frac{bT}{N}$, so as to minimize the variance of DP noises. 
    \item \textbf{Solution 3}: $T^*=0$.
    This solution means that there is no need to conduct any iteration. This is possible if the privacy budgets of clients are too small resulting in over huge variances of DP noises so that collaborative learning is useless. 
\end{itemize}
The best solution can be found by comparing three possible solutions as long as these parameters can be obtained in these solutions. 
In fact, only the solution 1 with $b^*=1$ is the feasible solution in practice to maximize the final model accuracy. This point will be clear when experiment results are presented. 


{\bf Discussion of Implementation:}  According to the above discussion,  it is necessary to obtain the values of a number of parameters  such as $G$, $\Gamma$ and $\gamma$ before we can derive $T^*$ and $b^*$.  
In practice, $\epsilon_i$ and $\delta_i$ are determined by the clients' privacy budgets, which can be reported by clients. We can also suppose that $N$ and $d$ are known by the PS before FedSGD-DPC is executed. 
For the rest parameters including $\mu$, $\lambda$, $\xi_1$ and $G$,  they are problem-related parameters. For most of them, their values can be estimated  without incurring much overhead with the assistance of clients by leveraging the existing algorithms introduced in \cite{luo2020cost,wang2019adaptive}, except the parameters $Y_0$ and $\Gamma$.  We propose the following method to approximately estimate $Y_0$ and $\Gamma$. 
\begin{itemize}
    \item Estimation of $Y_0$: Each client can easily deduce the optimal local parameters $\theta_i^{*}$. Given initial $\theta_0$, each client returns the gap $\|\theta_0-\theta_i^{*}\|_2^2$ to the PS. The PS approximately estimate $Y_0$ as $\sum_{i\in \mathcal{N}}\frac{d_i}{d} \|\theta_0-\theta_i^{*}\|_2^2$.
    \item Estimation of $\Gamma$: With the local optimal parameter $\theta_i^{*}$,  client $i$  returns $F_i^*$ to the PS. Then, $\Gamma$ is approximately estimated as $   \max_{\forall i \in \mathcal{N}}F_i^*-\sum_{i\in\mathcal{N}}\frac{d_i}{d}F_i^*,$  
\end{itemize}`

\section{Analysis of FedSGD-DPC with Gaussian Mechanism}\label{sec:gaussian}

\subsection{Convergence Rate with Gaussian Mechanism}
\label{Sec:AnalysisGaussian}

We can largely reuse the method in the last section to analyze the convergence rate of FedSGD-DPC with the Gaussian mechanism. 

We highlight two critical differences between the convergence rate with the Gaussian mechanism and that with the Laplace mechanism here. 
\begin{enumerate}
\item Recall that the variance of the aggregated Gaussian noises derived in Proposition~\ref{pro:noiseGau} is independent with $b$, and  its growth rate with $T$ is much slower than that with Laplace noises.
\item For each global iteration, only $q$ fraction of local samples in each client can be used to conduct the local iteration, where $q$ could be much less than $1$. 
\end{enumerate}
Intuitively, 1) can improve the model accuracy with a smaller noise variance, whereas 2) can lower the model accuracy because fewer samples are used for local iterations. We need to derive the convergence rate to quantify the overall effect, and derive the optimal $T$ and $b$. 

\begin{assumption}
\label{ASP:VarianceBound}
Let $\zeta$ denote any sample selected from a client's dataset. The variance of stochastic gradients in each client is bounded: 
 $\mathbb{E}\|\nabla F_i(\theta_t^i, \mathcal{D}_i)- \nabla F_i(\theta_t^i, \zeta)\|_2^2\leq \Lambda_i^2$, $\forall \zeta\in \mathcal{D}_i$. 
\end{assumption}

Since not all local samples are used to conduct the local iteration, we need Assumption~\ref{ASP:VarianceBound} to bound the variance of stochastic gradients in each client. 
We further define $\mathbf{g}_t^{b,q} =\frac{N}{b} \sum_{i\in\mathcal{N}}\frac{d_i}{d}\nabla F_i(\theta_t, \mathcal{B}_t^i)$ where $\mathcal{B}_t^i$ is a randomly selected subset of $\mathcal{D}_i$ for iteration $t$ and $\frac{|\mathcal{B}_t^i|}{|\mathcal{D}_i|}=q$.  Let $\mathbf{g}_t^q = \sum_{i\in\mathcal{N}}\frac{d_i}{d}\nabla F_i(\theta_t, \mathcal{B}_t^i)$ to represent the global gradients achieved with randomly selected samples.

\begin{lemma}[Bounding the variance of SGD]\label{lemma:SGD variance q}
Let Assumptions~\ref{assumption:variance of gradient} and \ref{ASP:VarianceBound} hold, the variance of gradients of sampled clients with randomly selected samples is bounded:
\begin{equation*}
\mathbb{E}
\left\{\left\|\mathbf{g}_t^{b, q} - \mathbf{g}_t\right\|_2^2\right\}\le2\frac{N-b}{N-1}\frac{G^2}{b}+\sum_{i\in \mathcal{N}} \frac{d_i}{qd^2} \Lambda_i^2.
\end{equation*}
\end{lemma}
The proof is presented in Appendix~\ref{APP:Lemma4}. By leveraging Lemmas \ref{lemma:unbaised sample}, \ref{lemma:one step SGD} and \ref{lemma:SGD variance q}, we prove the main theorem as below.

\begin{theorem}\label{theorem:SGD upper buond Gau}
Let Assumptions~\ref{assumption:smooth},~\ref{assumption:strong_convex} and~\ref{assumption:variance of gradient} hold. By setting $\eta_t=\frac{2}{\mu}\frac{1}{t+\gamma}$ where $\gamma=2\frac{\lambda}{\mu}$, the upper bound of the convergence rate of FedSGD-DPC using the Laplace mechanism is
\begin{equation*}
\begin{split}
&Y_t\le\frac{1}{t+\gamma}\left(\frac{4}{\mu^2}\omega'_0+\gamma Y_0\right)+
\frac{1}{\mu^2}\frac{1}{t+\gamma}\omega'_1 ,\\
&\omega_0'=2\frac{N-b}{N-1}\frac{G^2}{b}+\sum_{i\in \mathcal{N}} \frac{d_i}{qd^2} \Lambda_i^2 +2\lambda\Gamma,\\
&\omega_1' = \frac{4c_2^2pT\xi_2^2}{d^2}\sum_{i\in \mathcal{N}} \frac{1}{\epsilon_i^2}\log{\frac{1}{\delta_i}}.
\end{split}
\end{equation*}
\end{theorem}
Theorem~\ref{theorem:SGD upper buond Gau} can be proved in a similar way as that of Theorem~\ref{theorem:SGD upper buond}, and the detailed proof is presented in Appendix \ref{APP:Theorem4}.
Through comparing Theorem~\ref{theorem:SGD upper buond} and Theorem~\ref{theorem:SGD upper buond Gau}, we can observe
\begin{itemize}
    \item Asymptotically, $Y_T$ using the Gaussian mechanism has a better convergence rate since the term $\frac{1}{T+\gamma}\omega_1'$ approaches $O(1)$ as $T$ approaches infinity. In comparison. $Y_T$ using the Laplace mechanism is $O(\frac{1}{T})+O(T)$ which goes to infinity as $T$ increases. 
    \item The drawback of FedSGD-DPC with the Gaussian mechanism lies in enlarged $\omega_0'$ due to the term $\sum_{i\in \mathcal{N}} \frac{d_i}{qd^2} \Lambda_i^2$.  If $q$ is smaller, this term is larger, and it implies that it takes more iterations to reduce $\omega_0'$ to a certain small value.
\end{itemize}

\subsection{Optimizing $T$ and $b$ with Gaussian Mechanism}

Again, we define $U(T,b) =Y_T$ for this case. It is easy to minimize $U(T,b)$ by rewriting $U(T,b)$ as below.
\begin{equation*}
    \begin{split}
        &\text{minimize }\quad U(T,b)=\frac{E_1}{b(T+\gamma)}+\frac{E_2}{T+\gamma}+E_3,\\
        &\text{s.t. }\quad 0\le T,1\le b\le N,\\
        & E_1 =\frac{8G^2}{\mu_2}\frac{N}{N-1},\\
        & E_2 = \frac{4}{\mu^2}\left(\frac{-2G^2}{N-1}+\sum_{i\in \mathcal{N}} \frac{d_i}{qd^2} \Lambda_i^2 +2\lambda\Gamma\right)+\gamma Y_0\\
        & \quad -\frac{4}{\mu^2}\left( \frac{\gamma c_2^2p\xi_2^2}{d^2}\sum_{i\in \mathcal{N}} \frac{1}{\epsilon_i^2}\log{\frac{1}{\delta_i}}\right),\\
        & E_3 = \frac{4}{\mu^2}\left( \frac{ c_2^2p\xi_2^2}{d^2}\sum_{i\in \mathcal{N}} \frac{1}{\epsilon_i^2}\log{\frac{1}{\delta_i}}\right).
    \end{split}
\end{equation*}
It is not difficult to verify that $U(T, b) $ is strictly biconvex, and can be solved efficiently by the method introduced in the last section. There are two feasible solutions, which are discussed as below. 
\begin{itemize}
\item  \textbf{Solution 1}: $T^*=\infty$ and $b^*=N$. This solution is feasible when the privacy budget is not very small and $Y_0$ is large enough. In this case, the variance of DP noises is not so significant such that the collaborative learning between clients should be executed for as many iterations as possible. The size of $\mathcal{P}_t$ should be as large as possible, and thus we should set $b=N$. $Y_T$ will finally converge to the constant number $E_3$.\footnote{In fact, for any arbitrary $b$, $Y_T$ will approach  $E_3$. The convergence rate will be fastest by setting $b=N$. } 

Note that it may be difficult to maintain the Gaussian mechanism by setting  $T^*=\infty$. It is also impracticable to set  $T^*=\infty$ in a real system. The implication of our analysis is that we should set $T$ as large as possible for FL with the Gaussian mechanism. 

\item \textbf{Solution 2}: $T^* = 0$. This solution is feasible when the privacy budget is too small and $Y_0$ is not very large. It implies that the initial model parameters are already good enough because the variance of  DP noises is too large such that any further learning is useless. 
\end{itemize}

{\bf Remark I:} Using both the Laplace mechanism and the Gaussian mechanism, we can see that the solution $T^*=0$ exists. In theory, if $\epsilon_i$ and $\delta_i$ are infinitesimal small, the noise variance can be arbitrarily large. If the variance of DP noises is too huge, it will totally distort all useful parameter information, and as a result FL is not necessary any more. However, FL does not make sense if the privacy budget is too tight, and thus $T^*=0$ should be very rare in practice. 

{\bf Remark II:} Although our study only includes two specific DP mechanisms, our framework is applicable if a different DP mechanism is adopted through updating the terms $\omega_1$ and $\omega'_1$  in the convergence rates.

\section{Experiment}
\label{Sec:Exp}
In this section, we conduct experiments to evaluate the correctness and effectiveness of our proposed solutions for the optimal numbers of queries and replies in FL with DP. 

\subsection{Experimental Settings}
\subsubsection{Models and Datasets}

We evaluate our analysis results through conducting experiments with MNIST and FMNIST datasets \cite{caldas2018leaf}. MNIST contains 70,000 handwritten digit images. Each image is with 28$\times$28 gray-scale pixels. We randomly select 60,000 for training and the rest 10,000 images are used for testing. FEMNIST is a superset of MNIST with 62 different classes and widely used in evaluating FL system. We train convex multinomial logistic regression (LR) model with 7,840 parameters with the MNIST dataset. In the LR model, the inputs and the outputs are related in $y=\theta^T x$, where $\theta^T$ represents a 2-D array with dimension $784\times10$. A CNN model with 110,526 parameters is trained to classify FEMNIST dataset.  In the CNN model, we adopt 2 convolution layers with 416 and 12,832 parameters respectively. Each layer is followed by a ReLu activation function and a $2\times2$ max pooling layer. One fully connected layer with 97,278 ($1,569\times62$) parameters is adopted for the output layer.

\subsubsection{Data Distribution}
Data samples are distributed on clients in a non-iid manner. For the MNIST dataset, we divide the training set into $N$ groups with equal size, but images allocated to each client are selected out of only two digits. According to \cite{caldas2018leaf}, the FEMNIST dataset is inherently non-iid generated by 3,500 clients. We use the method introduced in \cite{caldas2018leaf} to generate 70,037 samples, which are distributed on exact $N$ clients.

\begin{figure*}[ht]
\centering
\subfigure[Laplace Mechanism]{
    \label{fig:iter_l}	
    \includegraphics[width=6.0cm]{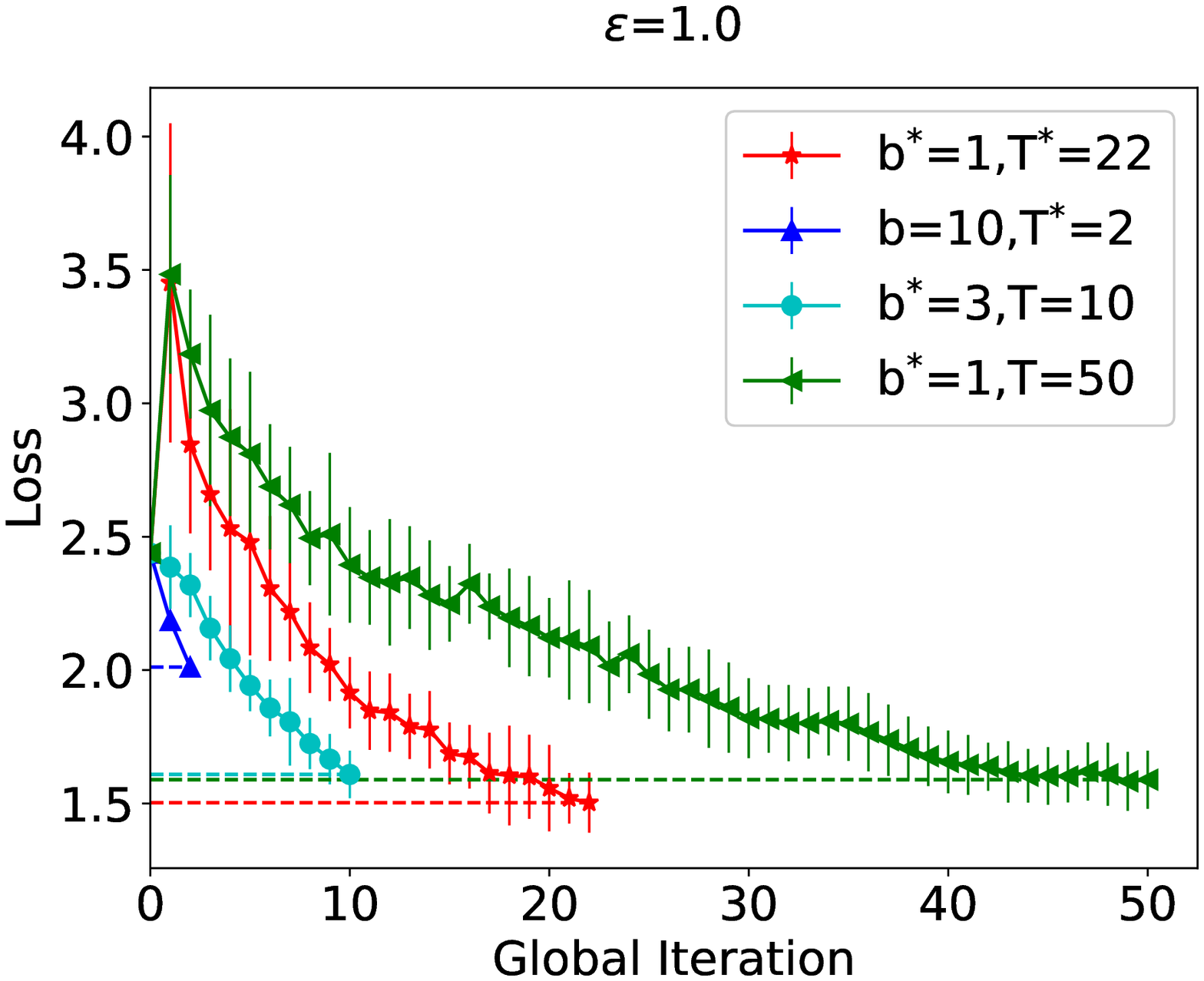}
    \includegraphics[width=6.0cm]{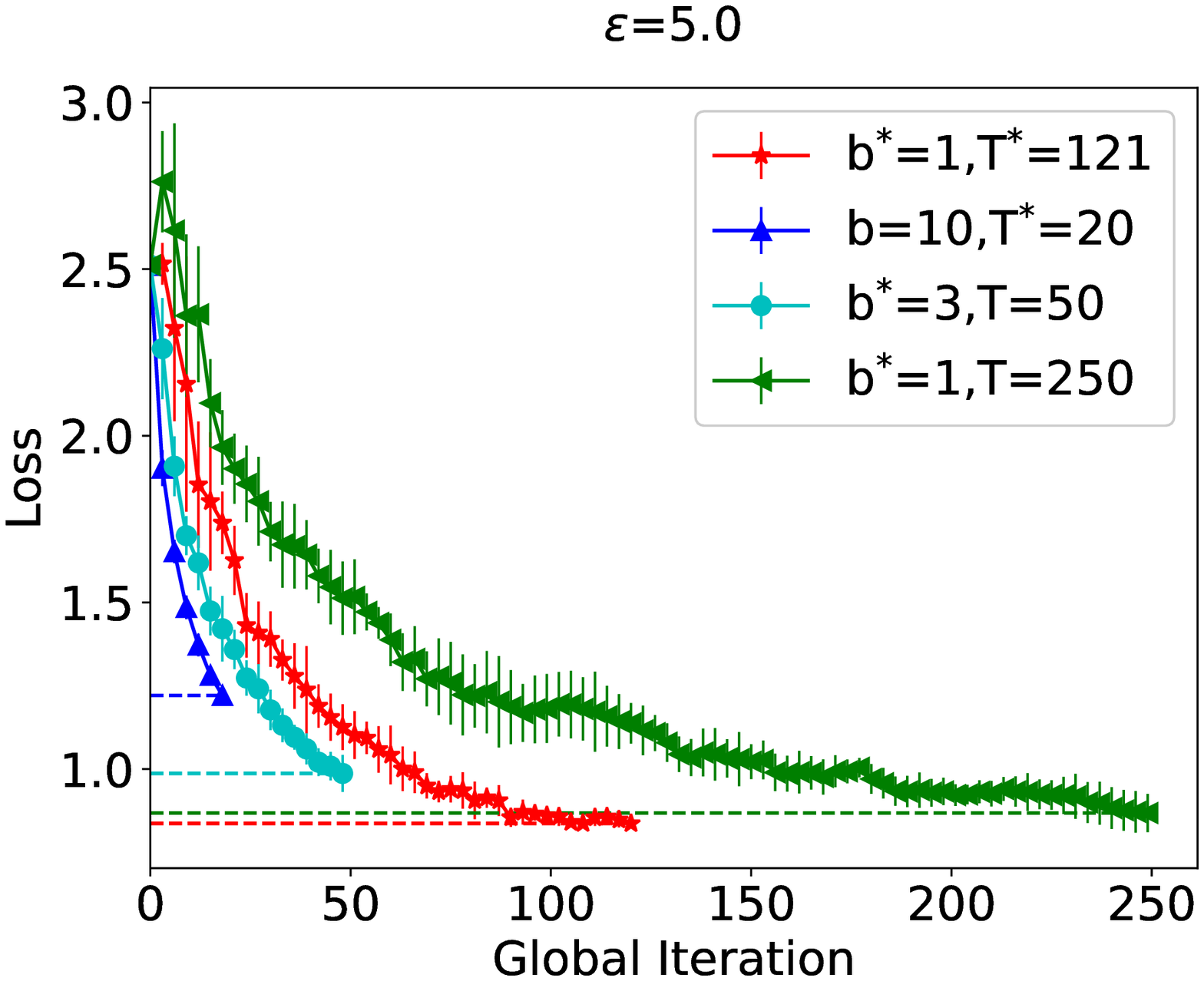}
    \includegraphics[width=6.0cm]{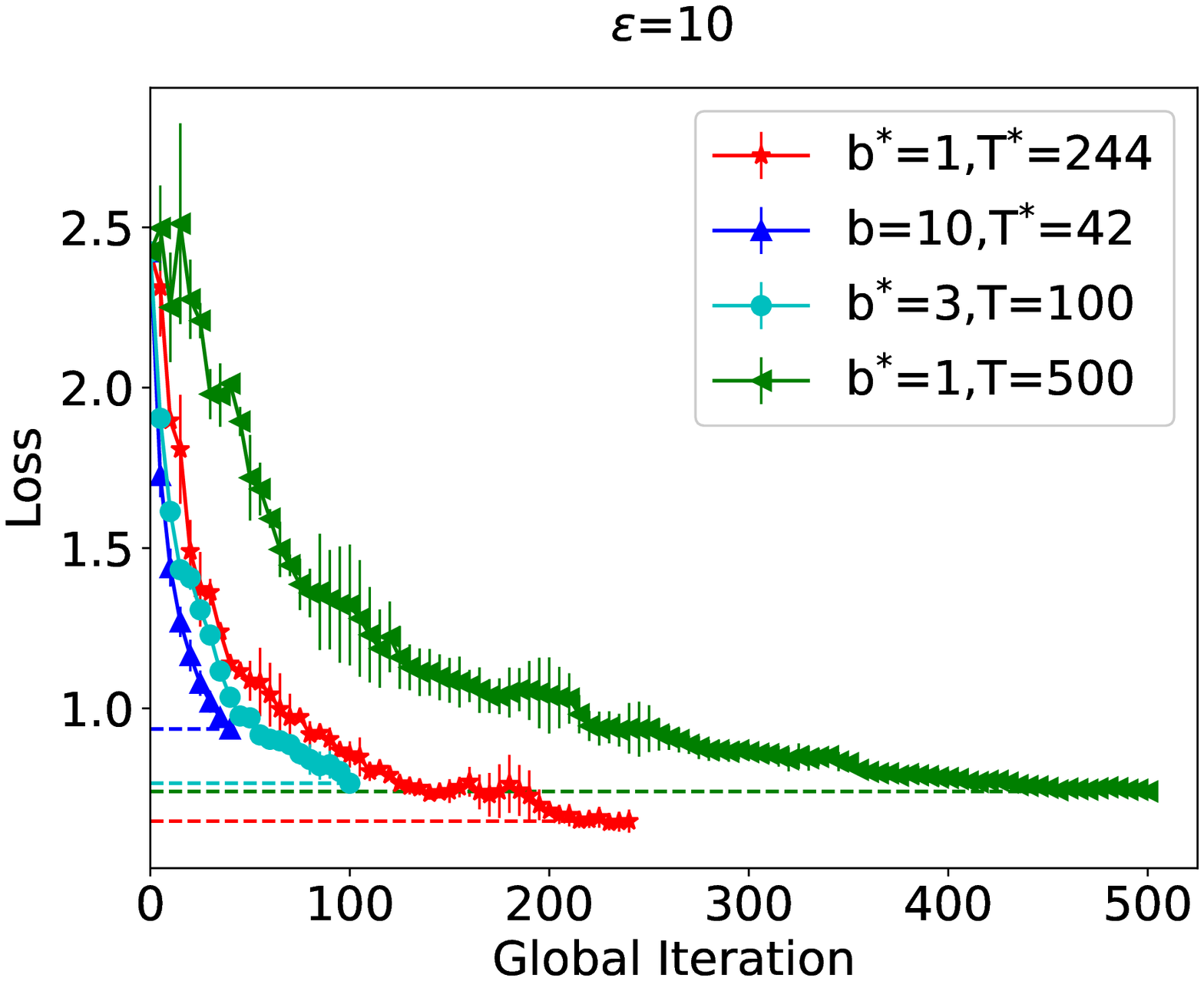}
}
\subfigure[Gaussian Mechanism]{
    \label{fig:iter_g}	
    \includegraphics[width=6.0cm]{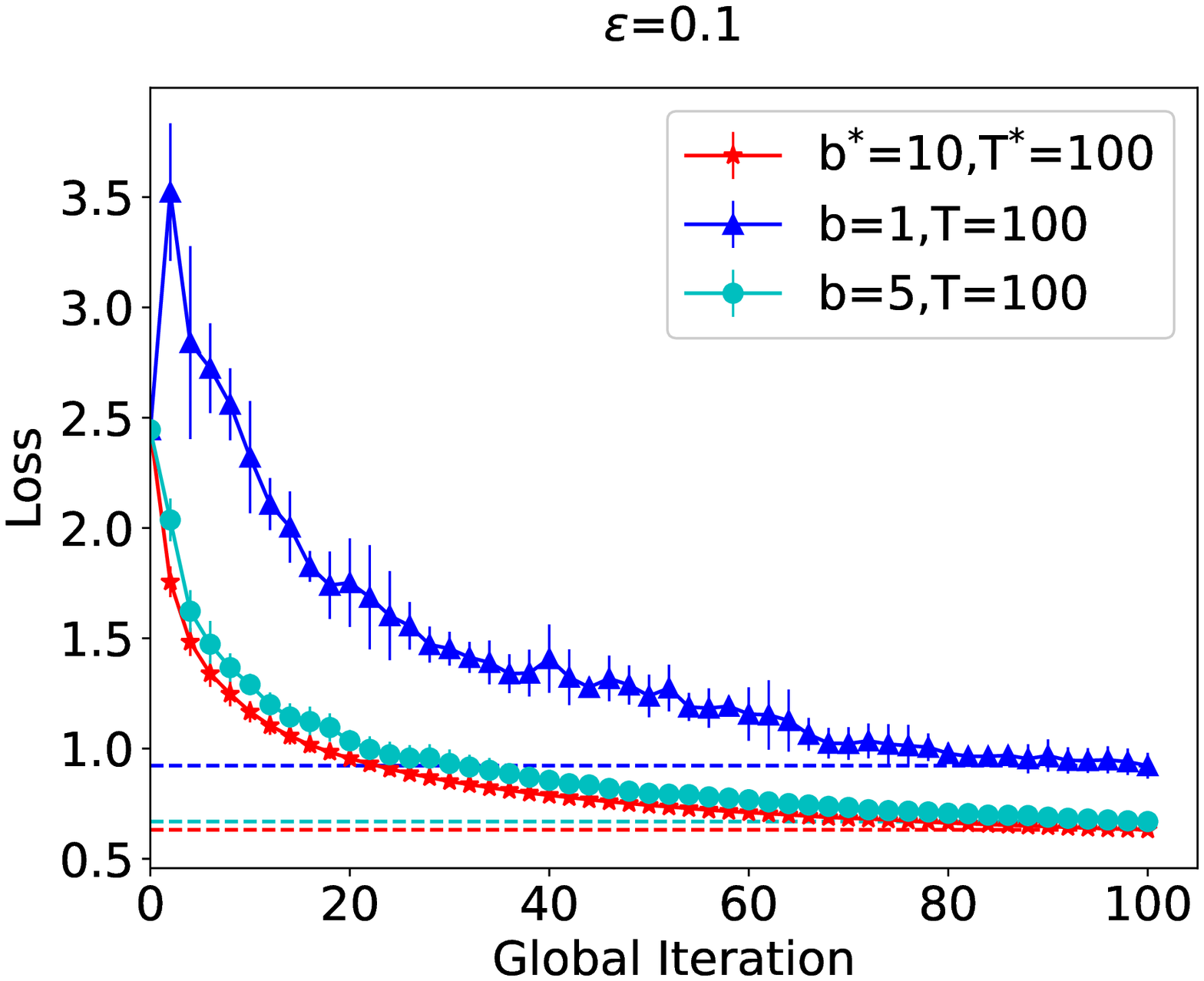}
    \includegraphics[width=6.0cm]{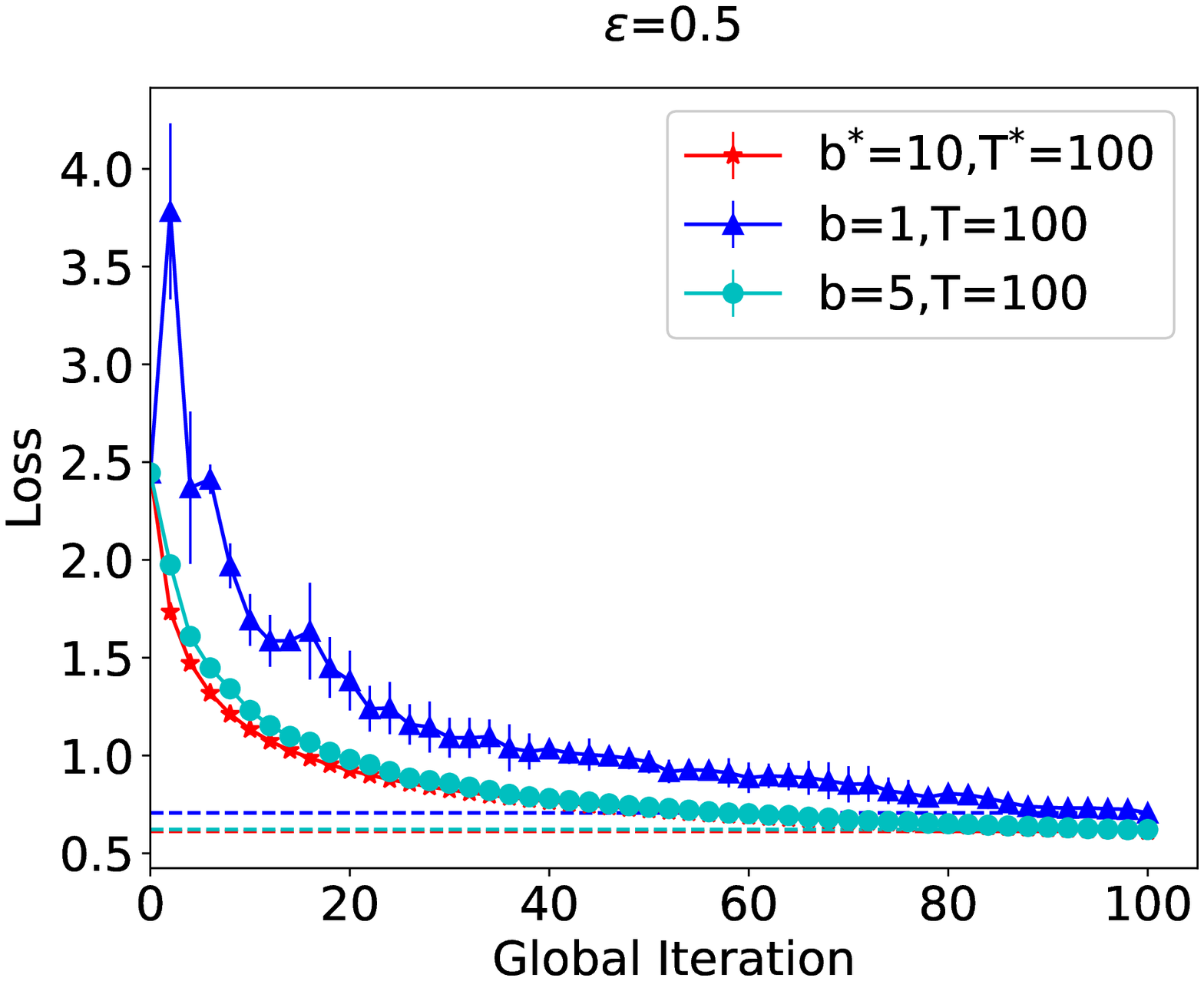}
    \includegraphics[width=6.0cm]{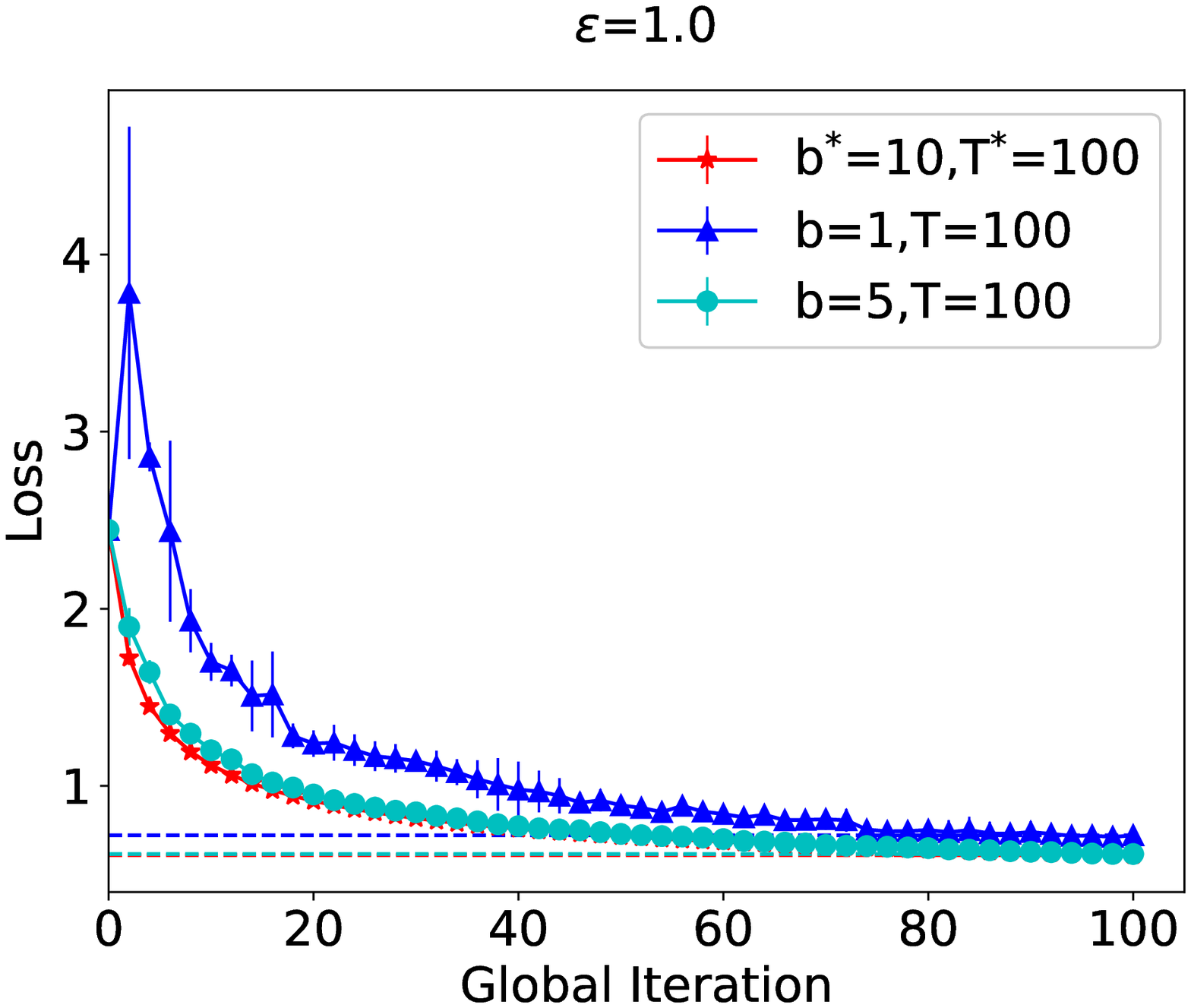}
}
\caption{Comparison of loss functions per global iteration for MNIST by setting different values for $b$ and $T$ in FL with DP.  There are $N=10$ clients, $b^*$ and $T^*$ are determined by our solutions, which give $b^*=1$ for the Laplace mechanism and $ b^*=10$  for  the Gaussian mechanism. 
}
\label{fig:loss_mn}	
\end{figure*}

\subsubsection{Experimental Settings}

In our experiments, we set the number of total clients as $N=10$ in FL.  The learning rate to train both models is $\eta$ = 0.05 initially, and decays gradually with the number of iterations. 
By default, the number of participating clients in each global iteration, \emph{i.e.}, $b$, is set as 1, 5 or 10, and the total number of iterations $T$ is selected from [10, 50, 100, 150, 200, 250, 500] (except for $T^*$) for the Laplace mechanism. $T$ is set as $100$ for the Gaussian mechanism.  Each case in our experiments is repeated for 10 times, and we plot the average performance together with error bars in our results. 

For the Laplace mechanism, the clipping bound of gradients is $\xi_1 = 300$ (\emph{i.e.}, \emph{l}$_1$-norm of gradients) for MNIST and $200$  for FEMNIST respectively. For the Gaussian mechanism, the clipping bound is 10 (\emph{l}$_2$-norm of gradients) for both datasets. The privacy budget of each client is set as follows:  $\epsilon$ is selected from the set [1, 5, 10] (representing strong, moderate and weak privacy protection needs) for the Laplace mechanism, while $\epsilon$ is selected from [0.1, 0.5, 1.0] and $\delta$ is fixed at  $10^{-5}$ for the Gaussian mechanism according to  \cite{wu2019value, Abadi_2016}. The sampling probability $q$ of the Gaussian mechanism  is $q=0.01$. 

\subsubsection{Baselines}

The main purpose of our experiments is to verify that setting FedSGD-DPC with $b^*$ and $T^*$  obtained with our analysis can achieve the highest model accuracy. 
For comparison,  we also implement FedSGD-DPC by setting different values for $b$ and $T$ in our experiments. 
\begin{itemize}
    \item It is the algorithm proposed in \cite{wu2019value} by setting a fixed $T$ and $b$ in FedSGD-DPC with the Laplace mechanism.
    \item It is the algorithm proposed in \cite{Abadi_2016} by setting a fixed $T$ and $b$ in FedSGD-DPC with the Gaussian mechanism.
\end{itemize}
Our evaluation can demonstrate the performance improvement achieved by optimizing $b$ and $T$ in our work. 

The FedSGD-DPC with the optimal solution derived by our analysis is denoted by $b^*$ and $T^*$. For baselines, we set $b$ and $T$ differing from $b^*$ and $T^*$.
For example, $b=1$, $T=100$  denotes that FedSGD-DPC is implemented with $b=1$ and $T=100$. Its performance should be different from that with $b^*$ and $T^*$.

\subsubsection{Evaluation Metrics}
In the experiments, we use two metrics to evaluate the performance of machine learning models. The first one is the cross entropy loss, which is used to measure the difference between predicted probability distribution and true probability distribution. The second one is the model accuracy which is defined as the proportion of correctly identified samples in the test set.

\begin{figure*}[ht]
\centering
\subfigure[Laplace Mechanism]{
    \includegraphics[width=6.0cm]{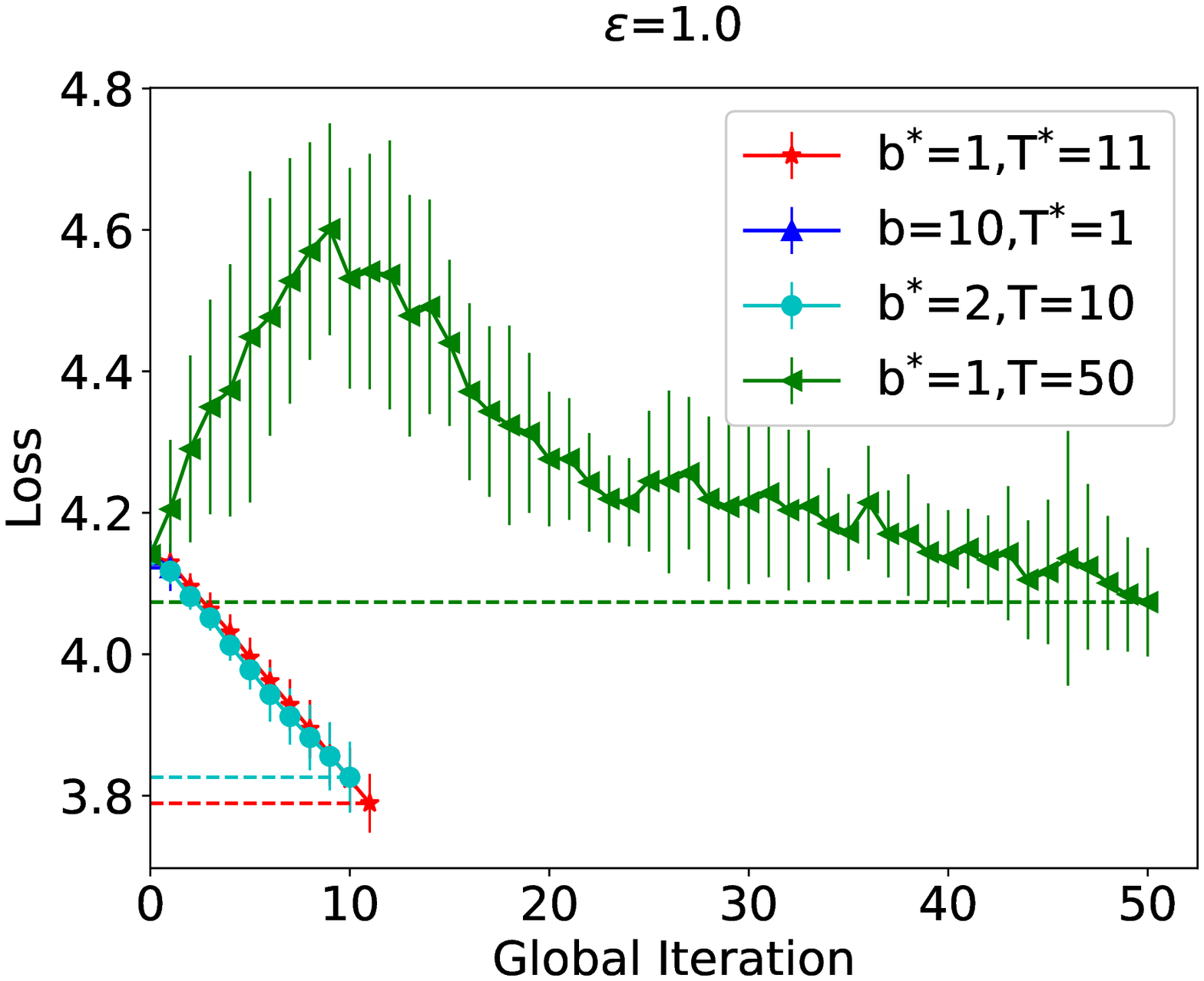}
    \includegraphics[width=6.0cm]{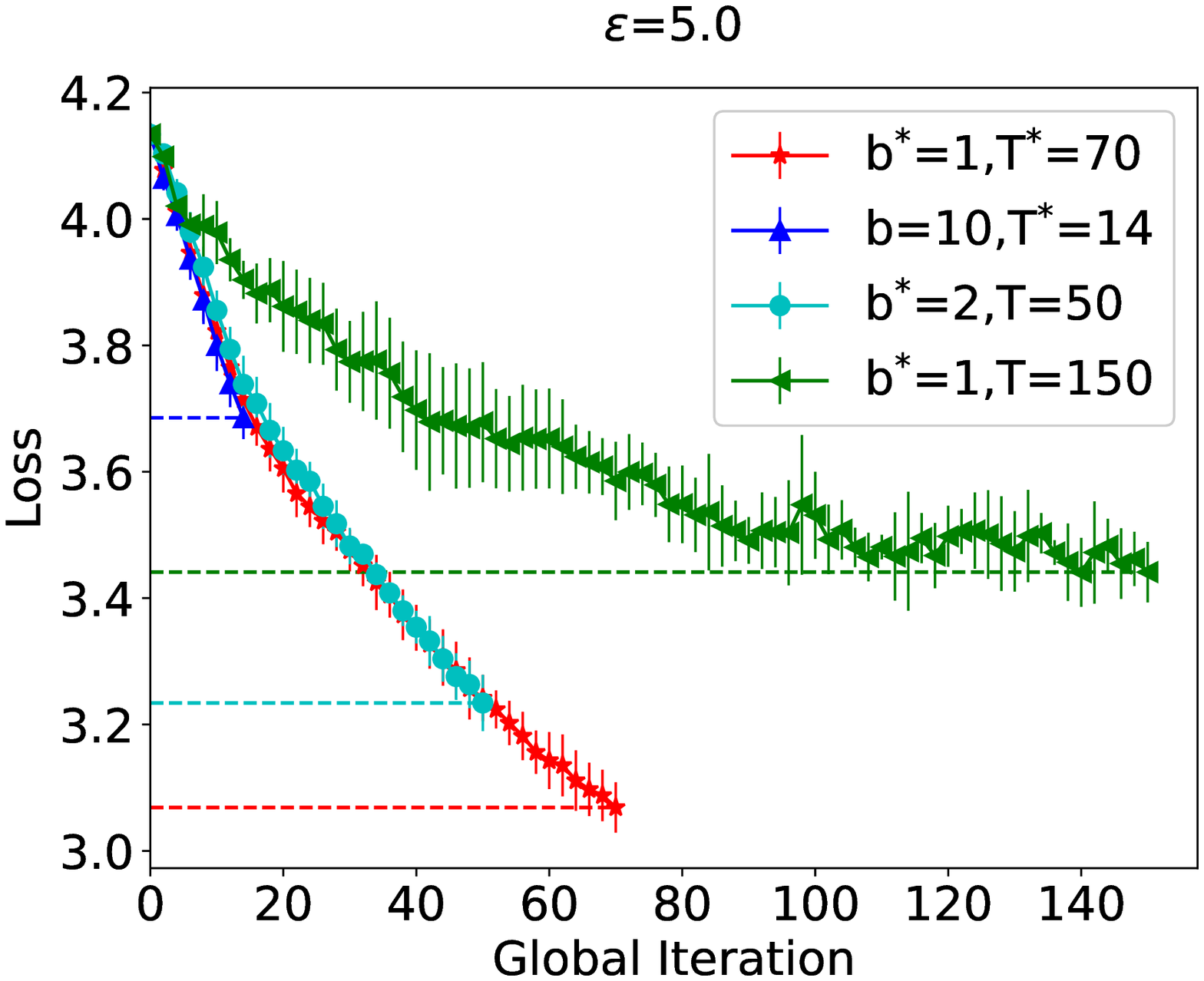}
    \includegraphics[width=6.0cm]{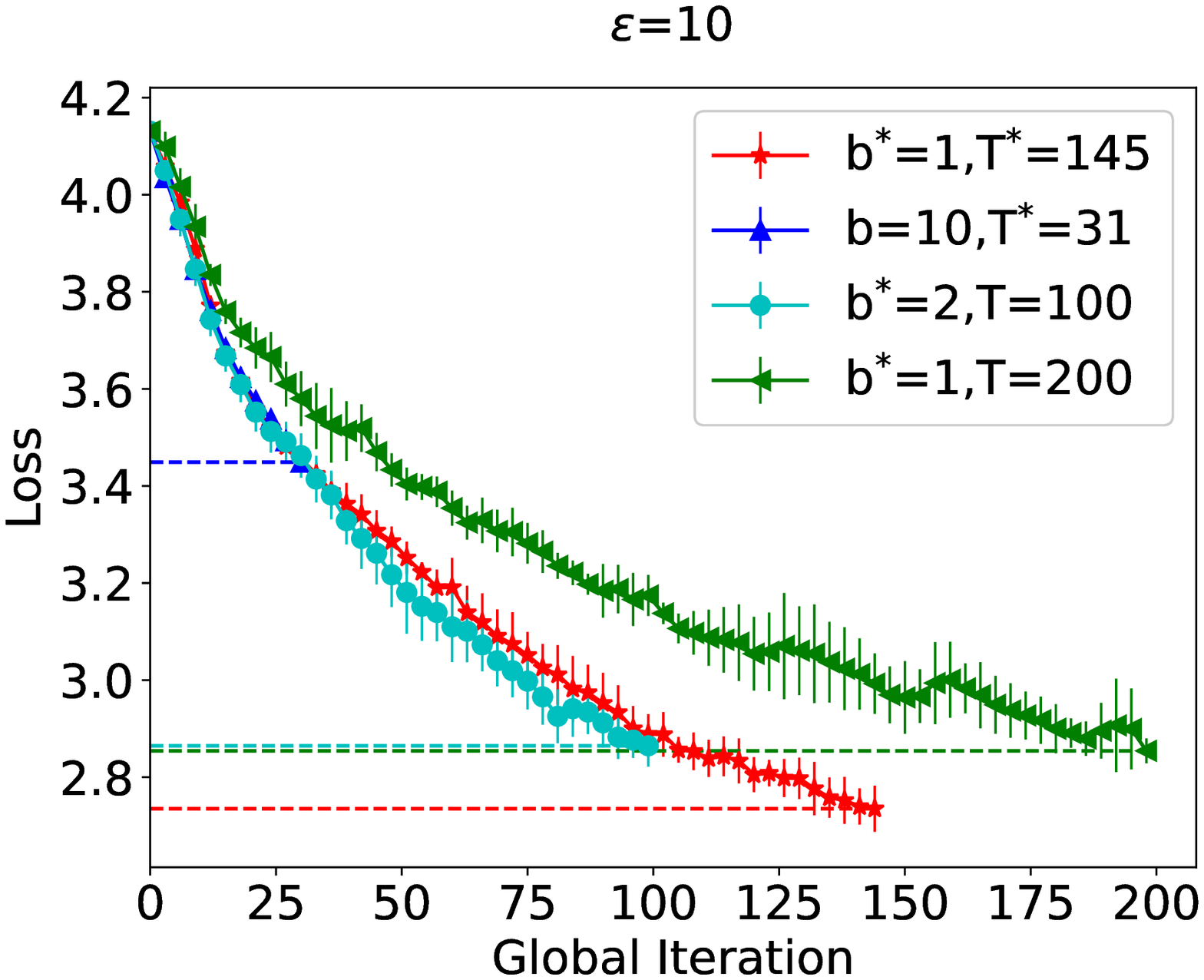}
}
\subfigure[Gaussian Mechanism]{
    \includegraphics[width=6.0cm]{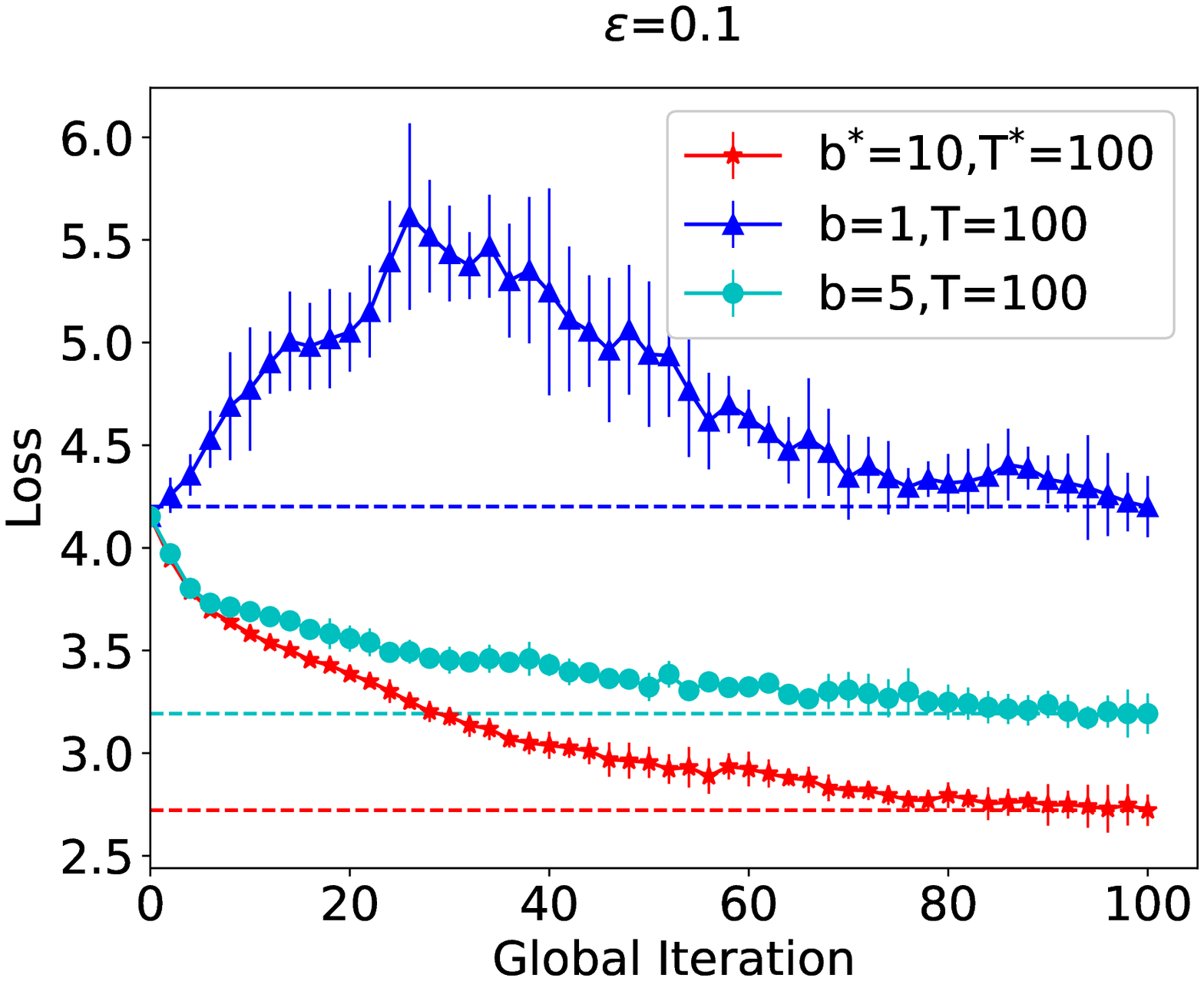}
    \includegraphics[width=6.0cm]{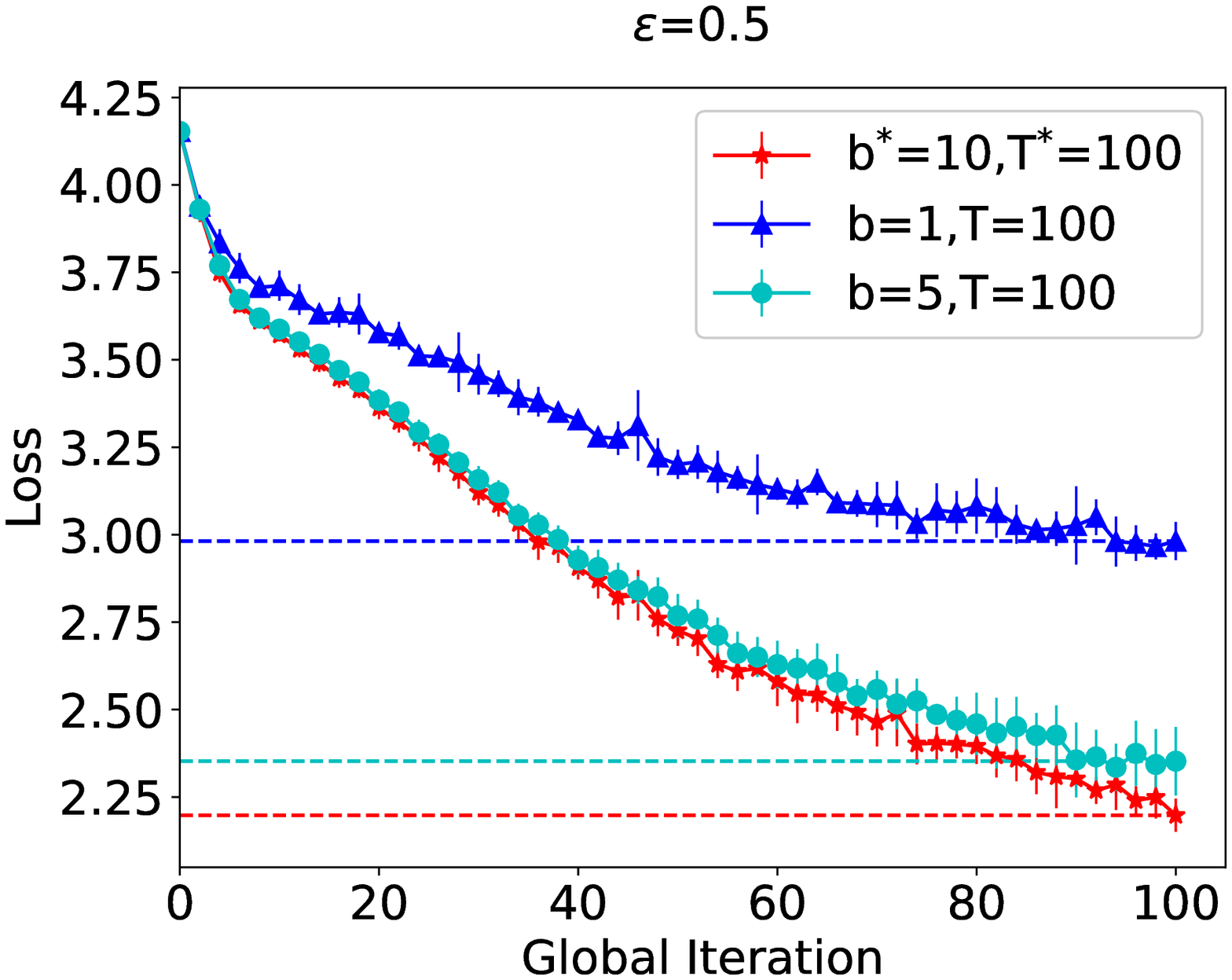}
    \includegraphics[width=6.0cm]{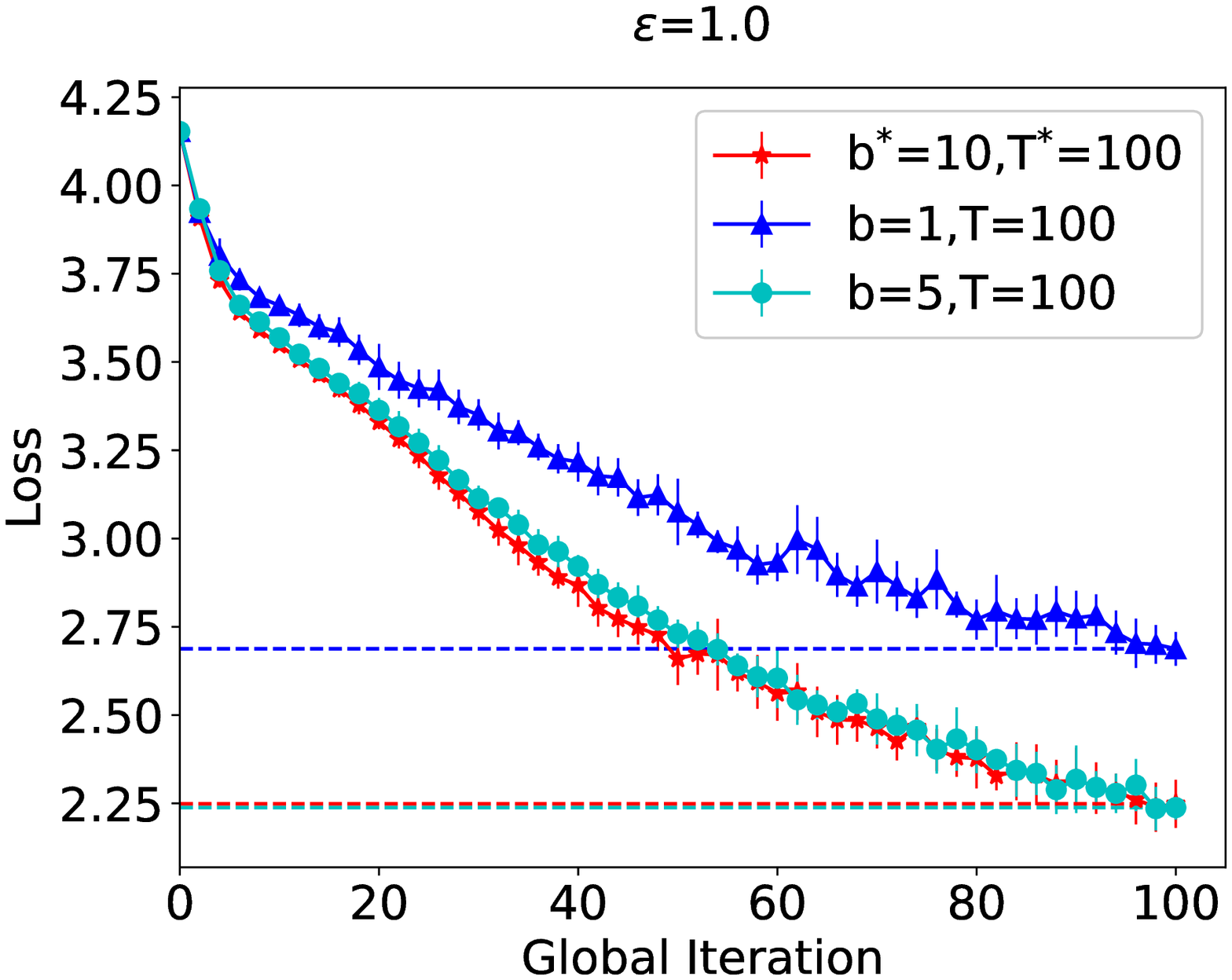}
}
\caption{Comparison of loss functions per global iteration for FEMNIST by setting different values for $b$ and $T$ in FL with DP.  There are $N=10$ clients, $b^*$ and $T^*$ are determined by our solutions, which give $b^*=1$ for the Laplace mechanism and $ b^*=10$  for  the Gaussian mechanism.}
\label{fig:loss_fe}	
\end{figure*}

\subsection{Experimental Results}
\subsubsection{Performance Comparison per Iterations}
In this experiment, we set different values for $b$ and $T$ for model training in FL with DP to demonstrate the importance to properly select $b$ and $T$. 

To observe the detailed model training process, we display the loss function together with error bars of the trained model after each global iteration in Fig.~\ref{fig:loss_mn} for MNIST and Fig.~\ref{fig:loss_fe} for FEMNIST, respectively. We compare the optimal setting with $b^*$ and $T^*$  and other settings with different values for $b$ and $T$, under different privacy budgets for both the Laplace and Gaussian mechanisms. Through observing 
Fig.~\ref{fig:loss_mn}  and Fig.~\ref{fig:loss_fe}, we can find that:
\begin{itemize}
    \item In both Fig.~\ref{fig:loss_mn}  and Fig.~\ref{fig:loss_fe}, poorly setting $b$ and $T$ can significantly inflate the final loss function, indicating that it is vital to optimize the choice of $b$ and $T$ in FL with DP. 
    \item Due to the disturbance  of DP noises, it is not wise to excessively query each client by setting very large $b$ and $T$ for the Laplace mechanism. As we can see, when $\epsilon =1$, the final loss function is very large if we set  $T=50$, which is much worse than that when $T=10 $  or $11$ for the FEMNIST case. This result is consistent with our analysis in Theorem~\ref{theorem:SGD upper buond}.
    \item In contrast, the loss of the Gaussian mechanism is minimized when we set  $b=10$ and $T=100$. The reason is that the influence of Gaussian noises will not diverge to infinity with the increase of the number of queries, and we can set relatively large $b$ and $T$. 
    \item By properly setting $b^*$ and $T^*$ in FL with DP using our solution, we can guarantee that the final loss function is minimized in this experiment for all cases.
\end{itemize}


\begin{figure}
\centering
\subfigure[Laplace Mechanism]{
	\includegraphics[width=4.3cm]{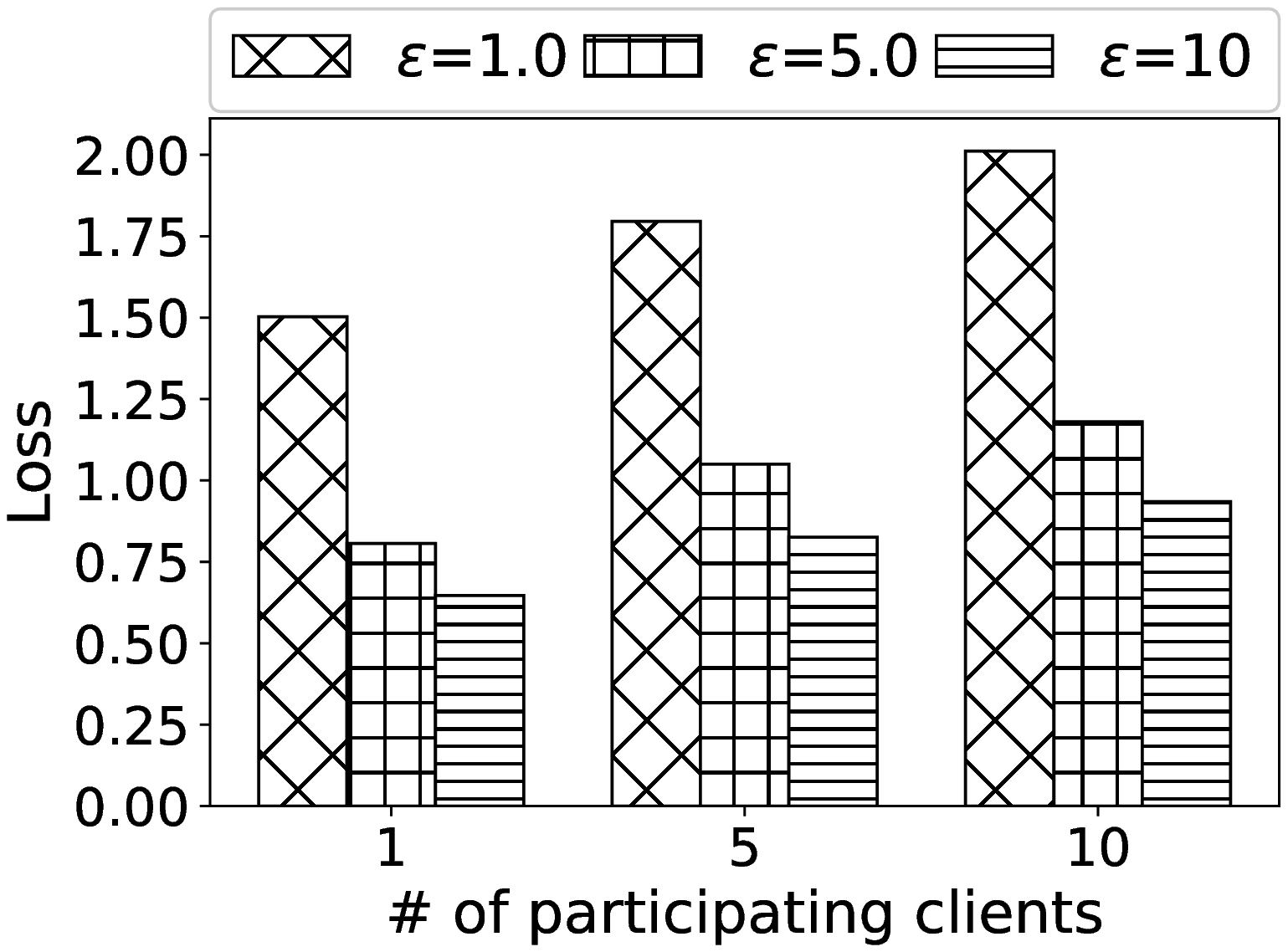}
    \includegraphics[width=4.3cm]{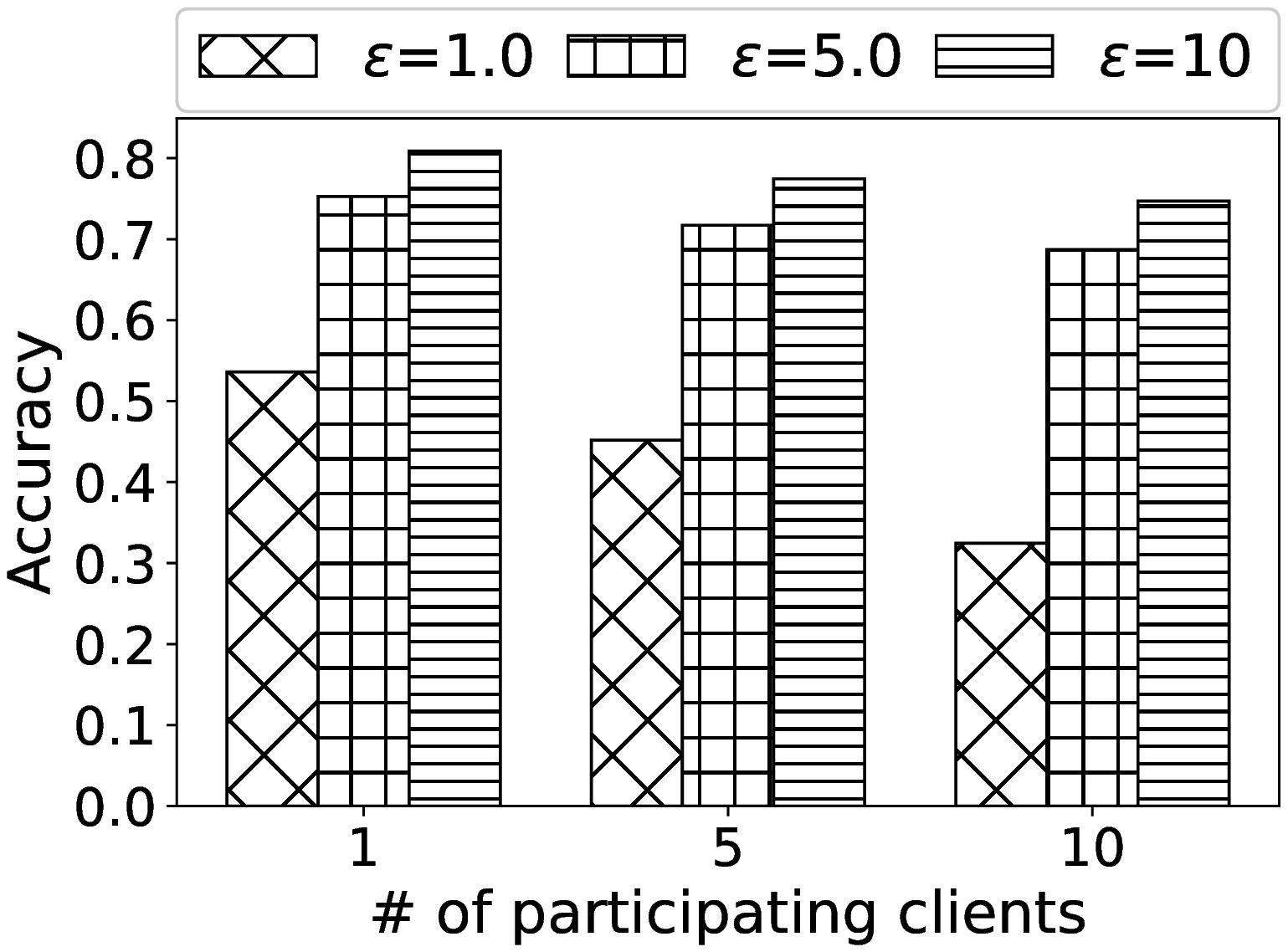}
}	
\subfigure[Gaussian Mechanism]{
	\includegraphics[width=4.3cm]{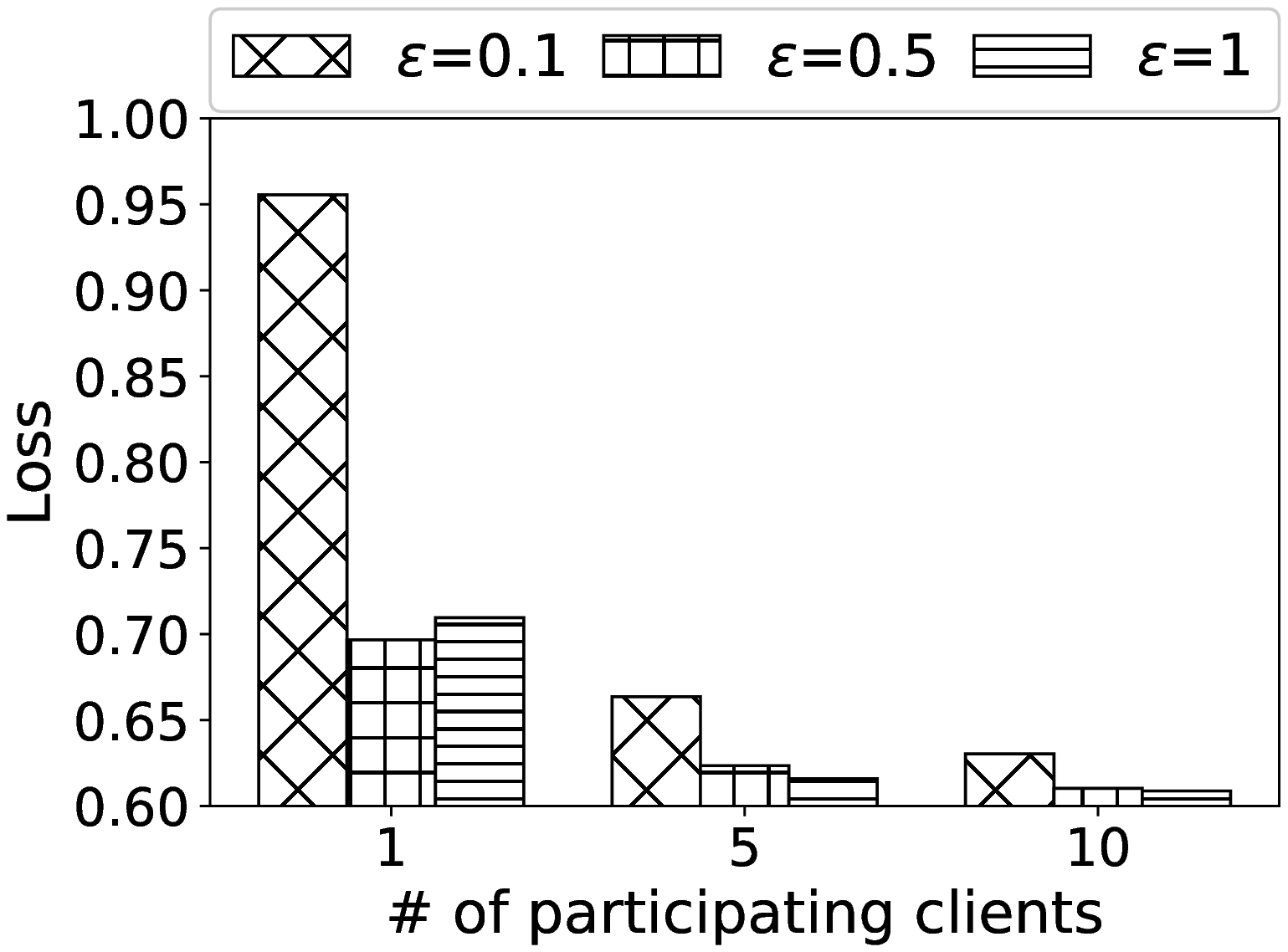}
    \includegraphics[width=4.3cm]{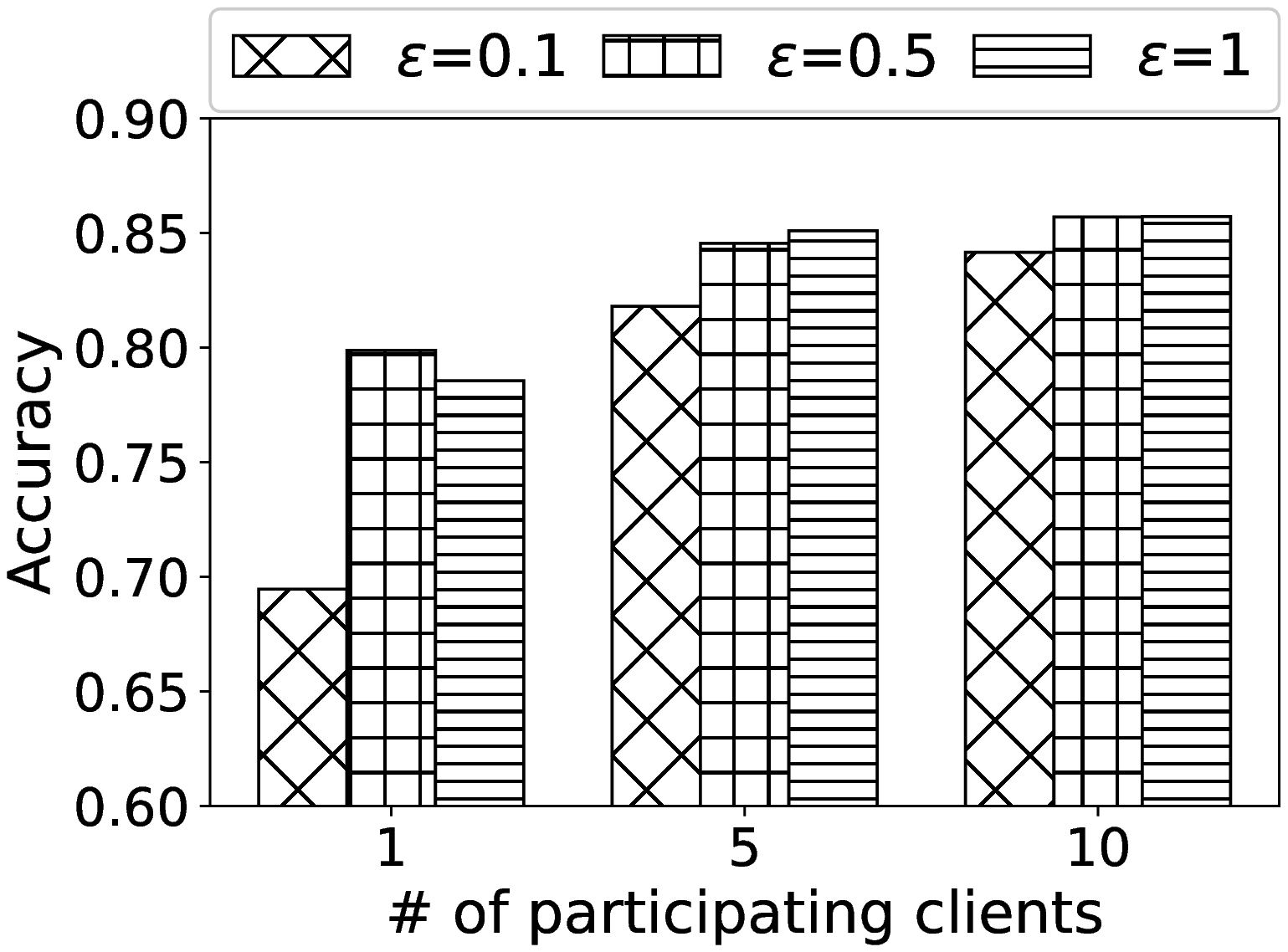}
}	
\caption{Comparison of FL performance with different $b$, \emph{i.e.}, the number of participating clients in each global iteration, on the MNIST dataset. }
\label{fig:sampling_mn}	
\end{figure}

\begin{figure}[ht]
\centering
\subfigure[Laplace Mechanism]{
	\includegraphics[width=4.3cm]{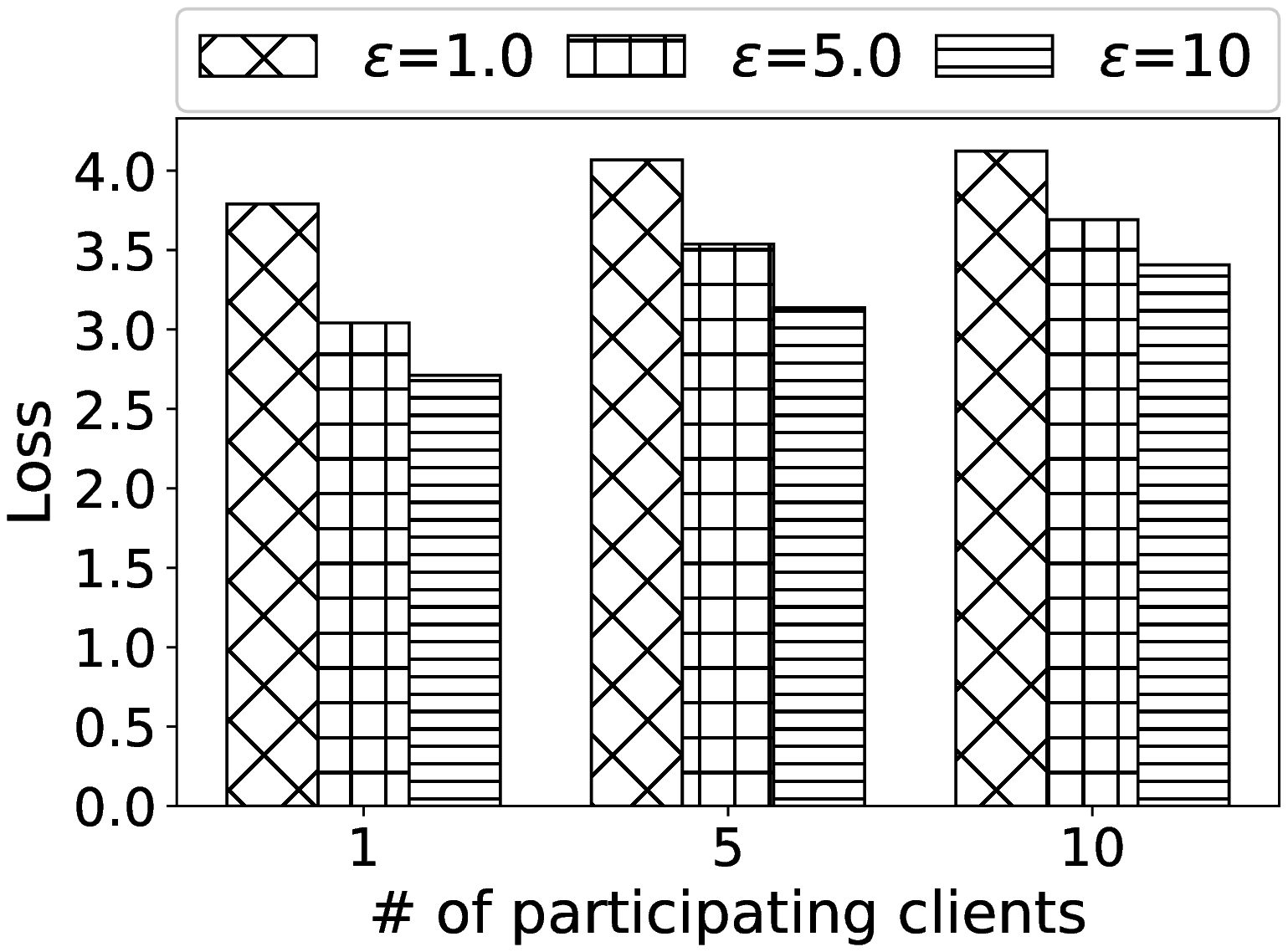}
	\includegraphics[width=4.3cm]{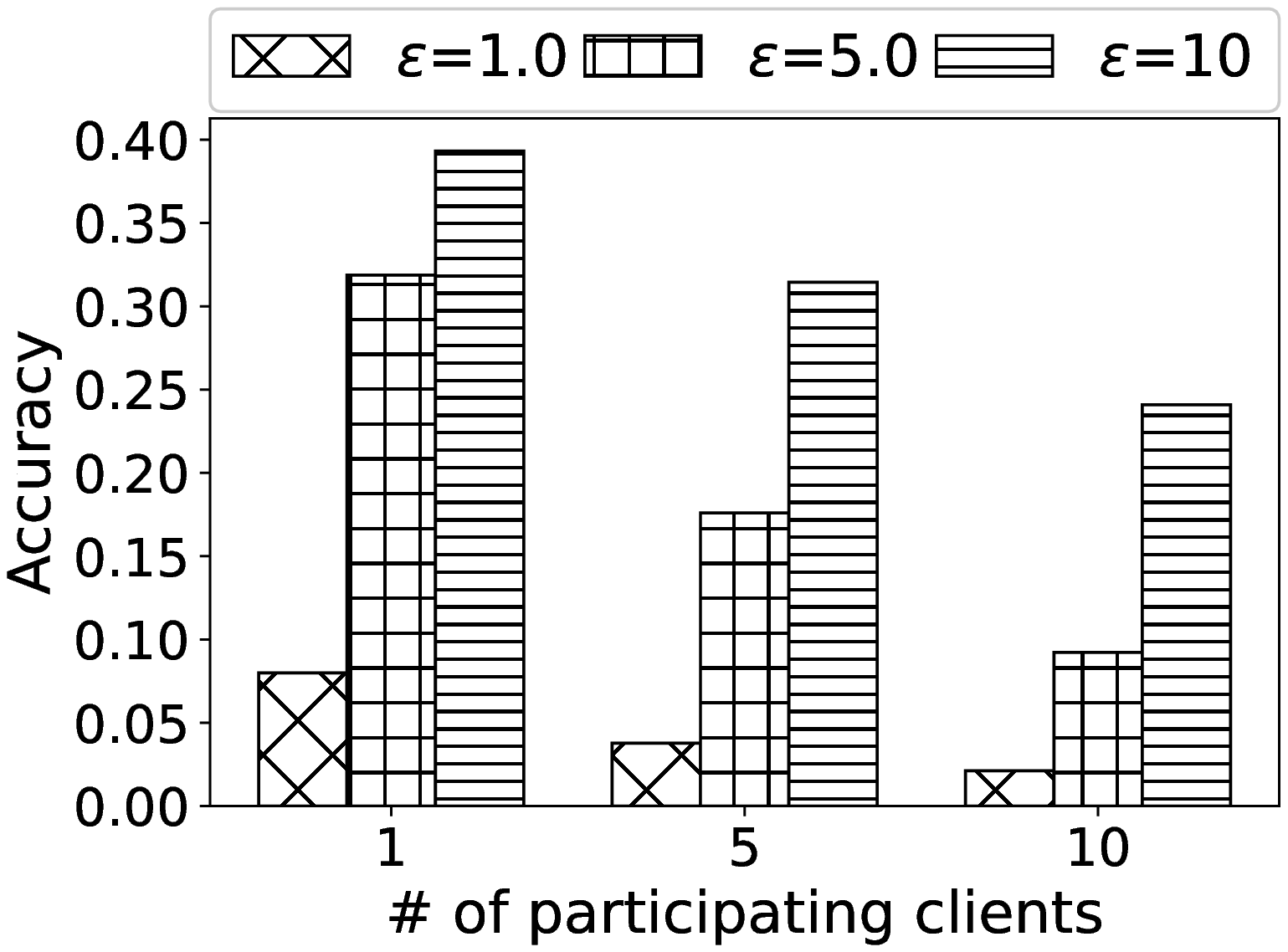}
}	
\subfigure[Gaussian Mechanism]{
	\includegraphics[width=4.3cm]{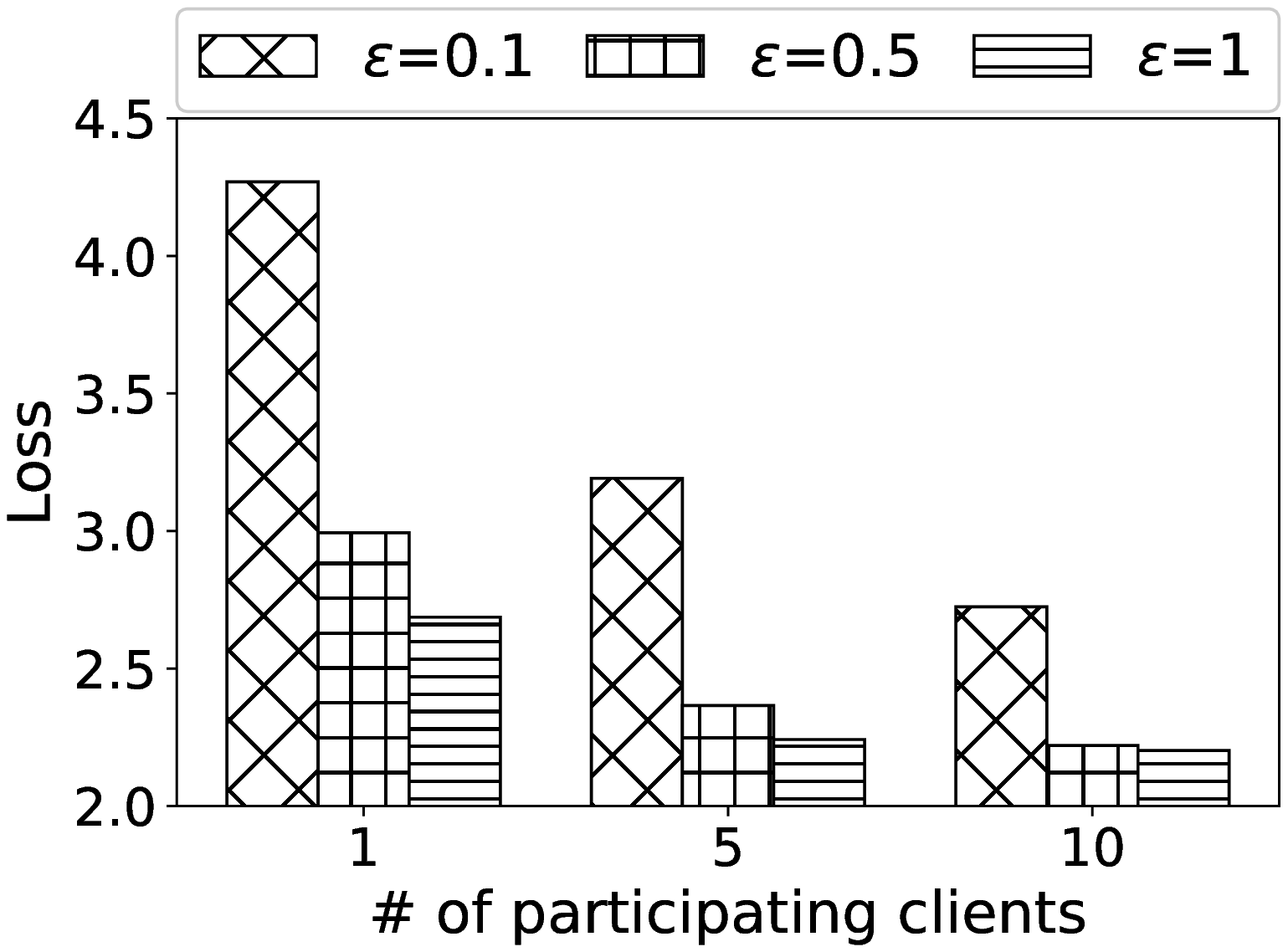}
	\includegraphics[width=4.3cm]{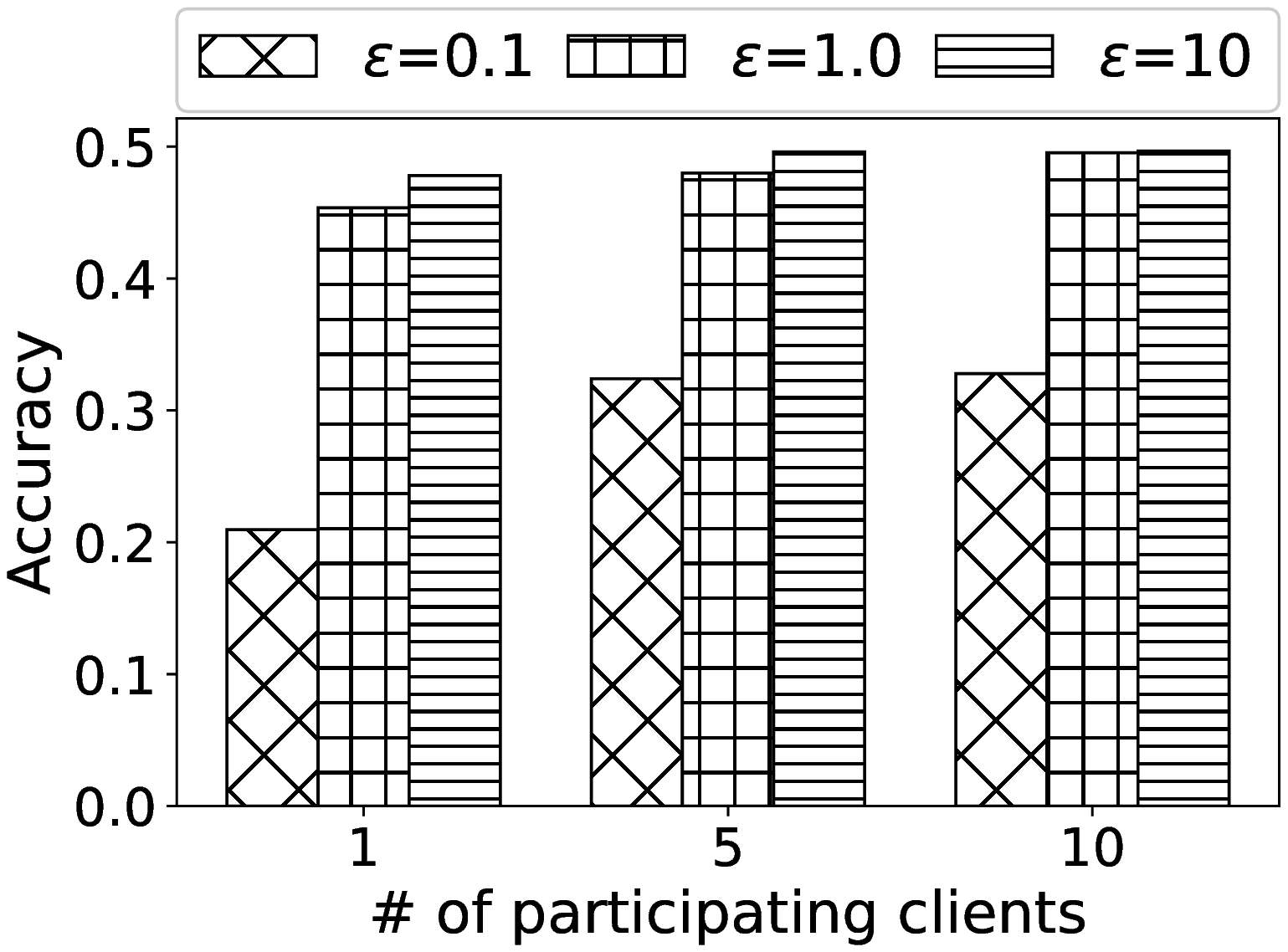}
}	
\caption{Comparison of model performance with different $b$, \emph{i.e.}, the number of participating clients in each global iteration, on the FEMNIST dataset. }
\label{fig:sampling_fe}	
\end{figure}

\subsubsection{Comparison with Different $b$}

Recall that $b$ is the number of clients selected to participate each round of global iteration. The value of $b$ influences the number of replies each client needs to respond the PS. Thus, we conduct this experiment by enumerating $b$ as $1$, $5$  and $10$ to observe its influence for both DP mechanisms. 
 For each enumerated $b$, we derive the optimal $T^*$ based on our analysis. The derived values of $T$ are displayed in Table~\ref{tab:optimal}.  For the Gaussian mechanism, according to our analysis, $T$ should be set as large as possible. Empirically, we fix $T=100$ since the decrease of loss function is insignificant anymore when  $T>100$.

\renewcommand{\arraystretch}{1.5}
\begin{table*}[ht]
    \centering
    \begin{tabular}{p{0.8cm}p{0.1cm}cccccccccccc}
    \toprule
    \multirow{2}{*}{Dataset}& 
    \multicolumn{1}{c}{}&
     \multicolumn{3}{c}{$T^*(\textit{fix } b=5)$}&
    \multicolumn{3}{c}{$T^*(\textit{fix } b=10)$}&
    \multicolumn{3}{c}{$b^*(\textit{fix } T=50)$}&
    \multicolumn{3}{c}{$T^* (b^*=1)$}\cr\cline{2-14}
    &$\epsilon$&$1.0$&$5.0$&$10$&$1.0$&$5.0$&$10$&$1.0$&$5.0$&$10$&$1.0$&$5.0$&$10$\cr
    \midrule
    MNIST&&5&32&67&2&20&42&1&3&5&22&121&244\cr\hline
    FEMNIST&&2&22&48&1&14&31&1&2&3&11&70&145\cr
    \bottomrule
    \end{tabular}
    \caption{$T^*$ or $b^*$ for Laplace mechanism on the MNIST and FEMNIST dataset.}
    \label{tab:optimal}
\end{table*}

The comparisons of the loss function and model accuracy are presented in Fig.~\ref{fig:sampling_mn} for MNIST and Fig.~\ref{fig:sampling_fe} for FEMNIST, respectively under the Laplace and Gaussian mechanisms. From this experiment, we can draw the following findings:
\begin{itemize}
    \item The value of $b$ can significantly affect both the final loss function and the model accuracy. Poorly setting $b$ can substantially lower the model accuracy, \emph{e.g.}, setting $b=10$ under the Laplace mechanism with $\epsilon=1$. 
    \item For the Laplace mechanism, the loss function is higher and the model accuracy is lower if $b$ is bigger. This result is consistent with our conclusion in Theorem~\ref{theorem:SGD upper buond}, in which the influence of the noise term on the model convergence increases with $b$.  
    \item In contrast, the loss function is smaller and the model accuracy is higher if we set a larger $b$ under the Gaussian mechanism. According to Theorem~\ref{theorem:SGD upper buond Gau}, the influence of the noise term keeps stable even if $T$ approach infinity, which is independent with the value $b$. Thus, setting a larger $b$ can accelerate the convergence of the term without noises, \emph{i.e.}, $\omega_0'$ term in Theorem~\ref{theorem:SGD upper buond Gau}. 
\end{itemize}


\begin{figure}
\centering
\subfigure[Laplace Mechanism]{
	\includegraphics[width=4.3cm]{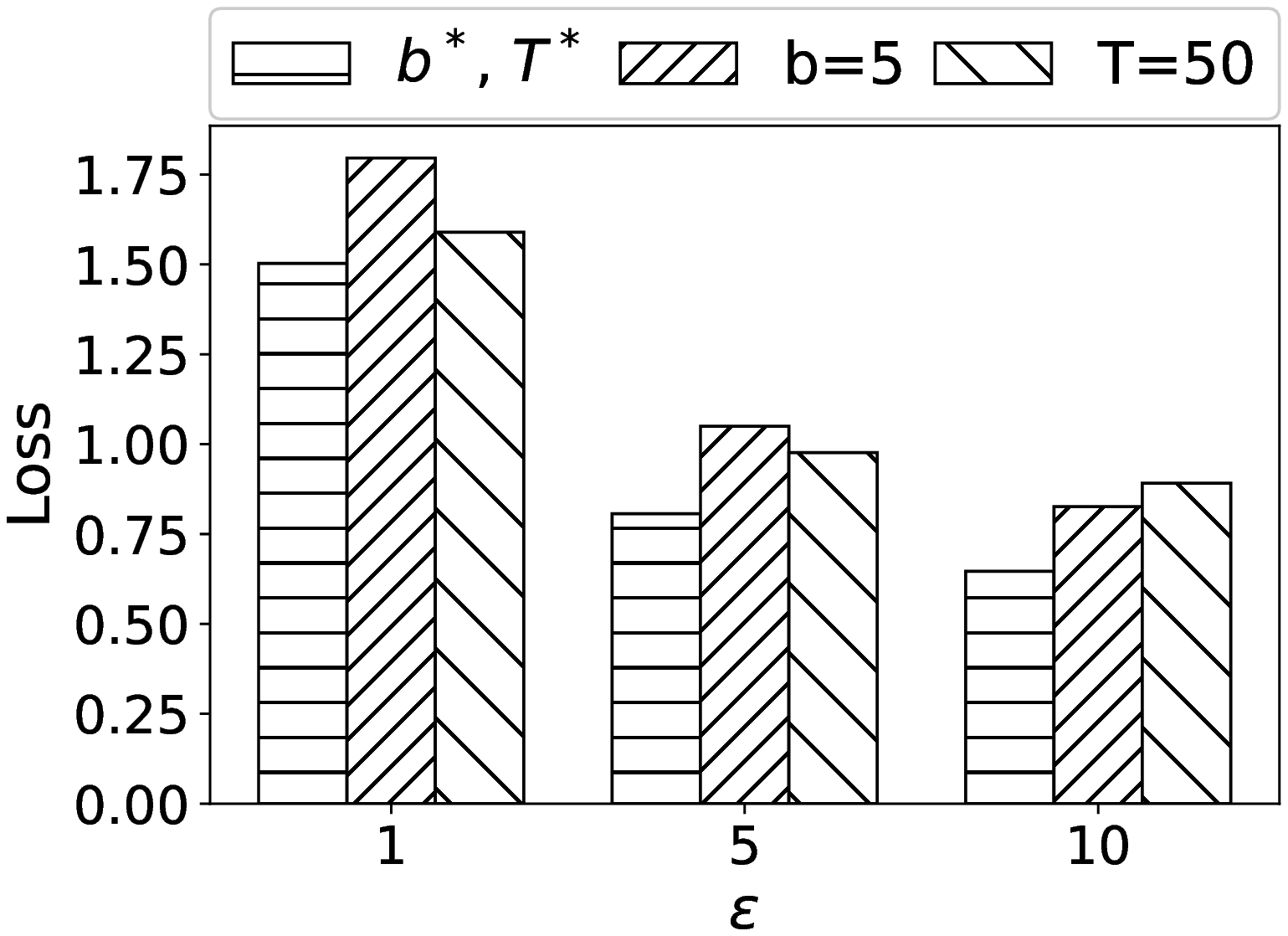}
	\includegraphics[width=4.3cm]{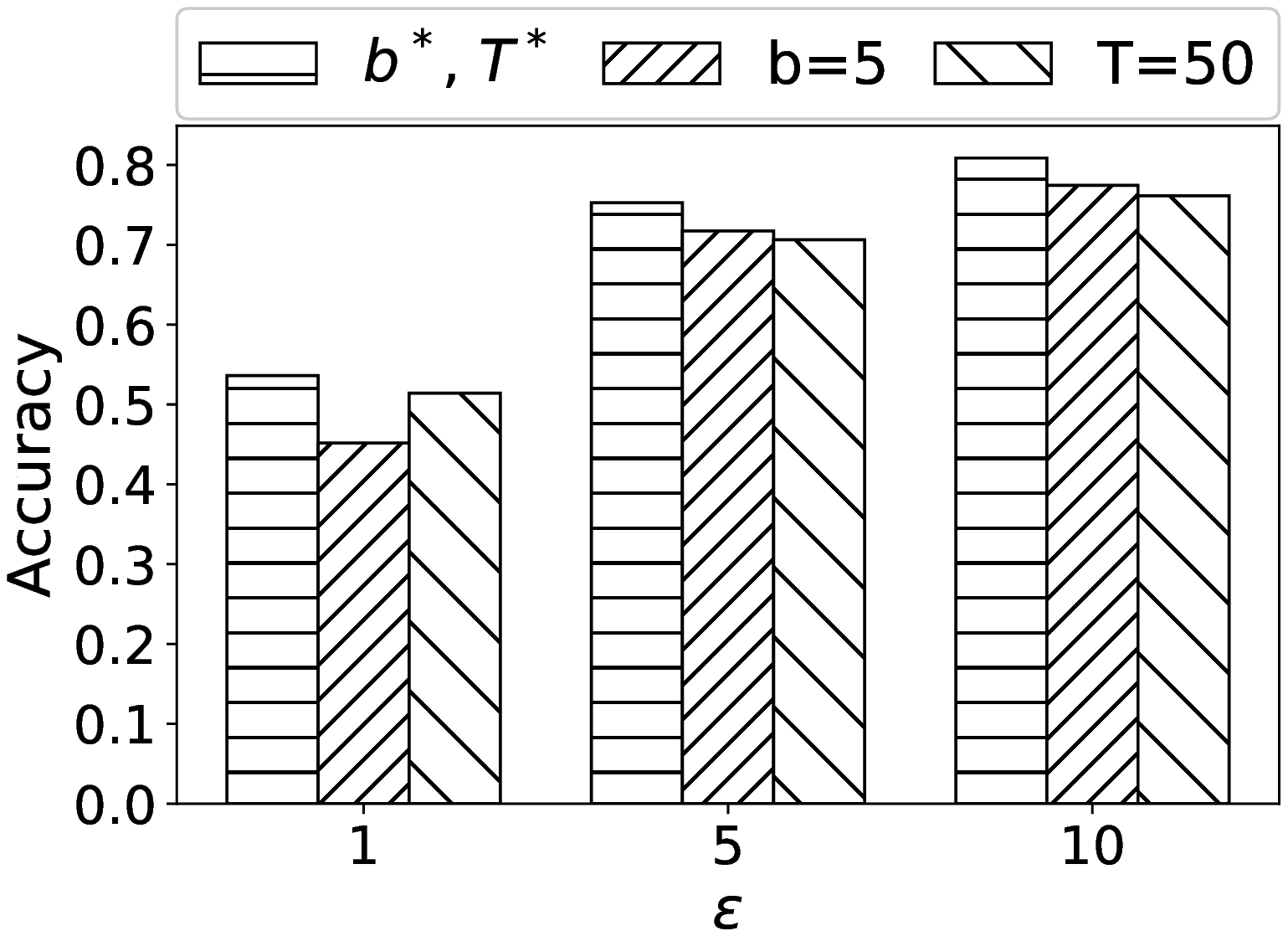}
}
\subfigure[Gaussian Mechanism]{
	\includegraphics[width=4.3cm]{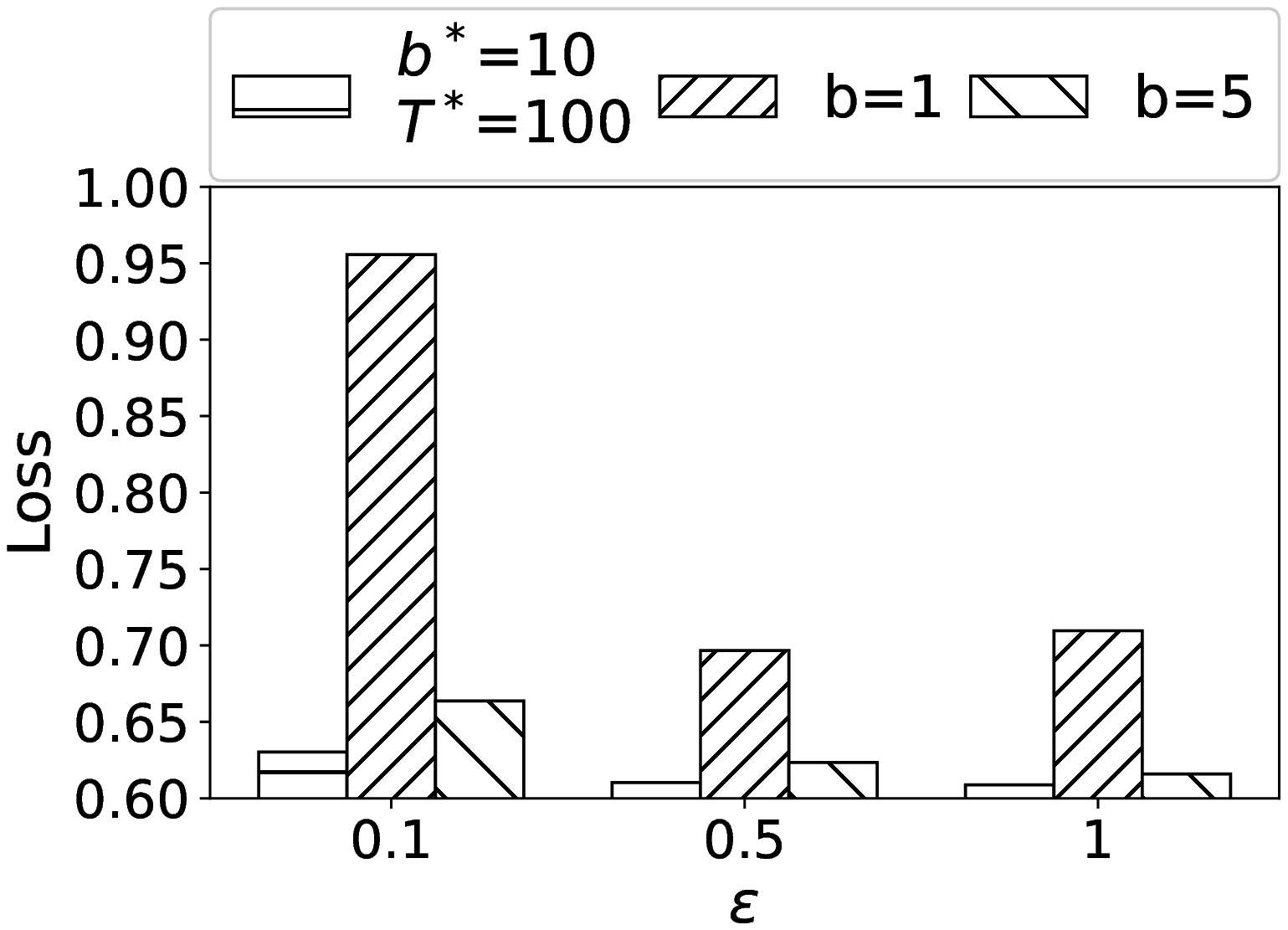}
	\includegraphics[width=4.3cm]{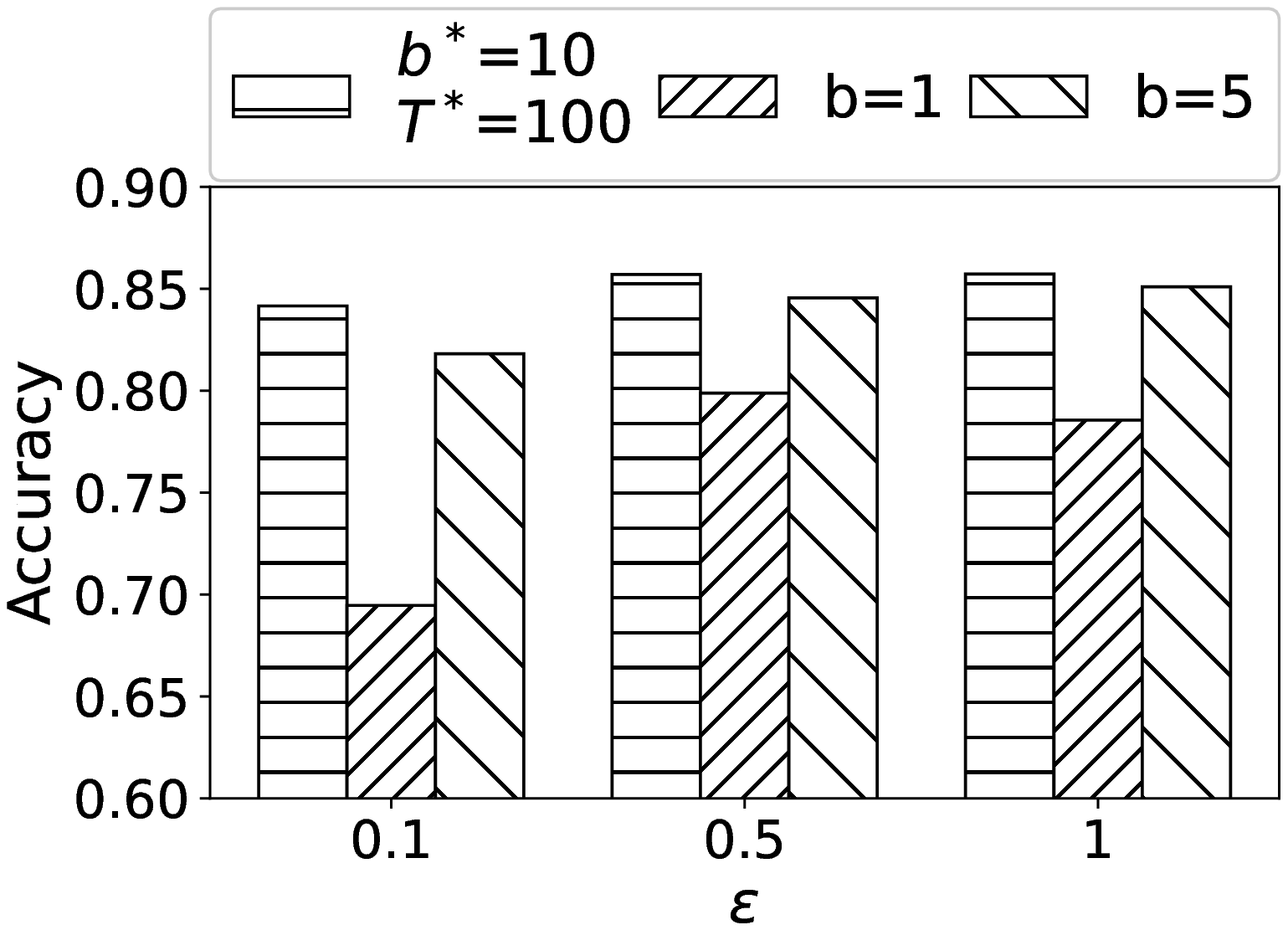}
}
\caption{Comparison of model performance with different privacy budgets on the MNIST dataset. }
\label{fig:eps_mn}	
\end{figure}

\begin{figure}[ht]
\centering
\subfigure[Laplace Mechanism]{
	\includegraphics[width=4.3cm]{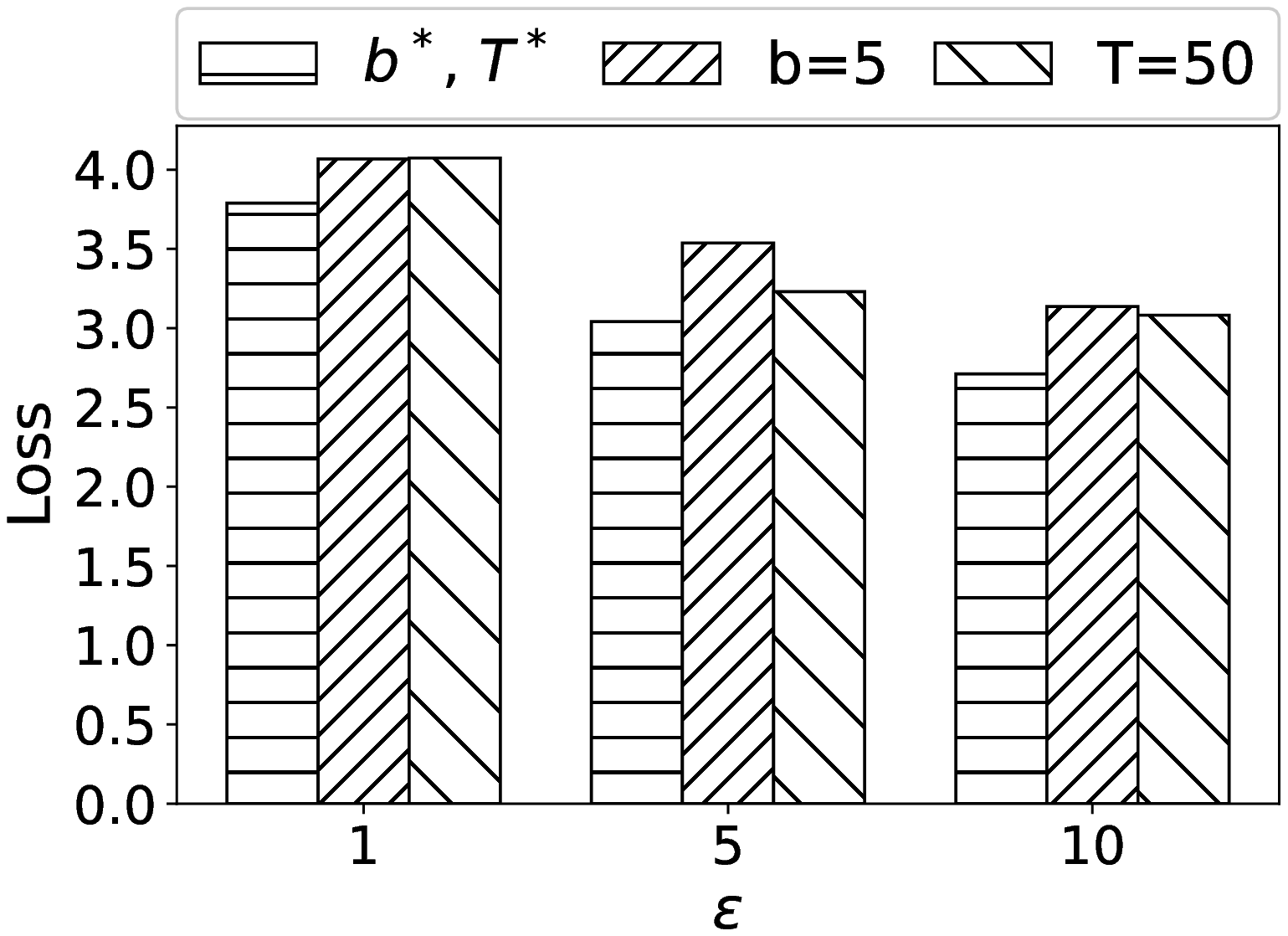}
	\includegraphics[width=4.3cm]{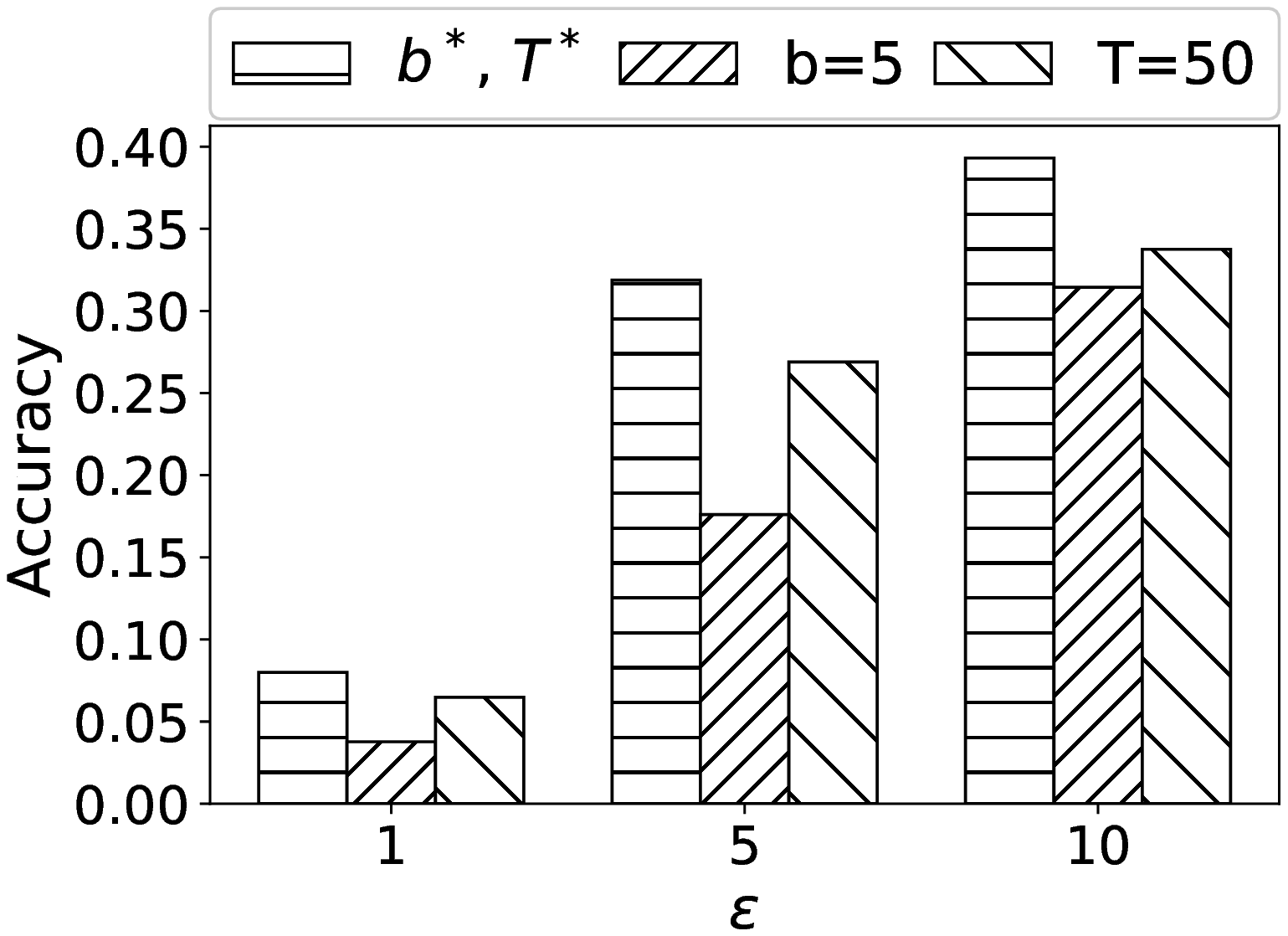}
}
\subfigure[Gaussian Mechanism]{
	\includegraphics[width=4.3cm]{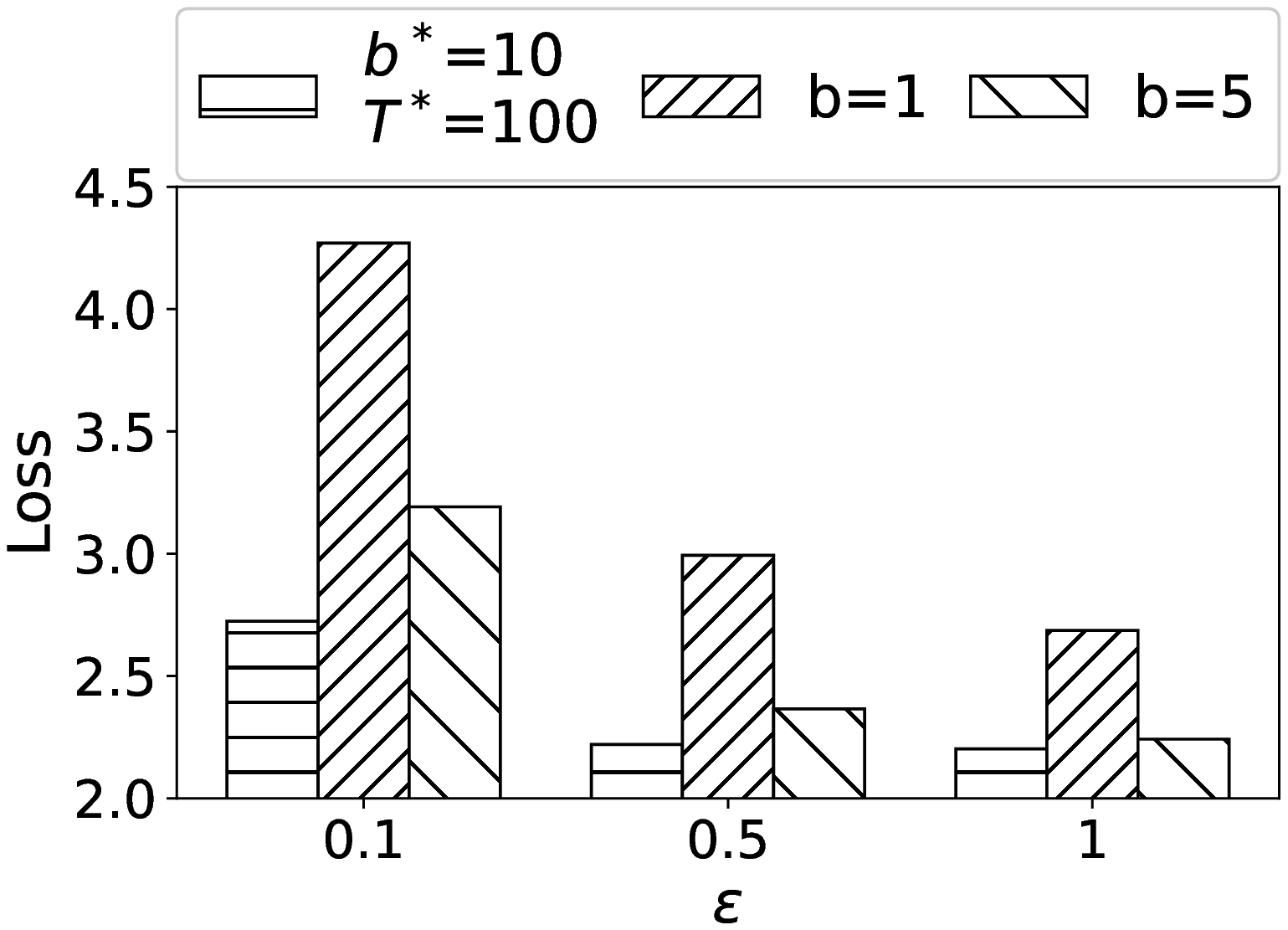}
	\includegraphics[width=4.3cm]{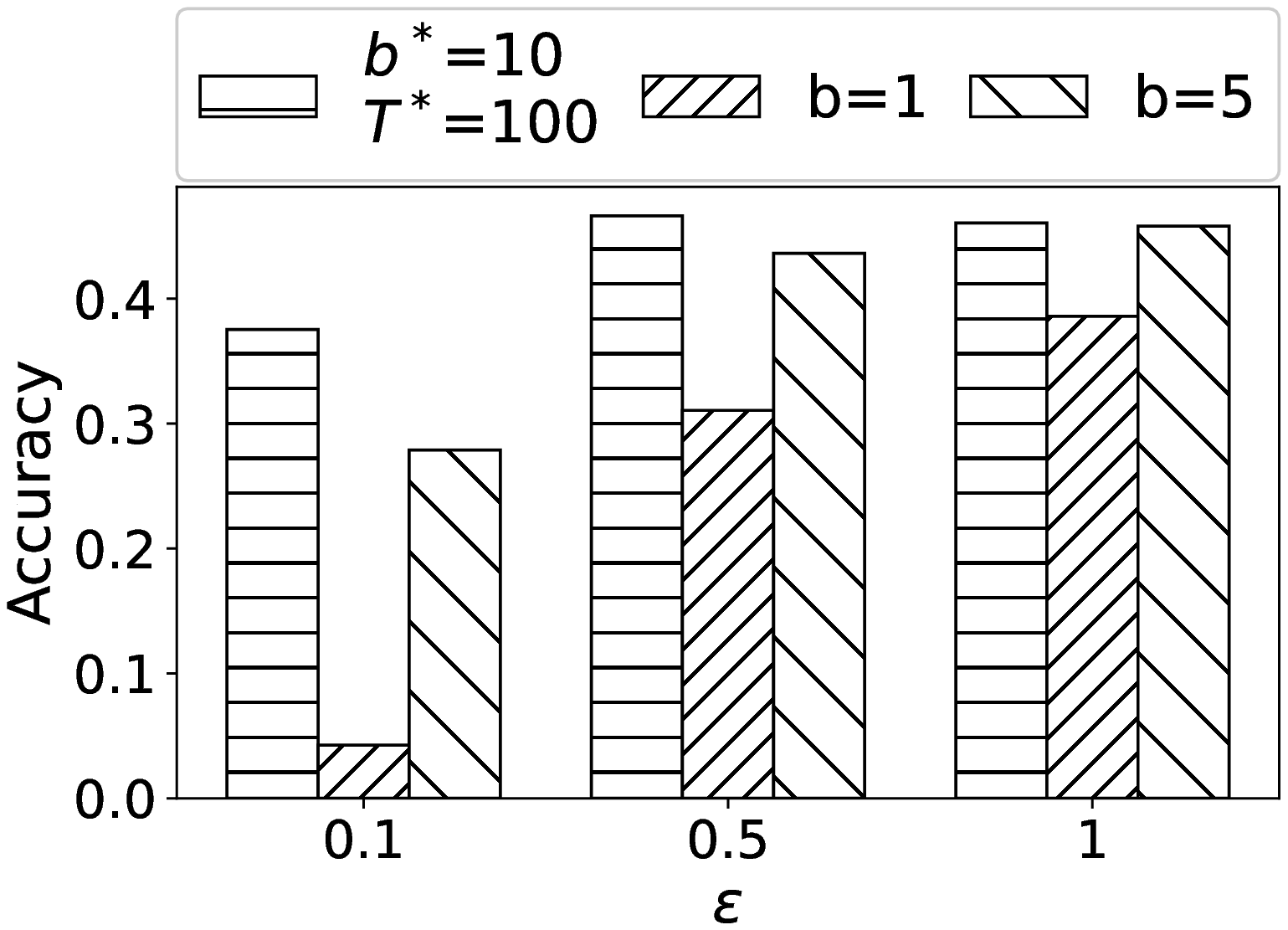}
}
\caption{Comparison of model performance with different privacy budgets on the FEMNIST dataset. }
\label{fig:eps_fe}	
\end{figure}

\subsubsection{Comparison with Different Privacy Budgets}
Next, we compare the FL performance  by enumerating the privacy budgets. The experiment results are presented in Fig.~\ref{fig:eps_mn} for MNIST  and Fig.~\ref{fig:eps_fe} for FEMNIST, respectively.
For the Laplace mechanism, we compare the optimal setting with $b^*, T^*$ with other settings which are set according to Table~\ref{tab:optimal} as well. 
For the Gaussian mechanism, according to Theorem~\ref{theorem:SGD upper buond Gau}, we should always set a large $T$ to minimize the loss function. Thus, we fix $T=100$ for the Gaussian mechanism, and set $b = 1, 5$ or $10$. 
From this experiment, we can observe that
\begin{itemize}
    \item A smaller privacy budget will inject DP noises with larger variances, and hence the loss function is higher and the model accuracy is lower for all cases. 
    \item The setting with $b^*, T^*$ is the best one achieving the optimal learning performance than other settings. 
    \item The model accuracy can deteriorate substantially if  $b$ and $T$ are chosen  inappropriately, especially when the privacy budget is  small, \emph{e,g,}, $\epsilon =1$ for the Laplace mechanism or $\epsilon =0.1$ for the Gaussian mechanism.   
    
\end{itemize}

\begin{figure}
\centering
\subfigure[Laplace Mechanism]{
	\includegraphics[width=4.3cm]{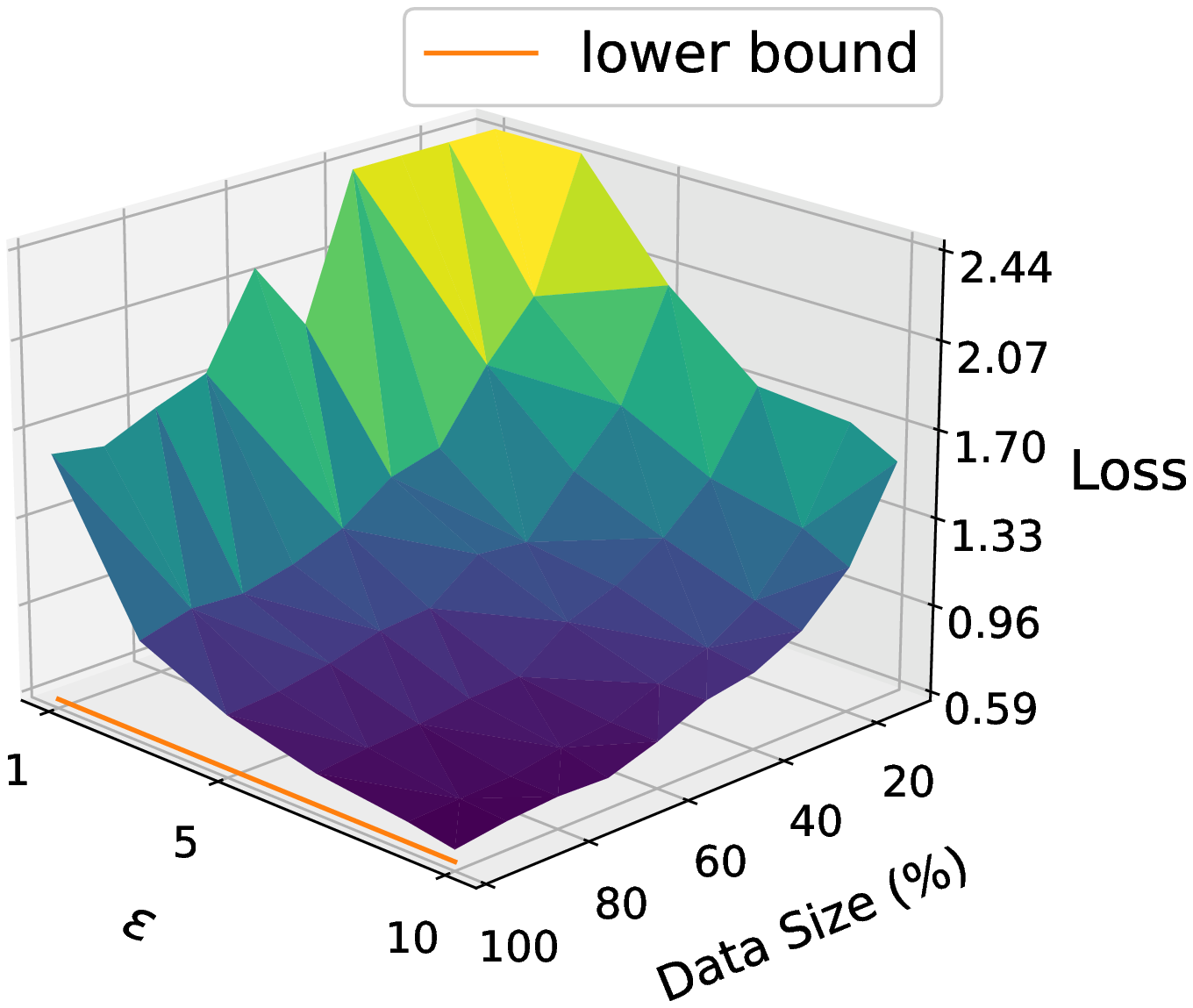}
	\includegraphics[width=4.3cm]{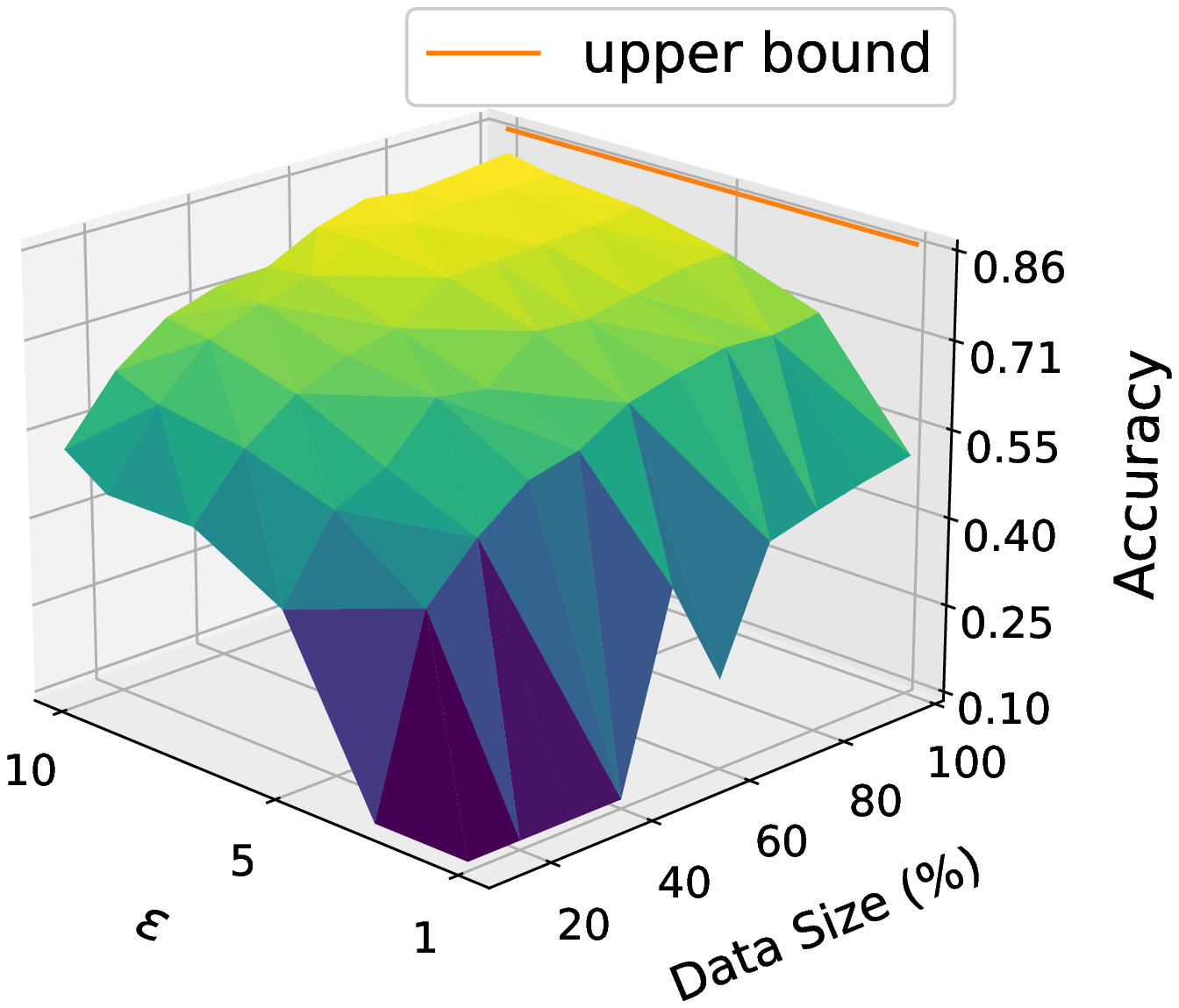}
}	
\subfigure[Gaussian Mechanism]{
	\includegraphics[width=4.3cm]{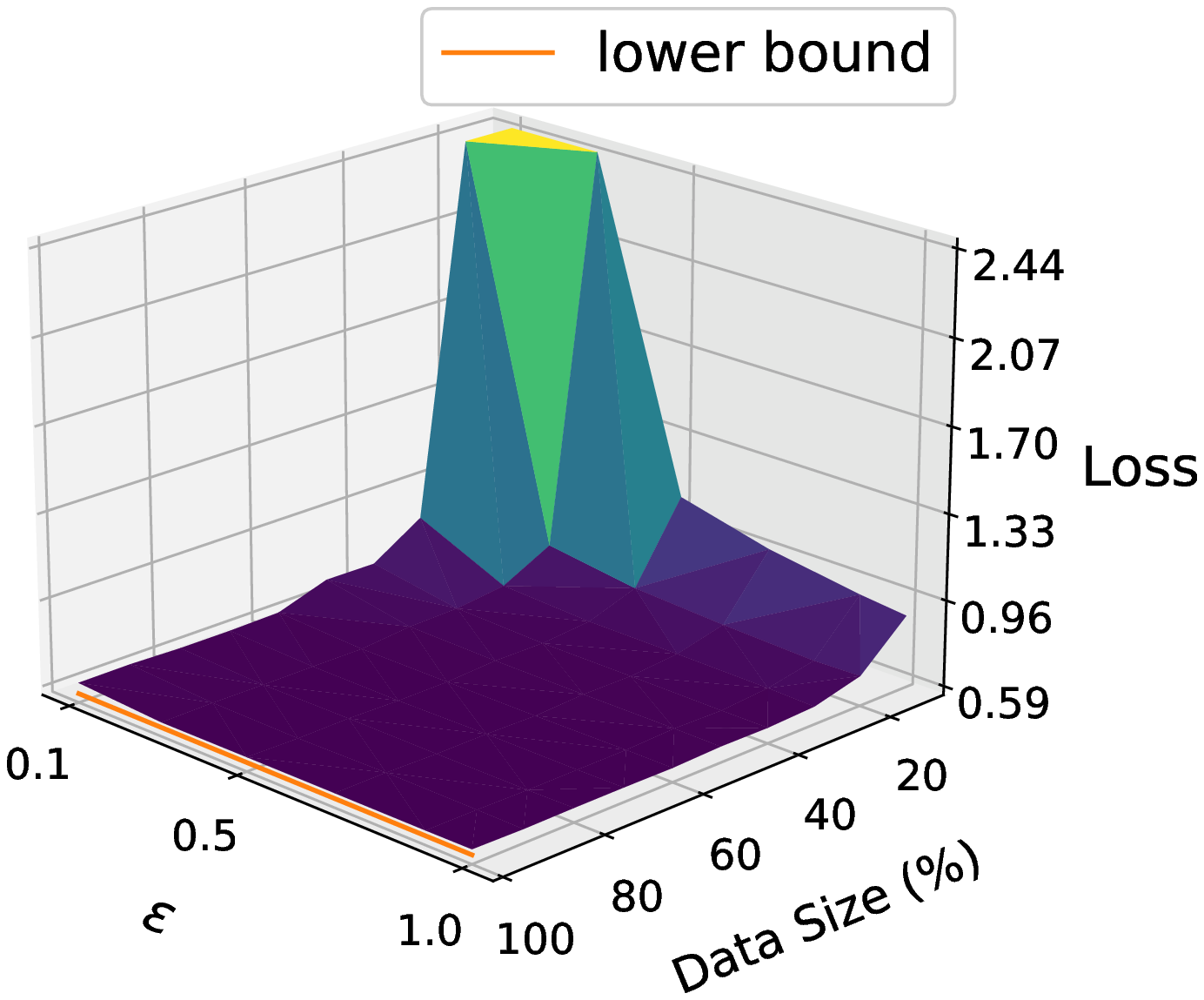}
	\includegraphics[width=4.3cm]{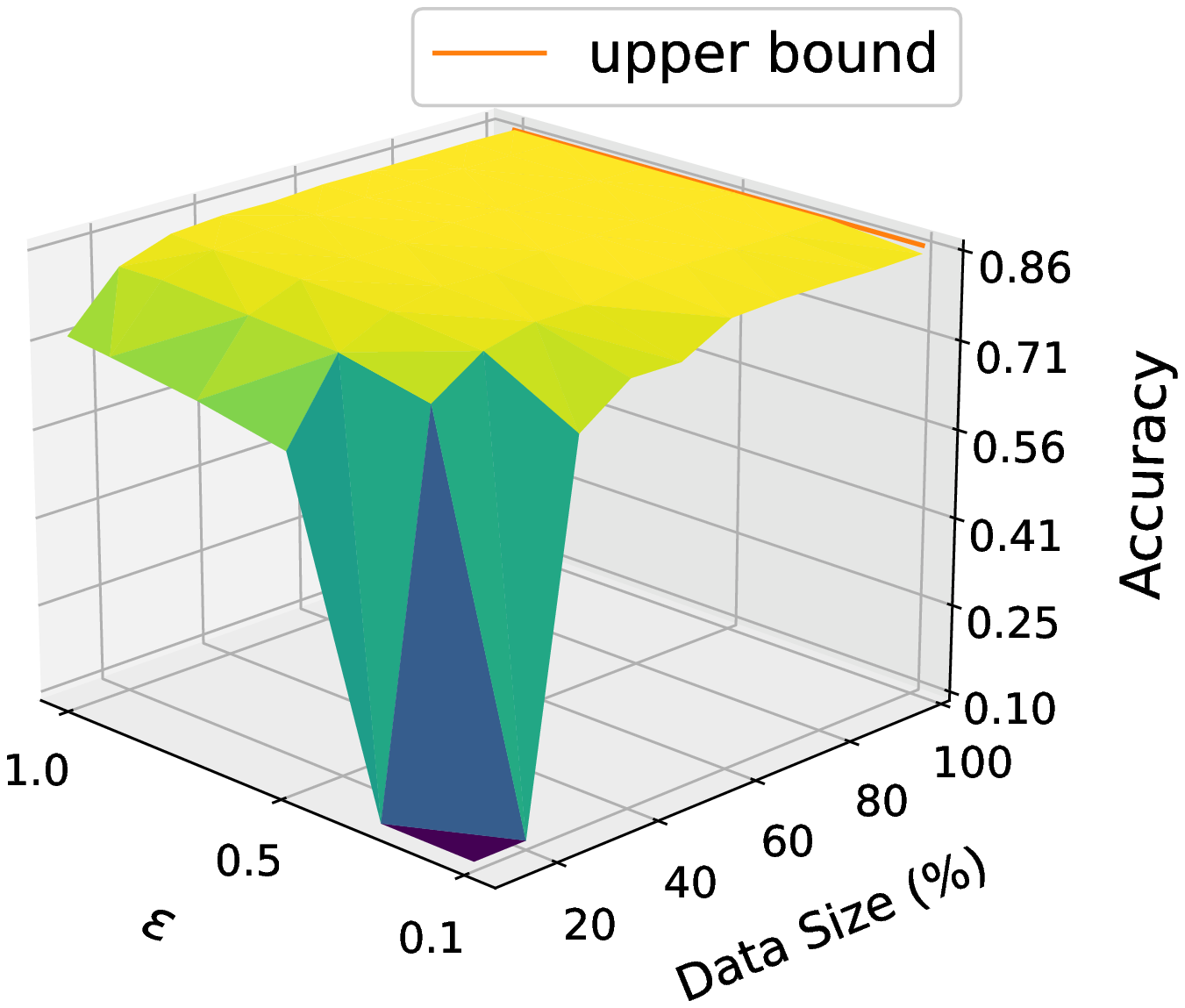}
}	
\caption{Optimal model performance  for MNIST with different data sizes and privacy budgets.  }
\label{fig:datasize_mn}	
\end{figure}

\subsubsection{Practicality of DP in FL}

\begin{figure}
\centering
\subfigure[Laplace Mechanism]{
	\includegraphics[width=4.3cm]{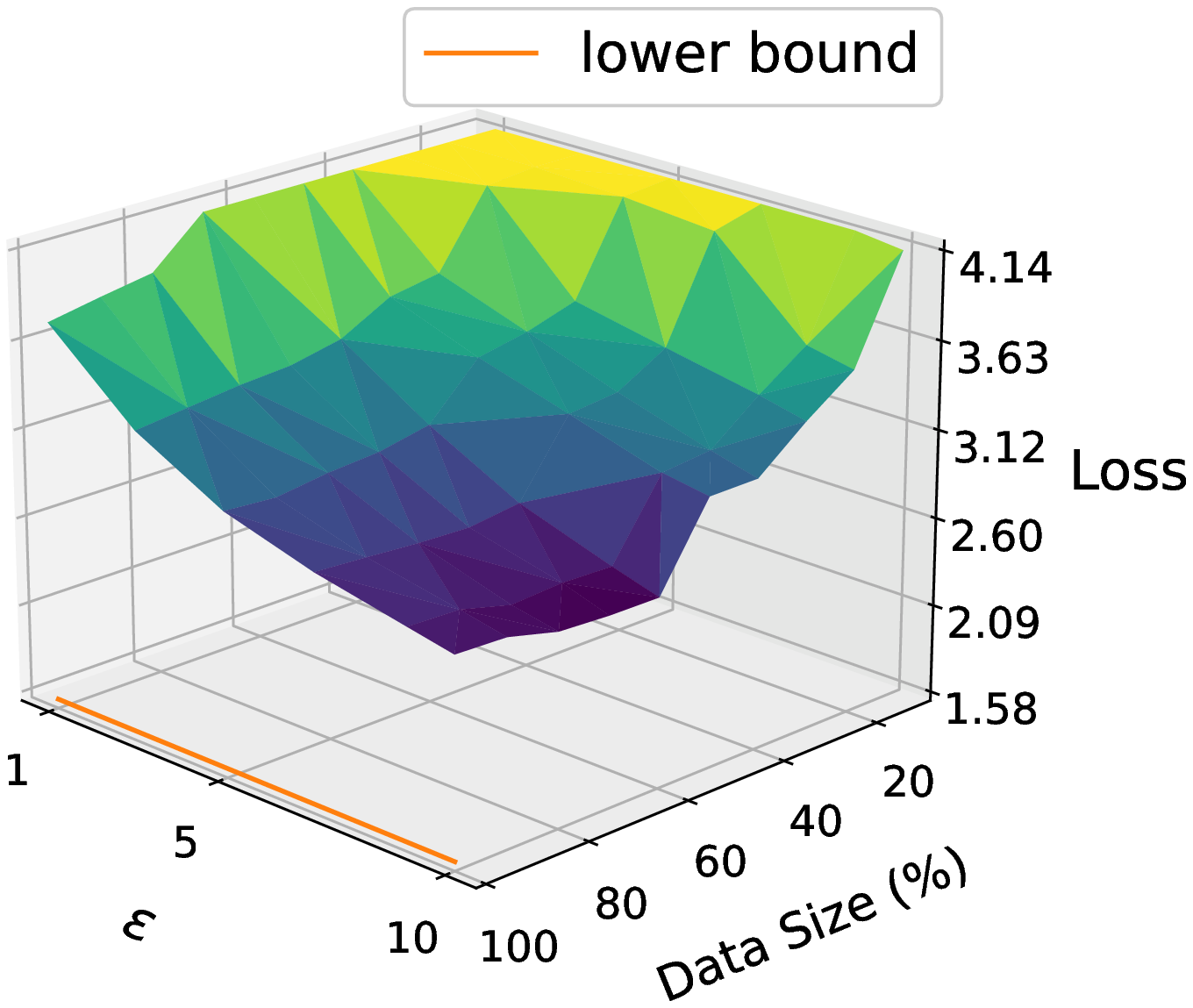}
	\includegraphics[width=4.3cm]{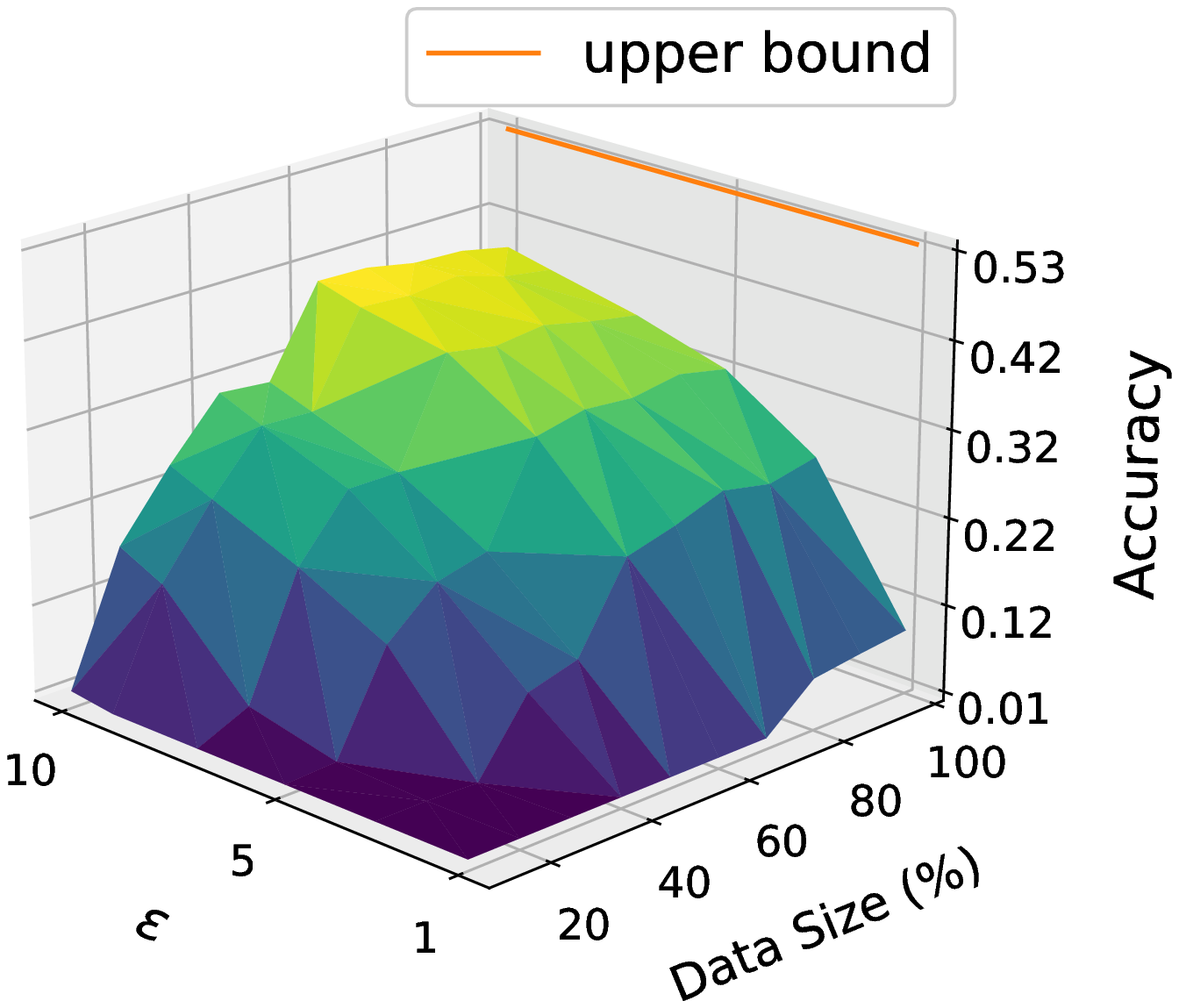}
}	
\subfigure[Gaussian Mechanism]{
	\includegraphics[width=4.3cm]{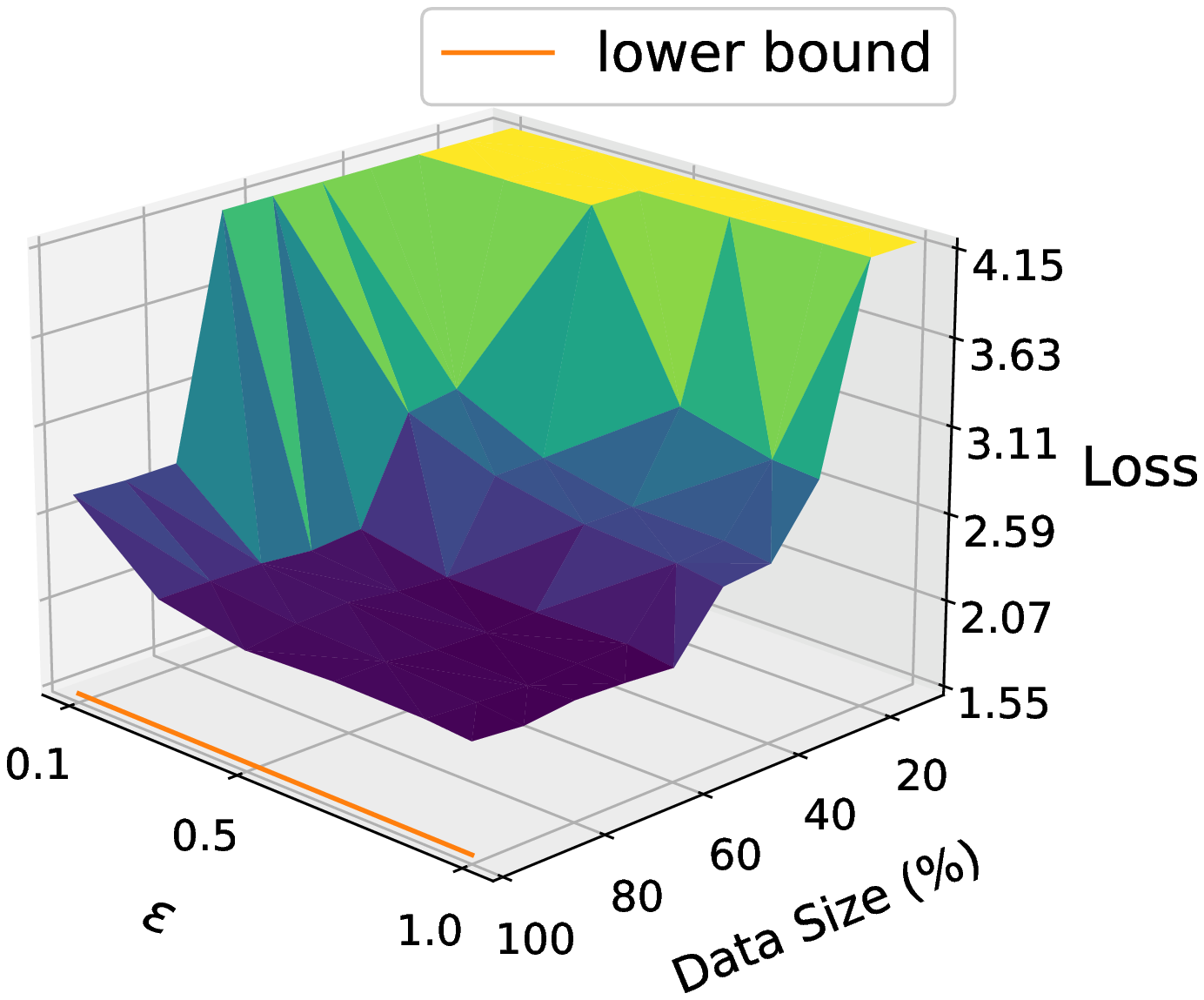}
	\includegraphics[width=4.3cm]{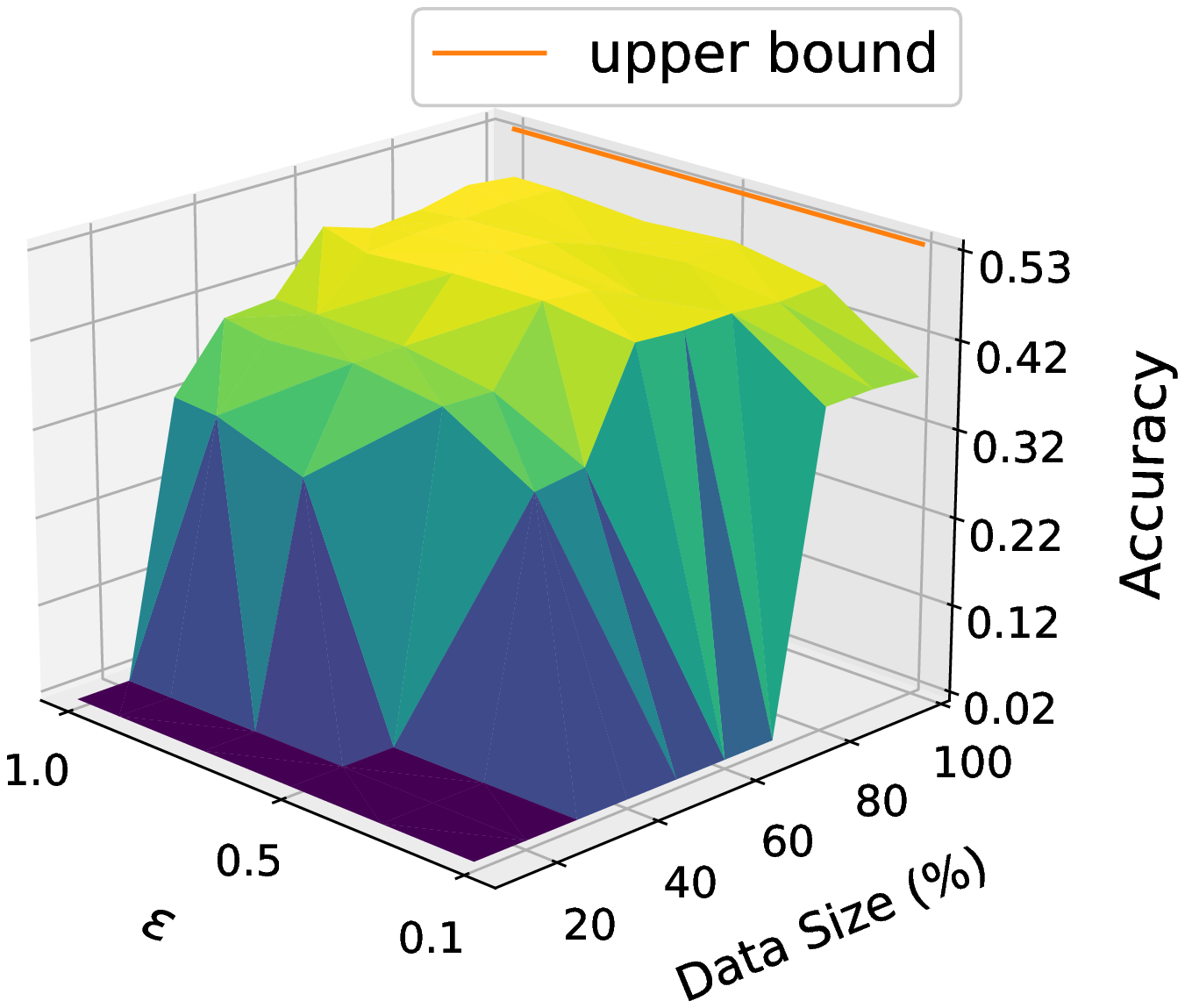}
}	
\caption{Optimal model performance  for FEMNIST with different data sizes and privacy budgets. }
\label{fig:datasize_fe}	
\end{figure}

DP mechanisms can inevitably impair the model accuracy in FL.  Thus, it is worth to exploring the practicability of  DP mechanisms in FL. In other words, a DP mechanism is not practicable if it lowers the model accuracy significantly. In this experiment, we evaluate the practicability of DP mechanisms by varying the privacy budget and the portion of samples used by each client, \emph{i.e.}, data size for training. Meanwhile, we set $b^*, T^*$ in FL which can achieve the optimal learning performance.  
The privacy budget of each client is varied from  $1$ to  $10$ for the Laplace mechanism, and from $0.1$ to $1$ for the Gaussian mechanism. The portion of data samples used by each client for FL is varied from 10\% to 100\% with 10\% intervals. Experiment results are presented in Fig.~\ref{fig:datasize_mn} for MNIST  and Fig.~\ref{fig:datasize_fe} for FEMNIST, from which we can observe:
\begin{itemize}
    \item The model accuracy is lower if the privacy budget is smaller or the portion of used data samples is smaller. 
    \item The Laplace mechanism can impair the model accuracy significantly even if we set $b^*, T^*$ in FL.  The model accuracy is very poor if either the privacy budget or the portion of used samples is small. This implies that it is not easy to apply the Laplace mechanism in practice. 
    \item In contrast, the model accuracy of FL with the Gaussian mechanism is very high even when the privacy budget is very small and only a small fraction of samples are used for training. Thus, the Gaussian mechanism is more friendly for  practice.   
\end{itemize}


\section{Conclusion and Future Work}
\label{Sec:Conclusion}
Disclosing plain parameter information can leak data privacy in FL.   Differential privacy is an effective mechanism to protect parameters by disturbing original parameters with noises. However, it is unknown to what extent the DP mechanism can impair model accuracy, and how can we minimize this negative influence.  In this work, we investigate this problem by optimizing the numbers of queries of the PS and replies of clients  in FL in order to maximize the final model accuracy with two popular DP mechanisms (\emph{i.e.}, the Laplace and Gaussian mechanisms). Through conducting convergence rate analysis, we prove that this is a biconvex optimization problem. 
We derive the closed-form solutions for the optimal numbers of queries and replies, based on which we further  discuss the implications of each solution.  Theoretical analysis are validated with comprehensive experiments conducted with MNSIT and FEMNIST datasets. Our work can provide theoretical basis for setting  the numbers of queries and replies  when incorporating DP into FL. 

The design space of FL with DP is very broad. Optimizing the numbers of queries and replies is only one of the ways to prohibit the negative influence of DP noises. It is also prospective to optimize FL with DP with other approaches such as designing more advanced DP mechanisms and more sophisticated model average algorithms.



\bibliographystyle{IEEEtran}
\bibliography{IEEEabrv,b}


\appendix

\subsection{Proof of Proposition~\ref{pro:noise}}
\label{APP:Prop1}
\begin{proof}
We use the fact that $\mathbf{w}_{t}^b$ are IID random variables from $b$ clients. One can derive
\begin{equation*}
\begin{split}
    \mathbb{E}\left\{\left\|\mathbf{w}_{t}^b\right\|_2^2\right\}
    &=\frac{N^2}{b^2d^2}\mathbb{E}\left\{\left\|\sum_{i\in\mathcal{P}_t}d_i\eta_t\mathbf{w}_{t}^i\right\|_2^2\right\},\\
    &=\frac{\eta_t^2N^2}{b^2d^2}\sum_{i\in\mathcal{P}_t}\mathbb{E}\left\{\left\|d_i\mathbf{w}_{t}^i\right\|_2^2\right\},\\
    &=\frac{8\eta_t^2pbT^2\xi_1^2}{Nd^2}\sum_{i\in \mathcal{N}}\frac{1}{\epsilon_i^2}.
\end{split} 
\end{equation*}
The last equality holds because $\mathbf{w}_{t}^b$ are independent and $\mathbb{E}\left\{\mathbf{w}_{t}^b\right\}=0$.

\end{proof}

\subsection{Proof of Lemma~\ref{lemma:unbaised sample}}
\label{APP:Lemma1}
\begin{proof}
By taking expectation  of $\nu_t^b$, we obtain
\begin{equation}\label{equ:unbiased 1}
\begin{split}
\mathbb{E}\left\{\bar{\nu}_t^b\right\}&=\frac{N}{b}\mathbb{E}\left\{\sum_{i\in\mathcal{P}_t}\frac{d_i}{d}\nu_t^i\right\}\\
&=\frac{N}{b}\sum_{\mathcal{P}\in\phi_b(\mathcal{N})}\frac{1}{|\phi_b(\mathcal{N})|}\left(\sum_{i\in\mathcal{P}}\frac{d_i}{d}\nu_t^i\right),
\end{split}
\end{equation}
where $\phi_b(\mathcal{N})$ is the set of all subsets of $\mathcal{N}$ with size equal to $b$. Given $|\phi_b(\mathcal{N})|$ subsets, there are totally $|\phi_b(\mathcal{N})|b$ clients sampled and each client is sampled $\frac{cb}{N}$ times. Through exchanging the order of summation, one can get 
\begin{equation*}
        \sum_{\mathcal{P}\in\phi_b(\mathcal{N}) }\sum_{i\in\mathcal{P}}\frac{d_i}{d}\nu_t^i=\sum_{i\in\mathcal{N}}\frac{|\phi_b(\mathcal{N})|b}{N}\frac{d_i}{d}\nu_t^i.
\end{equation*}
By substituting this into \eqref{equ:unbiased 1}, we finish the proof.
\end{proof}

\subsection{Proof of Lemma~\ref{lemma:SGD variance}}
\label{APP:Lemma2}
\begin{proof}
Since we let $\mathcal{B}_t^i = \mathcal{D}_i$, from the definition of $\mathbf{g}_{t}^b$ and $\mathbf{g}_t$, one can get $\mathbf{g}_t=\sum_{i\in\mathcal{N}}\frac{d_i}{d}\nabla F_i(\theta_t^i)$ and $\mathbf{g}_t^b=\frac{1}{b}\sum_{i\in\mathcal{P}_t}\nabla F_i(\theta_t^i)$.

	\begin{equation*}
	\begin{split}
	&\mathbb{E}\left\{\left\|\mathbf{g}_t^b-\mathbf{g}_t\right\|_2^2\right\},\\
	&=\frac{1}{b^2}\mathbb{E}\left\{\left\|\sum_{i\in\mathcal{P}_t}\left(\nabla F_i(\theta_{t})-\mathbf{g}_t\right)\right\|_2^2\right\},\\
	&=\frac{1}{b^2}\mathbb{E}\left\{\left\|\sum_{i\in\mathcal{N}}\mathbb{I}(i\in\mathcal{P}_t)\left(\nabla F_i(\theta_{t})-\mathbf{g}_t\right)\right\|_2^2\right\},\\
	&=\frac{1}{b^2}\sum_{i\in\mathcal{N}}\mathbb{P}(i\in\mathcal{P}_t)\left\|\nabla F_i(\theta_{t})-\mathbf{g}_t\right\|_2^2+\\
	&\quad\frac{1}{b^2}\sum_{i,j\in\mathcal{N},i\neq j}\mathbb{P}(i,j\in\mathcal{P}_t)\big<\nabla F_i(\theta_{t})-\mathbf{g}_t,\nabla F_j(\theta_{t})-\mathbf{g}_t\big>,\\
	&=\frac{1}{bN}\sum_{i\in\mathcal{N}}\left\|\nabla F_i(\theta_{t})-\mathbf{g}_t\right\|_2^2+\\
	&\quad\frac{(b-1)}{bN(N-1)}\sum_{i,j\in\mathcal{N},i\neq j}\big<\nabla f_i(\theta_{t})-\mathbf{g}_t, \nabla F_j(\theta_{t})-\mathbf{g}_t\big>,\\
	&=\frac{N-b}{bN(N-1)}\sum_{i\in\mathcal{N}}\left\|\nabla F_i(\theta_{t})-\mathbf{g}_t\right\|_2^2,\\
	\end{split}
	\end{equation*}
	where $\mathbb{P}$ indicates the probability. We use $\mathbb{P}(i\in\mathcal{P}_t)=\frac{b}{N}$ and $\mathbb{P}(i,j\in\mathcal{P}_t,i\neq j)=\frac{b}{N}\frac{b-1}{N-1}$.
	Here $\mathbb{I}(i\in \mathcal{P}_t)$ is an indicator with value $0$ if $i$ is in $\mathcal{P}_t$ and $0$ otherwise. $\left< , \right>$ is the inner product operator . 
	
	The last equation holds because
		\begin{equation*}
		\begin{split}
		&\sum_{i\in\mathcal{N}}\left\|\nabla F_i(\theta_{t})-\mathbf{g}_t\right\|_2^2\\
		&+\sum_{i,j\in\mathcal{N},i\neq j}\big<\nabla F_i(\theta_{t})-\mathbf{g}_t,\nabla F_j(\theta_{t})-\mathbf{g}_t\big>\\
		&=\left(\sum_{i\in\mathcal{N}}\left(\nabla F_i(\theta_{t})-\mathbf{g}_t\right)\right)^2=0.
		\end{split}
		\end{equation*}
By using Assumption~\ref{assumption:variance of gradient}, we complete the proof.
\end{proof}

\subsection{Proof of Lemma~\ref{lemma:one step SGD}}
\label{APP:Lemma3}
\begin{proof}
	We start from the definition of $\bar{\nu}_{t+1}$,
	\begin{equation}\label{equ:SGD upper bound 0}
	\begin{split}
	\left\|\bar{\nu}_{t+1}-\theta^*\right\|_2^2
	&=\left\|\theta_{t}-\eta_{t} \mathbf{g}_t-\theta^*\right\|_2^2\\
	&=\left\|\theta_{t}-\theta^*\right\|_2^2+\eta_t^2\left\|\mathbf{g}_t\right\|_2^2\\
	&\quad-2\eta_t\left<\theta_{t}-\theta^*,\mathbf{g}_t\right>.
	\end{split}
	\end{equation}
	For the second term, using that $F_i$ is $\lambda$-smooth and $\eta_{t}\le\frac{1}{\lambda}$, we have
	\begin{equation}\label{equ:SGD upper bound 1}
	\begin{split}
	\eta_t^2\left\|\mathbf{g}_t\right\|_2^2&=\eta_t^2\left\|\sum_{i\in\mathcal{N}}\frac{d_i}{d}\nabla F_i(\theta_{t})\right\|_2^2\le\eta_t^2\sum_{i\in\mathcal{N}}\frac{d_i}{d}\left\|\nabla F_i(\theta_{t})\right\|_2^2\\
	&\le\eta_t^2\sum_{i\in\mathcal{N}}\frac{d_i}{d}2\lambda \left(F_i(\theta_{t})-F_i^*\right),\\
	\end{split}
	\end{equation}
	where we use that the norm is convex in the first inequality.
	Because $F_i$ is $\mu$-strong convex, the last term can be bounded as
	\begin{equation}\label{equ:SGD upper bound 2}
	\begin{split}
	&-2\eta_t\left<\theta_{t}-\theta^*,\mathbf{g}_t)\right>\\
	&=-2\eta_t\sum_{i\in\mathcal{N}}\frac{d_i}{d}\left<\theta_{t}-\theta^*,\nabla F_i(\theta_{t})\right>\\
	&\le-2\eta_t\sum_{i\in\mathcal{N}}\frac{d_i}{d}\left(F_i(\theta_{t})-F_i(\theta^*)+\frac{\mu}{2}\left\|\theta_{t}-\theta^*\right\|_2^2\right)\\
	&=-2\eta_t\sum_{i\in\mathcal{N}}\frac{d_i}{d}\left(F_i(\theta_{t})-F_i(\theta^*)\right)-\mu\eta_{t}\left\|\theta_{t}-\theta^*\right\|_2^2.\\
	\end{split}
	\end{equation}
	
	Combining (\ref{equ:SGD upper bound 1}) and (\ref{equ:SGD upper bound 2}) into (\ref{equ:SGD upper bound 0}), we obtain
	\begin{equation*}
	\begin{split}
	\left\|\bar{\nu}_{t+1}-\theta^*\right\|_2^2
	&\le(1-\mu\eta_{t})\left\|\theta_{t}-\theta^*\right\|_2^2\\
	&\quad+2\lambda\eta_t^2\sum_{i\in\mathcal{N}}\frac{d_i}{d} \left(F_i(\theta_{t})-F_i^*\right)\\
	&\quad-2\eta_t\sum_{i\in\mathcal{N}}\frac{d_i}{d}\left(F_i(\theta_{t})-F_i(\theta^*)\right).
	\end{split}
	\end{equation*}
	For the last two terms, we show that
	\begin{equation*}
	\begin{split}
	&2\lambda\eta_t^2\sum_{i\in\mathcal{N}}\frac{d_i}{d} \left(F_i(\theta_{t})-F_i^*\right)-2\eta_t\sum_{i\in\mathcal{N}}\frac{d_i}{d}\left(F_i(\theta_{t})-F_i(\theta^*)\right)\\
	&=2\eta_{t}(\lambda\eta_{t}-1)\sum_{i\in\mathcal{N}}\frac{d_i}{d}\left(F_i(\theta_{t})-F(\theta^*)\right)+2\lambda\eta_{t}^2\Gamma\\
	&=2\eta_{t}(\lambda\eta_{t}-1)\left(F(\theta_{t})-F(\theta^*)\right)+2\lambda\eta_{t}^2\Gamma.\\
	\end{split}
	\end{equation*}
	Using that $\eta_{t}\le\frac{1}{\lambda}$ again, we have $2\eta_{t}(\lambda\eta_{t}-1)\le0$ and $F(\theta_{t})-F(\theta^*)\ge0$. We finally obtain
	\begin{equation*}
	\left\|\bar{\nu}_{t+1}-\theta^*\right\|_2^2\le(1-\mu\eta_{t})\left\|\theta_{t}-\theta^*\right\|_2^2+2\lambda\eta_{t}^2\Gamma.
	\end{equation*}
\end{proof}

\subsection{Proof of Theorem~\ref{theorem:SGD upper buond}}

\label{APP:Theorem3}
\begin{proof}

Starting from the updating rule in FedSGD-DPC, we have
\begin{equation}\label{equ:lower bound 0}
\begin{split}
\left\|\theta_{t+1}-\theta^*\right\|_2^2
&=\left\|\bar{\nu}_{t+1}^b-\theta^*-\mathbf{w}_t^b\right\|_2^2\\
&=\left\|\bar{\nu}_{t+1}^b-\theta^*\right\|_2^2+\left\|\mathbf{w}_t^b\right\|_2^2\\
&\quad-2\left<\bar{\nu}_{t+1}^b-\theta^*,\mathbf{w}_t^b\right>.\\
\end{split}
\end{equation}
The second term is the variance of the aggregated DP noises.   hown in Proposition~\ref{pro:noise} and the expectation of the last term is zero. For the first term, we have
\begin{equation}\label{equ:lower bound 1}
\begin{split}
\left\|\bar{\nu}_{t+1}^b-\theta^*\right\|_2^2
&=\left\|\nu_{t+1}^b-\bar{\nu}_{t+1}+\bar{\nu}_{t+1}-\theta^*\right\|_2^2,\\
&=\left\|\nu_{t+1}^b-\bar{\nu}_{t+1}\right\|_2^2+\left\|\bar{\nu}_{t+1}-\theta^*\right\|_2^2\\
&\quad+2\left<\nu_{t+1}^b-\bar{\nu}_{t+1},\bar{\nu}_{t+1}-\theta^*\right>.
\end{split}
\end{equation}
Then, one can derive
\begin{equation*}
    \begin{split}
        \left\|\theta_{t+1}-\theta^*\right\|_2^2&=\left\|\nu_{t+1}^b-\bar{\nu}_{t+1}\right\|_2^2+\left\|\bar{\nu}_{t+1}-\theta^*\right\|_2^2+\left\|\mathbf{w}_t^b\right\|_2^2\\
        &\quad-2\left<\bar{\nu}_{t+1}^b-\theta^*,\mathbf{w}_t^b\right>\\
        &\quad+2\left<\nu_{t+1}^b-\bar{\nu}_{t+1},\bar{\nu}_{t+1}-\theta^*\right>.\\
    \end{split}
\end{equation*}
Taking expectation on both sides and using Lemma~\ref{lemma:unbaised sample} and~\ref{lemma:one step SGD}, we obtain
\begin{equation*}
    \begin{split}
        Y_{t+1}\le&(1-\mu\eta_t)Y_t+2\eta_t^2\lambda\Gamma+\mathbb{E}\left\{\left\|\mathbf{w}_t^b\right\|_2^2\right\}\\
        &+\mathbb{E}\left\{\left\|\nu_{t+1}^b-\bar{\nu}_{t+1}\right\|_2^2\right\},\\
        =&(1-\mu\eta_t)Y_t+2\eta_t^2\lambda\Gamma+\mathbb{E}\left\{\left\|\mathbf{w}_t^b\right\|_2^2\right\}\\
        &+\eta_t^2\mathbb{E}\left\{\left\|\mathbf{g}_t^b-\mathbf{g}_t\right\|_2^2\right\},\\
    \end{split}
\end{equation*}
where we use $\mathbb{E}\left\{\mathbf{w}_t^b\right\}=0$.
Plugging Proposition~\ref{pro:noise} and Lemma~\ref{lemma:SGD variance}, we obtain
$Y_{t+1}\le(1-\mu\eta_{t})Y_t+\eta_{t}^2 (2\frac{N-b}{N-1}\frac{\sigma^2}{b}+2\lambda\Gamma+\frac{8pbT^2\xi_1^2}{Nd^2}\sum_{i\in \mathcal{N}}\frac{1}{\epsilon_i^2})$.

Let $\omega = 2\frac{N-b}{N-1}\frac{\sigma^2}{b}+2\lambda\Gamma+\frac{8pbT^2\xi_1^2}{Nd^2}\sum_{i\in \mathcal{N}}\frac{1}{\epsilon_i^2}$ and $\eta_t=\frac{\alpha}{t+\gamma}$, where $\alpha>\frac{1}{\mu}$ and $\gamma>0$ such that $\eta_0\le\min(\frac{1}{\mu},\frac{1}{\lambda})=\frac{1}{\lambda}$, we have
\begin{equation*}
    Y_{t}\le \frac{\varphi}{t+\gamma},
\end{equation*}
where $\varphi=\max(\frac{\alpha^2\omega}{\alpha\mu-1},\gamma Y_0)$.
See detailed proof in Theorem 1 in~\cite{Li2020On}. 
By setting $\alpha=\frac{2}{\mu},\gamma=2\frac{\lambda}{\mu}$, we obtain

\begin{equation*}
    \begin{split}
          Y_t\le\frac{1}{\mu}\frac{1}{t+\gamma}\left(\frac{4}{\mu}\omega+2\lambda Y_0\right).
    \end{split}
\end{equation*}
The proof can be completed by rearrange the right hand side of the above inequality. 
\end{proof}

\subsection{K.K.T. conditions for $U(T,b)$}\label{sec:K.K.T.}
We define the Lagrangian function of $U(T, b)$ as
\begin{equation*}
\mathcal{L}(T,b,\rho_1,\rho_2,\rho_3)=U(T,b)-\rho_1T+\rho_2(1-b)+\rho_3(b-N),
\end{equation*}
where $\rho_1$, $\rho_2$ and $\rho_3$ are Lagrangian multipliers.
The K.K.T. conditions of $\mathcal{L}(T,b,\rho_1,\rho_2,\rho_3)$ are
\begin{equation*}
\begin{split}
\frac{2C_2bT}{T+\gamma}-\frac{1}{(T+\gamma)^2}\left(\frac{C_1}{b}+C_2bT^2+C_3\right)-\rho_1&=0\\
\frac{1}{T+\gamma}\left(C_2T^2-\frac{C_1}{b^2}\right)-\rho_2+\rho_3&=0\\
\rho_1T=0,\,\rho_2(1-b)=0,\,\rho_3(b-N)&=0\\
-T\le0,\,1-b\le0,\,b-N&\le0\\
\rho_1,\,\rho_2,\,\rho_3&\ge0
\end{split}
\end{equation*}

\subsection{Proof of Lemma~\ref{lemma:SGD variance q}}
\label{APP:Lemma4}
\begin{proof}
We start from the definition of $\mathbf{g}_{t}^{b, q}$,
\begin{equation}
	\begin{split}
\left\|\mathbf{g}_t^{b, q} - \mathbf{g}_t\right\|_2^2	&=	\left\|\mathbf{g}_t^{b, q} -\mathbf{g}_t^q+\mathbf{g}_t^q  -\mathbf{g}_t\right\|_2^2,\\
&=\left\|\mathbf{g}_{t}^{b,q}-\mathbf{g}_t^q\right\|_2^2+\left\|\mathbf{g}_t^q  -\mathbf{g}_t\right\|_2^2\\
	&\quad+2\eta_t\left<\mathbf{g}_{t}^{b,q}-\mathbf{g}_t^q, \mathbf{g}_t^q  -\mathbf{g}_t\right>
	\end{split}
	\end{equation}
By taking expectation on both sides, the first term has been bounded in Lemma~\ref{lemma:SGD variance}, while the second term can be bounded by $\sum_{i\in \mathcal{N}} \frac{d_i}{qd^2} \Lambda_i^2$ by utilizing Lemma 2 in \cite{}.  The last term is $0$ since random samples generate unbiased estimation of global values in expectation, and thus $\mathbb{E}\{\mathbf{g}_{t}^{b,q}-\mathbf{g}_t^q\} = \mathbb{E}\{\mathbf{g}_t^q  -\mathbf{g}_t\} =0$.
\end{proof}

\subsection{Proof of Theorem~\ref{theorem:SGD upper buond Gau}}
\label{APP:Theorem4}
\begin{proof}

Following the proof of Theorem~\ref{theorem:SGD upper buond}, we have
\begin{equation*}
    \begin{split}
        Y_{t+1}\le&(1-\mu\eta_t)Y_t+2\eta_t^2\lambda\Gamma+\mathbb{E}\left\{\left\|\mathbf{w}_t^b\right\|_2^2\right\}\\
        &+\mathbb{E}\left\{\left\|\nu_{t+1}^b-\bar{\nu}_{t+1}\right\|_2^2\right\},\\
        =&(1-\mu\eta_t)Y_t+2\eta_t^2\lambda\Gamma+\mathbb{E}\left\{\left\|\mathbf{w}_t^b\right\|_2^2\right\}\\
        &+\eta_t^2\mathbb{E}\left\{\left\|\mathbf{g}_t^{b, q}-\mathbf{g}_t\right\|_2^2\right\},\\
    \end{split}
\end{equation*}
where we use $\mathbb{E}\left\{\mathbf{w}_t^b\right\}=0$. Plugging Proposition~\ref{pro:noiseGau} and and Lemma~\ref{lemma:SGD variance q}, we obtain
$Y_{t+1}\le(1-\mu\eta_{t})Y_t+\eta_{t}^2 (2\frac{N-b}{N-1}\frac{\sigma^2}{b}+\sum_{i\in \mathcal{N}} \frac{d_i}{qd^2} \Lambda_i^2+ 2\lambda\Gamma+\frac{c_2^2pT\xi_2^2}{d^2}\sum_{i\in \mathcal{N}} \frac{1}{\epsilon_i^2}\log{\frac{1}{\delta_i}})$.

Let $\omega' =  2\frac{N-b}{N-1}\frac{\sigma^2}{b}+\sum_{i\in \mathcal{N}} \frac{d_i}{qd^2} \Lambda_i^2+ 2\lambda\Gamma+\frac{c_2^2\eta_t^2pT\xi_2^2}{d^2}\sum_{i\in \mathcal{N}} \frac{1}{\epsilon_i^2}\log{\frac{1}{\delta_i}}$ and $\eta_t=\frac{\alpha}{t+\gamma}$, where $\alpha>\frac{1}{\mu}$ and $\gamma>0$ such that $\eta_0\le\min(\frac{1}{\mu},\frac{1}{\lambda})=\frac{1}{\lambda}$. Following the proof of Theorem~\ref{theorem:SGD upper buond}, we derive $Y_{t}\le \frac{\varphi}{t+\gamma}$ again where $\varphi=\max(\frac{\alpha^2\omega'}{\alpha\mu-1},\gamma Y_0)$.
By setting $\alpha=\frac{2}{\mu},\gamma=2\frac{\lambda}{\mu}$, we obtain $  Y_t\le\frac{1}{\mu}\frac{1}{t+\gamma}\left(\frac{4}{\mu}\omega'+2\lambda Y_0\right)$. The proof is completed by rearrange the right hand side of the inequality. 

\end{proof}

\subsection{Notation List}

To facilitate the understanding of our analysis, most notations used in our paper are listed in the table with a brief explanation. 
\begin{table}[!htbp]
\renewcommand\arraystretch{1.2}
\centering
\caption{Major notations used in the paper}
\label{tab:notations}
\begin{tabular}{|c|c|}
\toprule
Symbol & Description\\
\midrule
$\mathcal{N}$ & the set of all clients \\
\hline
$N$ & the number of all clients \\
\hline
$p$ & the dimension of model parameters\\
\hline
$i$ & the id of a particular client \\
\hline
$d_i$ & the number of data samples owned by client $i$\\
\hline
$d$ & the total number of data samples of all clients\\
\hline
$\mathcal{D}_i$ & the set of data samples in client $i$\\
\hline
$\zeta$ & represents a sample\\
\hline
$b$ & the number of clients participate each global iteration\\
\hline
$\mathcal{B}_{t}^i$ & the sample batch selected by client $i$ \\
& to participate global iteration $t$\\
\hline
$q$ & the sampling probability of clients \\
& using the Gaussian mechanism\\
\hline
$\theta_t^i$ & model parameters owned by client $i$ after $t$ iterations\\
\hline
$\theta^*$ & the optimal model parameters that can\\
 & minimize the global loss function\\
\hline
$\bar{\theta}_t$ & the global average of model parameters over \\
& all clients  after $t$ iterations\\
\hline
$\nu_t^i$ & the local model parameters in client $i$ after $t$ iterations\\
\hline
$\bar{\nu}_t$ & the global average  of  local model parameters \\
& over all clients after $t$ iterations\\
\hline
$F$ & the loss function\\
\hline
$\nabla F$ & the gradients of function $F$\\
\hline
$\bar{\mathbf{g}}_t$ &  global gradients computed with the complete dataset \\
\hline
$\bar{\mathbf{g}}_t$ &  global gradients computed with the complete dataset over \\
& all clients after $t$ iterations \\
\hline
$\bar{\mathbf{g}}_t^b$ &  gradients computed over the complete dataset of $b$\\
& participating clients after $t$ iterations \\
\hline
$\bar{\mathbf{g}}_t^{b,q}$ &  gradients computed over the sampled dataset of $b$\\
& participating clients after $t$ iterations \\
\hline
$t$ & the index of global iterations \\
\hline
$\eta_t$ & the learning rate at $t^{th}$ iteration\\
\hline
$T$ & the total number of global iterations\\
\hline
$\Gamma$ & the non-iid degree of teh data sample distribution\\
\hline
$\epsilon_i,\delta_i$ & privacy budget of client $i$\\
\hline
$\sigma_i$ & the standard deviation to generate Gaussian noises\\
& by client $i$ \\
\hline
$\mathbb{I}_p$ & the identity matrix with dimension $p$ \\
\hline
$\mathcal{M}$ & indicate a DP mechanism\\
\hline
$\xi_1$ & the sensitivity  of model parameters\\ &using the Laplace mechanism\\
\hline
$\xi_2$& the sensitivity  of model parameters \\
& using the Gaussian mechanism  \\
\hline
$\mathbf{w}_t^i$ & the DP noises generated by client $i$ in global iteration $t$\\
\hline
$\mathbf{w}_t^b$ & the aggregated DP noises over $b$ participating \\
& clients in global iteration $t$\\
\hline
 $Y_t$ & the expected gap between model parameters after $t$ \\
 &iterations and the optimal model parameters\\
\hline
 $G$ & the bound of the expected squared norm of\\
 & stochastic gradients in clients\\
\hline
$\lambda$ & the smooth constant of the loss function $F$\\
\hline
$\mu$ & the strongly convex constant of the loss function $F$\\
\hline
$\Lambda_i$ & the variance of stochastic gradients in client $i$  \\
\hline
$U(T,b)$ & the function to optimize $T$ and $b$ \\
\hline
$T^*$  & the optimal number of global iterations \\
\hline
$b^*$  & the optimal number of participating clients per iteration\\
\bottomrule
\end{tabular}
\end{table}

\end{document}